\documentclass[School=Harvard]{Dissertate}

\newcommand{\urlcolor}[1]{\textcolor{SchoolColor}{#1}}
\newcommand{\notered}[1]{\textcolor{black}{#1}}
\newcommand{\noteblue}[1]{\textcolor{black}{#1}}
\newcommand{\notegreen}[1]{\textcolor{black}{#1}}

\usepackage{fancyhdr}
\usepackage{lastpage}

\usepackage{mdframed}

\usepackage{chngcntr} %
\counterwithout{footnote}{chapter}

\usepackage{enumitem}

\setlist[itemize]{nosep} %

\usepackage{wrapfig}
\usepackage{tabularx}
\usepackage{tabularray}

\usepackage{type1cm}        %
\usepackage{makeidx}         %
\usepackage{graphicx}        %
\usepackage[bottom]{footmisc}%

\usepackage{lipsum}
\usepackage{amsmath}
\usepackage{refcount} %

\usepackage[capitalise]{cleveref} %

\usepackage{keyfloat}
\usepackage{listings}
\definecolor{lightgray}{gray}{0.9}
\usepackage{amsmath}
\usepackage{siunitx}[=v2] %

\usepackage{booktabs}
\usepackage{longtable}
\usepackage{tikz} %
\usepackage{etoolbox} %

\usepackage{array} %
\usepackage{tabularx} %

\usepackage[T1]{fontenc}    %
\usepackage{cleveref}       %
\usepackage{wrapfig}
\usepackage{csquotes}
\MakeOuterQuote{"} %

\usepackage{babel}

\usepackage{array}

\makeindex             %

\newcolumntype{P}[1]{>{\centering\arraybackslash}p{#1}}

\usepackage{lmodern}
\usepackage{textcomp}

\usepackage{stringenc}
\newcommand*{\uni}{}
\DeclareRobustCommand*{\uni}[1]{%
  \begingroup
    \StringEncodingConvert\x{%
      \pdfunescapehex{%
        00%
        \ifnum"#1<"100000 0\fi
        \ifnum"#1<"10000 0\fi
        \ifnum"#1<"1000 0\fi
        \ifnum"#1<"100 0\fi
        \ifnum"#1<"10 0\fi
        #1%
      }%
    }{utf32be}{utf8}%
    \everyeof{\noexpand}%
    \endlinechar=-1 %
  \edef\x{%
    \endgroup
    \scantokens\expandafter{%
      \expandafter\unexpanded\expandafter{\x}%
    }%
  }\x
}

\usepackage{hwemoji}

\usepackage{tcolorbox} %

\usepackage[disable]{todonotes}

\newcommand{\code}[1]{\lstinline{#1}}

\ExplSyntaxOn

\newcounter{savedenv}
\seq_new:N \g_ephraim_savedenv_seq

\NewDocumentEnvironment{savedenv}{ o +b }
 {
  \refstepcounter{savedenv}
  
  \IfValueT{#1}{\label{savedenv@#1}}
  \seq_gput_right:Nn \g_ephraim_savedenv_seq { #2 }
}{}

\NewDocumentCommand{\printsaved}{m}
 {%
  \clist_map_inline:nn { #1 }
   {
    \seq_item:Nn \g_ephraim_savedenv_seq { \getrefnumber{savedenv@##1} }
    \par
   }
 }
\NewDocumentCommand{\printallsaved}{}
 {
  \seq_use:Nn \g_ephraim_savedenv_seq { \par }
 }

\ExplSyntaxOff

\begin{document}
\pagestyle{plain}

\title{\uni{1F590} Machine Learning for Tangible Effects: Natural Language Processing for Uncovering the Illicit Massage Industry \& Computer Vision for Tactile Sensing}

\author{Rui Nancy Ouyang}
\advisor{Finale Doshi-Velez}
\advisor{David Parkes}
\advisor{Roberto Rigobon}

\degree{Doctor of Philosophy}
\field{Computer Science}
\degreeyear{2023}
\degreemonth{Sep.}
\department{School of Engineering and Applied Sciences}

\pdOneName{B.S.}
\pdOneSchool{Massachusetts Institute of Technology}
\pdOneYear{2013}

\pdTwoName{M.S.}
\pdTwoSchool{Harvard University}
\pdTwoYear{2019}
\maketitle
\begingroup
\hypersetup{linkcolor=black, urlcolor=black}
 \copyrightpage
 \abstractpage
 \tableofcontents
 \listoffigures 
 
	\newpage \thispagestyle{fancy} \vspace*{\fill}
	\scshape \noindent

To my parents. Though I am not always forthcoming about all the trials and tribulations in my life, I know deep down in my heart that I can always turn to both of you. Your love gives me the courage to keep moving forward through life.

	\vspace*{\fill} \newpage \rm

	\chapter*{Acknowledgments}
	\noindent
	This endeavour would not have been possible without my advisor, Prof. Roberto Rigobon, who has been a great source of support as I redefined my research focus and without whom my research would not be possible. I am also extremely grateful to my committee, Prof. Finale Doshi-Velez and Prof. David Parkes, as they came to my aid through this journey and provided consistent gentle encouragement to finish and graduate, as well as Prof. Milind Tambe for providing valuable feedback during my qualification exams. I would like to express my deepest gratitude to my collaborators, John McGrath at IBM and my collaborator and mentor, Julie Braun. I am indebted to my best friends Marcela Rodriguez and Irina Tolkova for giving me the strength to continue my graduate studies during the pandemic (and holding me accountable to it). My roommate Erons Ohienmhen gave invaluable support in proofreading and logistical support, and my other roommates Arianna McQuillen, Ond\v{r}ej B\'{i}\v{z}a, and Gagan Khandate kept my life in balance the last few years with hiking and sailing trips. Thank you to my partner, Diony Rosa, for the everyday support and reminding me of all the good in humanity whenever my work made me feel cynical. Mark York, Eric Lu, Mark Goldstein, David Holland, Anitha Gollamundi, for late nights and long discussions. Thank you to John Aleman, for so many sailing and life adventures, Sarah Cheng and Ilia Lebedev for listening to me vent, and Eric Marion for helping me actually get the thesis out the door. Thank you to Sasha Zaranek, who convinced me to apply to graduate school, and Prof. Sangbae Kim and Albert Wang for having me as an assistant while I waited to start graduate school. I would also like to thank John Girash, who helped me navigate the incredible bureaucracy of Harvard. A shout-out to Radhen Patel, working with you was a highlight of my graduate career. Thanks to my psychiatrist, Blake Ritter, and a thank you to Ankur Mehta, who gave me a place to stay for several months so that I could softly land back in reality. I can only hope to pay it forward one day. Thank you also to Zoz Brooks, for supporting my robotics adventures, and to Elsa Riachi and Carrie Chai at Scotiabank. Thanks to Ellie Simonson and Lilly Chin at MIT, who helped me navigate the process of shifting my research focus. Thank you so much to all my friends and acquaintances who found my work cool and exciting when I didn't -- your enthusiasm kept me going. Thank you to all the security and janitorial staff, and finally, thank you to my cat, Rosie -- your high-fives earned respect and smiles from many people in my life. I cannot hope to fully acknowledge everyone who contributed to this thesis, and will endeavour to make sure I thank people in real life.

\pagebreak

\section*{Declaration of Authorship}

The work in \Cref{chap:reviews} and \Cref{chap:datasets} is original independent work by the author, Rui Ouyang, as advised by Professor Roberto Rigobon. The work in \Cref{chap:synthetic} is original independent work by the author Rui Ouyang.  A version of \Cref{chap:digger} has been published as "Digger Finger: GelSight Tactile Sensor for Object Identification Inside Granular Media" at the 17th International Symposium on Experimental Robotics (ISER), Malta, Nov. 15 to 18, 2021 \cite{patel2021digger}. The following authors contributed: Radhen Patel, Rui Ouyang, Branden Romero, Edward Adelson. For \Cref{chap:digger}, my contributions included physical prototyping that led to the final iteration described in this paper, as well as creating the supplementary materials listed in \Cref{sec:highlights}. Finally, the work in \Cref{chap:fiducial} is original independent work by the author, Rui Ouyang, as advised by Professor Robert D. Howe. \noindent A version of \Cref{chap:fiducial} has been published as "Low-Cost Fiducial-Based 6-axis Force-Torque Sensor" at the 2020 International Conference on Robotics and Automation (ICRA), held virtually \cite{Ouyang2020LowCostF6}.

	\vspace*{\fill} \newpage
	\setcounter{page}{1}
	\pagenumbering{arabic}

\endgroup

\doublespacing

\pagestyle{fancy}
\fancyhf{} %
\fancyhead[R]{\rightmark}
\fancyfoot[C]{\thepage} %
\renewcommand{\headrulewidth}{0.5pt}

\setcounter{chapter}{-1}  %

\chapter{Unexpected Directions: An Introduction}
\label{ch:intro}
\chaptermark{Thesis Introduction}

\section{The Questions}

I enjoy creative applications of serious technical research, and in my thesis I consider the following two questions:

\textsc{How can computer science be used to fight human trafficking?}  
and \\
\textsc{How can camera images turn into a sense of touch?} \\

\section{The Answers}

\textbf{How can computer science be used to fight human trafficking?} In the first two chapters, I present my work on:

\begin{enumerate}
    \item Finding indicators of illicit activity in Google Reviews
    \item Creating (and exploring the import of) a dataset of activity on a public forum for discussing illicit activity 
\end{enumerate}

I will specifically focus on sex trafficking in the illicit massage industry (although labor trafficking often goes hand-in-hand). Additionally, in the third chapter I cover how counter-trafficking is addressed in the banking and financial sectors. 
    
\begin{enumerate}[resume]
    \item Creating synthetic transaction (banking) data that mimics illegal activity
\end{enumerate}

\textbf{How can camera images turn into a sense of touch?} In the last two chapters, I present my work on:

\begin{enumerate}[resume]
    \item Using fiducial markers to turn movement into force and torque measurements
    \item Designing (Gelsight) stereophotogrammetry sensors for digging in sand and other granular environments
\end{enumerate}

\section{Overview}

\color{black}
In this section, I provide an overview of the concepts and motivation behind each of my chapters.

In the first chapter, I consider the erotic massage industry (which I refer to as the illicit massage industry, or IMI, given their legal status in the United States), where commercial sexual services are offered in addition to massage services. Although not all illicit massage businesses (IMBs) are settings for human trafficking, a better understanding of this industry would help create regulations and policy that could improve working conditions in the IMI in pursuit of a more just society. In this chapter, I investigate what information about the IMI can be learned from Google Maps, as opposed to websites specialized to the IMI.\footnote{Other IMI researchers previously considered the Yelp Reviews dataset. However, the default Yelp dataset contains information only for 11 metropolitan areas. We focus on Google Maps to create a dataset that covers all fifty states (and that could potentially extend to countries around the world, which is of interest since many employees in the IMI are immigrants).} From collaborators at The Network, I received a listing of business locations on a website specific to the IMI (Rubmaps), these locations cross-referenced against Google Maps.
By retrieving the review text on Google Maps for each cross-referenced location, I created a text corpus to analyze with natural language processing (NLP). Given the lack of a centralized list of massage parlors across the United States,  I then  queried Google Maps for additional locations for each city in our dataset, and retrieved reviews for these locations. The results demonstrate that  simple NLP techniques such as bag-of-words \cite{harrisDistributionalStructureBagWords1954} can be used to  detect illicit activity from Google Maps review text. I also consider certain hypotheses, such as whether reviewers would be more likely to mention money or ethnicity for IMBs (reflecting potentially more workers more vulnerable to trafficking), and evaluate the strength of the NLP classifier against simpler models such as simply looking at the businesses’ opening hours (also provided by Google Maps). I also demonstrate that  removing mentions of sensitive features (ethnicity or language) from the reviews dataset has no significant  impact  on the performance of the NLP classifier. In summary, this work contributes a classification model for IMBs that is the first to be trained on data from all fifty states instead of a subset,  as well as the first analysis of business hours as a potential feature for distinguishing illicit from licit massage parlors.

Whereas Chapter 1  focuses on the supply side of the IMI (what businesses were illicit or not),  the second chapter adopts a different viewpoint and considers what online text data is available about the demand side.    Multiple online forums exist where buyers of sexual services at IMIs (self-referenced as “mongers”) gather to discuss topics (e.g., law enforcement action or business turnover) and exchange often graphic reviews of sex workers. These forums potentially accelerate and normalize toxic beliefs and attitudes toward the IMI, such as those that may have led to the mass-shooting deaths of several people at massage parlors in Atlanta in 2020. The focus is on creating a dataset involving completely public data such that it could be freely released to the computer science community for further analysis. I  downloaded all posts at the forum AMPReviews.net and show three exploratory use cases of the dataset: expanding acronyms, investigating related domain names, and assessing monger’s concerns about their relationships. I also give possible new research questions that could be addressed with this dataset, ranging from IMI-specific questions (e.g., what is the distribution of user visits and spending? are there geographical hot spots?) to more general computational social science questions (e.g., do users normalize toxic attitudes toward the IMI?).

The third chapter  considers counter-trafficking in the financial sector. Human trafficking at its heart is an economic activity, and eventually the proceeds of trafficking will enter the formal financial system (a.k.a., money laundering). Thus, banks have a unique role to play in counter- trafficking. Furthermore, banks are under heavy regulatory pressure, facing fines in the range of hundreds of millions of dollars,  to institute strong anti-money laundering controls. Based on the author’s experiences at Scotiabank, which is one of the three largest banks in Canada, synthetic data could have helped assess the capability of cutting-edge academic tools and how they apply to real-life data at scale, where the truth label is lazy and potentially incredibly sparse. The work in this chapter  considers how to create synthetic data using an agent-based model, with behavioral assumptions drawn from regulatory agency operational alerts. Specifically, the model considers normal agents that transact with a Gaussian probability centered during the day and where suspicious agents transact with Gaussian probability centered in the evening. This supports the study of downstream applications such as anomaly detection, comparing the performance of different models such as isolation forest and Gaussian mixture models on this synthetic data. I then show how agent- based models can be modified to output network information about who transacts with who,  and incorporate additional agent types to   allow variation in  the likelihood of transactions between different agent types. I release the model code for public use.

The final two chapters (chapters four and five), turn to the question of creating a sense of touch from computer vision. Over the past decade, legged locomotion has significantly improved and to some extent research has moved to industry, and  the academic frontier of robotics  is moving from locomotion  to manipulation. Manipulation remains a relatively unsolved problem. For instance, manipulation is very helpful in  creating robots that are useful in the home, which is an important consideration given the aging population in industrialized nations. Although many manipulation objectives can be achieved by vision alone, such systems tend to be limited in the types of objects they can manipulate (e.g., glass can pose difficulties, as well as obstructed areas or objects hidden behind other objects). A sense of touch can be vital to manipulation in robotics, e.g., for detecting slip and recovering before an object is dropped.

Chapter 4  considers advances to the Gelsight-based line of sensors. Gelsight sensors consist of a miniature camera pointed at a piece of rubber-like clear gel that has a semi-specular spray coating on the surface. The gel is bonded to a clear acrylic structure that houses the camera. The acrylic acts as a light guide for three colors of light that come from three different directions. The light then bounces off the semi-specular coating, throwing into sharp contrast any object pressed into the gel. By contrasting the colors of shadows in the camera image, we can develop an understanding of the 3D structure (using stereo photometry), thereby creating a “tactile image”. This work  considers the advances in technology that are required to apply the Gelsight to a granular environment, for instance finding cables buried in sand. For this, the sensor is  miniaturized  by replacing two colors of LEDs with fluorescent paint, so that we only require  circuitry at the top of the sensor instead of along the sides. Another change is to  create a wedge-shape at the tip to facilitate digging through sand, trading-off evenly balanced illumination (which would make reconstructing a 3D model easier) with adapting to the environment. With these changes to the default Gelsight sensor, the sensor is still able to identify different shapes  through a suitably trained convolutional neural network. A vibrator motor is also introduced, which  allows the sensor to bypass the jamming nature of granular media.

The pros of this technology include easy adaptability to many shapes, by simply changing the shape of the gel surface and clear acrylic substrate, as well as its high resolution. The downsides include the need to cast and spray-coat elastomers, as well as durability issues and the need for custom circuits to power the LEDs.

The coating can be rubbed off by constant touch and the elastomer damaged by sharp or rough objects. In such cases, the sensor must be replaced manually (this contrasts  with our fingers, which constantly rejuvenate). Thus, in the fifth and final chapter, I consider a tactile sensor that provides less detailed information (six degrees-of-freedom, instead of thousands of multi-color pixels) but could be more suitable for applications where durability is important.

Force-torque sensors are widely used on existing industrial and collaborative robots, including being integrated into grippers and other devices on robot arms. In fact, they are available as plug-in additions to popular robot arm series like the Universal Robotics arms. However, these sensors are often tens of thousands of dollars. This work creates  an inexpensive alternative using consumer webcams and 3D printing, thus opening up six-axis force-torque sensing to a wider range of applications (e.g. should one need dozens or hundreds of such sensors). Specifically, a consumer webcam is removed from its casing and pointed  at a platform “floating” above it on four springs. Due to the flexibility of the springs, the platform can freely translate and rotate in 3D space. A 2D printed tag is glued to the underside (the side facing the camera) of the platform. Computer vision is used to  convert the camera image of the tag to a 6D pose (XYZ location and roll, pitch, and yaw). With calibration against known weights and torques, the 6D pose converts to three axes of force and three axes of torque information. The downsides are that calibration is required for the sensor to get absolute force and torque readings (in order to translate displacement to force and rotation to torque), and there are also  non-linearity and hysteresis issues. The benefits of this sensor is  its low cost and ease of fabrication, requiring only a consumer-grade 3D printer, a 2D printer, springs, and glue. In this chapter, I characterize this novel sensor design, and also report on the open sourcing of the design files.

\section{Highlight to Other Researchers}
\label{sec:highlights}

Section 0.3.1 details publicly-accessible source code, datasets, and design files that were produced during the course of this research and are public resources.

\begin{enumerate}
    \item A portion of the code for the first chapter can be found at \url{ https://github.com/nro-bot/illicit_massage_code}, with the final release of the Google Places code planned by the end of 2023.
    \item Code for this chapter may be found at \url{https://github.com/nro-bot/imi_datasets/}, where final release of the dataset planned for end of 2023. Information the results of any related hackathon(s) will be found at \url{https://hack4fem.github.io} or linked from \url{https://nrobot.dev}.
    \item  For the agent-based model, the code may be found at \url{https://github.com/nro-bot/fake-banking-data}.
    \item Although no design files are provided for this sensor itself, the published paper associated with this chapter \cite{Ouyang2020LowCostF6} includes an “exploded CAD” (hardware design) file and details of the LEDs and paint used, and this should be sufficient to replicate the results. 
    \item For the final chapter, the hardware and software files are released at \url{https://sites.google.com/view/fiducialforcesensor}, and specifically at the link \url{https://www.dropbox.com/sh/cnoyitxwunrdk00/AAB_jrrbwQmEwhs7nj5Z7ypsa}.
\end{enumerate}
\normalcolor

\chapter{Illegal Activity in Plain Sight: Finding Erotic Massage Businesses with Google Places Reviews} \label{chap:reviews}
\chaptermark{Illegal Activity in Plain Sight}

\captionsetup{labelfont=bf}

\graphicspath{{./figs/reviews/}}

\renewcommand{\KFLTtightframe}[1]{%
    \begin{minipage}{\KFLTimageboxwidth}
    \tcbox[size=fbox, colframe=black!70, arc=0mm, colback=white]{
    #1
    }
    \end{minipage}
}

\begin{savedenv}[fig_bow]
    \medskip
    \keyfig[H]{lw=0.5,ft,c={%
    Confusion matrix for Part 1, running a bag-of-words model with a logistic regression classifier.  Overall accuracy comes out to approximately 80\%. Note: Counts are averaged across 5-folds.},
    l=fig:bow%
    }{result_logreg.png}
\end{savedenv}

\begin{savedenv}[fig_baselines]
    \keyfig[H]{lw=0.8,f,c={%
    Confusion matrices for using the binary heuristic features directly
    as baselines (not as inputs into models).},
    l=fig:baselines%
    }{CONFUSION_featurepresence_with_acc.png}
\end{savedenv}

\begin{savedenv}[fig_barplot]
\medskip
   \keyfig[H]{lw=0.7,ft,c={%
    Barplot showing feature distribution for the four manually created
    binary features between labels 0 and 1 classes. All four show significant
    distributional differences.
    },l=fig:barplot%
    }{BARPLOT_covariates.png}
\end{savedenv}

 \begin{savedenv}[fig_hours]
    \keyfig[H]{f,lw=1,c={%
Heatmap of Closing Hours for the two labels. Note that values are
    raw numbers not normalized by the class imbalance. Based on this graph, we
    derived a threshold of 9pm as our "closing late" feature, and added an
    additional "open seven days a week" feature.%
         },
         l=fig:hours%
         }{closing_hours.png}
\end{savedenv}

\begin{savedenv}[fig_rmaps_homepage]
    \medskip
    \keyfig[H]{lw=0.8,ft,c={%
      \notered{Screenshot of a page from Rubmaps. For this chapter, business name and location data are used, while forum, name, price, and the review text are not used (the latter is behind a paywall).}},
      l=fig:rmaps_homepage%
      }{Rubmaps_censored.jpg}%
\end{savedenv}

\begin{savedenv}[fig_pipeline]
    \medskip
   \keyfig[H]{lw=0.8,ft,c={%
   \color{black}
   Diagram of classification pipeline for the bag-of-words models. Up to five reviews per business are returned by Places. The text is then concatenated, lowercased, non-alphabetic characters are removed, and stemmed. The text is then vectorized with \code{sklearn}'s \code{CountVectorizer}. The vector is reduced from \char`~4,000 to 150 dimensions for speed (an optional step as the training is already fast, on the order of a few minutes) and fed to a logistic regression classifier. We then train and evaluate the classifier with an 80/20 train/test split (5-fold cross validation).%
   \normalcolor
    },l=fig:pipeline%
    }{pipeline_logreg_bow.png}%
\end{savedenv}

\begin{savedenv}[fig_screenshot_places]
    \medskip
    \keyfig[H]{lw=0.4,ft,c={%
      Screenshot of a review found on Google Maps API. Some features used in later analysis are boxed in red.},
      l=fig:screenshot_places%
      }{screenshot_places.png}%
\end{savedenv}

\begin{savedenv}[fig_pipeline_ablate_bow_logreg]
    \medskip
    \keyfig[H]{lw=1,ft,c={%
      \notered{Pipeline of how text was processed for NORP and LANGUAGE tokens, which were removed prior to running through the pipeline in Part 1 of Stem, bag-of-words, and logistic regression pipeline.}},
      l=fig:pipeline_ablate_bow_logreg%
      }{pipeline_nerp_ablate_bow_logreg.png}%
\end{savedenv}

\begin{savedenv}[fig_pipeline_hours_vector]
    \medskip
    \keyfig[H]{lw=1,ft,c={%
      Diagram of pipeline for the hours-based logistic regression, which has two features (in essence a decision tree).},
      l=fig:pipeline_hours_vector%
      }{pipeline_hours_vector.png}%
\end{savedenv}

\begin{savedenv}[confusion_hours_features_logreg]
    \medskip
    \keyfig[H]{lw=0.5,ft,c={%
      Confusion matrix of the hours-based logistic regression (open late and open 7 days a week). The model struggles more with false positives than the text-based model.},
      l=fig:confusion_hours_features_logreg%
      }{CONFUSION_HOURS_logistic.png}%
\end{savedenv}

\section{Introduction}

The erotic massage industry provides not just massage and spa services but also commercial sex. In the United States, these businesses, like other forms of sex work, are illegal (barring a few places in Nevada), hence the designation as the illicit massage industry (IMI). The IMI is far from a niche industry and includes thousands of businesses found in all fifty states. \notered{Our dataset lists over a hundred locations in Manhattan alone that received visits in the first half of 2023.}\footnote{\notered{Specifically, the authors found that as of April 2023, the website listed 180 businesses with user-submitted reviews since Jan 1st, 2023, marked as erotic and not permanently closed.}} The non-profit Polaris estimated that in 2018, the combined annual revenue of these businesses in the United States was in the range of \$2.5 \textit{billion} dollars \cite{polaris-social}.

Due to increasing law enforcement and public scrutiny, as well as the potential for human trafficking of workers, a better understanding of the IMI is urgently needed.  According to interviews conducted by \cite{chin2019}, most employees are middle-aged married immigrant women, often with limited English, facing economic hardship and a lack of job opportunities. Many employees work in exploitative conditions ranging from simple labor law violations (e.g. lack of paid holidays) to complex psychological and physical forms of labor and sex trafficking. However, it should be re-emphasized that not all IMI businesses are venues for trafficking.

Rubmaps is a publicly accessible website specific to the illicit massage industry (IMI) which lists businesses along with the services users have gotten at any given location. These provide an important source of data for research into the IMI. A screenshot is shown in \cref{fig:rmaps_homepage}.

\printsaved{fig_rmaps_homepage}

\color{black}
However, Rubmaps and similar websites dedicated to commercial sex exist in a precarious situation where changing laws and policies can lead to law enforcement action shutting down the website(s), as happened in 2016 to Backpages.com \cite{BackpageChildsafeAi}. Thus, there is inherent value in having alternative data sources about the IMI. Since data about illegal activity is by nature very noisy, more data sources allow a clearer estimate of the true signal. %

\begin{mdframed}
\noindent\textbf{Content Warning:} This work may contain sexist and racist language.
Reader discretion is advised.
\end{mdframed}

\section{Our Work}

\textbf{Part 1}. We thus seek to create a list of IMI businesses that does not rely on Rubmaps. To do so, we find a list of spas and massage parlors using the Google Maps Places (henceforth referred to as "Places") API\footnote{Application Programming Interface} and build a model to classify those as illicit (1 label) or legal (0 label). We treat the Rubmaps listings as ground truth, where businesses listed in Rubmaps are illicit, and those not listed in Rubmaps are legal establishments (this is a strong assumption we will address in the discussion in \cref{sec:discussion}). Using only the Places review text, we use a count vectorizer (a.k.a. bag-of-words) with a logistic regression model and achieve nearly 80\% accuracy when compared to the original Rubmaps listings. This compares favorably to the null accuracy baseline of 62\% (if the classifier always outputs the majority class). We also consider using a transformers-based model, specifically DistilBERT \cite{sanh2019distilbert}, as the classifier. We found the transformer achieved similar accuracy but required much longer training and inference time.

Our work can be viewed as having two main contributions in this part:
\begin{enumerate}
    \item  As mentioned before, if Rubmaps shuts down, we can use our model to continue monitoring the United States IMI so long as Google Places continues to exist.
    \item Additionally, Rubmaps provides (publicly) only basic records such as business name and location.\footnote{Although Rubmaps does include (very explicit) user-submitted reviews, this requires paying the website operators as subscription fee.} By using Places data, we can also retrieve user reviews for each business \textit{without} paying Rubmaps.
\end{enumerate}

\textbf{Part 2}. In the second (and brief) part of this chapter, we consider a few domain expert claims about the IMI and whether this is reflected in the data. Specifically, we do find that reviews for illicit businesses are more likely to mention ethnicity or language and also cash. We also find that there are strong distributional differences between the opening hours of IMI businesses vs. regular massage  businesses. See \cref{fig:screenshot_places} for an example review that inspired these feature definitions.

\printsaved{fig_screenshot_places}

\textbf{Part 3}. In the third part of this chapter, we consider two strategies to mitigate potential bias in the dataset. First, we can avoid using text features at all and instead use more generic features. To this end, we trained a logistic regression classifier on hand-crafted hours features (specifically, binary features indicating 1. if a business is open seven days a week and 2. if a business is open later than 9 P.M.) and found it achieves 70\% or so accuracy. Second, we consider the possibility of removing sensitive text features, specifically text mentioning language or ethnicity from the dataset. Upon re-running the bag-of-words (BoW) and logistic regression pipeline, we find this ablation does not significantly impact the accuracy compared to using the full text corpus. 

The ultimate impact of our classifier lies in downstream analysis. From the law enforcement perspective, having a stable source of data is important for criminal justice proceedings. There is also no list of all spas and massage parlors, and although ours is certainly not comprehensive (the API returns only the twenty locations closest to the city center's GPS point), the rough ratio of illicit to legal businesses could still help direct resources to highly impacted communities. %
From the economics research perspective, the classifier opens the potential to study the IMI outside of the United States. Rubmaps only covers the United States, while Google Maps covers many countries around the world. This all the more important given that the IMI is a global industry.

\normalcolor

\section{Background}
\label{sec:imi_bkgd}

The following abbreviations are used. 

\begin{mdframed}
\noindent 
\notegreen{
\textbf{IMI} Illicit Massage Industry. \textbf{IMB} Illicit Massage Business. \textbf{RM} Rubmaps. \textbf{Places} The Google Maps Places API.
}
\end{mdframed}

\subsection{Trafficking within the Illicit Massage Industry}

We provide contextualization of the illicit massage industry in the United States, define human trafficking, and explain how the two are interlinked.

\subsubsection{The Illicit Massage Industry (in the United States)}
\label{sec:imi_stats}

We follow the definition of illicit massage business used in \cite{boucheEstimatingDemandIllicit2018}:

\begin{displayquote}
\textbf{Definition}. Illicit massage businesses are establishments with registered business names that ostensibly provide massage, wellness, and/or spa services while in fact deriving some clientele and revenue through the provision of commercial sex acts. \\
\end{displayquote}

The IMI is a large industry: in 2018, the IMI included over 11,000
businesses in the United States, according to Polaris \cite{polaris-social}, a
non-profit that works to end human trafficking in the US. \notegreen{As mentioned previously}, Polaris estimated a combined annual revenue for IMBs of \$2.5 billion in 2018. 

These businesses operate in plain sight and can be found throughout the United States. At the city and state levels, researchers have combined website data and in-person foot traffic estimate annual gross revenue. These estimates include over \$100 million in Houston \cite{boucheEstimatingDemandIllicit2018},  \$180+ million in Dallas \cite{DallasweismannmelissaEstimatingDemandGross2019},  and \$40+ million in Georgia \cite{rodgersbobStreetGraceIllicit2020}. Demographically, the most common demographics of employees are women in their thirties to fifties, who have children, arrived recently in from China or Korea, and speak very little English. The IMI has come to prominence over time with the scandal involving billionaire Robert Kraft \cite{narratives} as well as the 2020 mass shooting targeting employees at Asian massage parlors in Atlanta \cite{lam_double-edged_2021}.

\subsubsection{Historical Perspective on the IMI}

From a historical perspective, the prevalence and market dynamics of the IMI in the United States can be traced back to the official comfort stations (brothels) of the Japanese Imperial army \notegreen{and the patriarchal and nationalistic neo-Confucian ideals before that. The subsequent American military invasions in East Asia led to a hypersexualized depiction of Asian women, which was reinforced by popular media in the 21st century. This drives the modern-day demand behind the IMI.} For a comprehensive account of these historical, societal, and economic factors, please see \cite{bookHistoricalModernDayComfortStations2021} or the video at \cite{videothenetworkHumanTraffickingIllicit2022}. For a more personal account of the current IMI, please see the newspaper articles \cite{NYTimesCaseJaneDoe2018} and \cite{kulishNYTimesIllicitMassageCrimeNetwork2019} or the video at \cite{videothenetworkThisMyStory2023}. As they are a bit depressing we leave further discussion out of this thesis. 

\subsection{The Illicit Massage Industry in the United States}

 \subsubsection{Trafficking}
\label{sec:trafik}
Human trafficking is defined as follows in the United Nations Palermo Protocols, which was adopted in 2000 and now ratified by 178 parties \cite{UNProtocolPreventSuppress2000}:

\begin{displayquote}
\textbf{Definition}. "Trafficking in persons" shall mean the recruitment, transportation, transfer, harboring or receipt of persons, by means of the threat or use of force or other forms of coercion, of abduction, of fraud, of deception, of the abuse of power or of a position of vulnerability or of the giving or receiving of payments or benefits to achieve the consent of a person having control over another person, for the purpose of exploitation.
\end{displayquote}
 
Notably, trafficking does not require moving someone or something across country borders, a common misconception.

\subsubsection{Trafficking in the Illicit Massage Industry}

Trafficking is known to occur in the IMI. Indeed, several of the 116 women  
interviewed by \cite{chin2019} identified as victims of trafficking (while others described very coercive circumstances that easily met the definition, they may not have identified as victims for multiple reasons, such as confusion about how trafficking is defined, or a belief in their own autonomy).\footnote{For the study, women were
found through ads in ethnic newspapers and financially compensated for the
work-hours missed.}

For instance, employers may advertise in newspapers for massage work without mentioning sexual work (fraud) or confiscate passports (coercion). Women may take out loans to travel to the United States. Immigration agencies or persons may promise a job and a place to live, but set a system where the rent and other fees are such that it is almost impossible to leave the cycle of debt. As reported in \cite{chinjohnj.IllicitMassageParlors2019} and \cite{keyhanrochellePolarisHumanTrafficking2018}, traffickers use language barriers, immigration status, and cultural values to enforce the idea that the women involved have no one else to turn to, i.e. that there are no civil or legal protections and that they may be deported if they seek help, and that society will hold them accountable as prostitutes and victims of exploitation and trafficking.

In the same study, researchers found that psychological coercion (e.g. threats of law enforcement and deportation) are more common than physical coercion.  Without obvious physical coercion, in practice available evidence may not allow for a clear separation between trafficking and non-trafficking conditions. \cite{chin2019} found that from their perspective, many of the women chose IMB work as the best of a limited set of options (by injuries, allergies to nail salon chemicals, poor English, etc.). Additionally, severity can vary over time: workers could initially be trafficked into the industry and later choose to stay e.g. if unable to find alternative employment.\footnote{Choosing a trafficking narratives can backfire if the intent is to help IMB employees. In the United States, if support for IMB workers is conditioned on their being trafficking victims, but trafficking cannot be proved, then instead the same workers will be charged with prostitution. In the aforementioned Robert Kraft case in 2019, Kraft's charges were dropped, whereas the IMB employees ended up with criminal records and thousands of dollars in fines \cite{narratives}.}

Law enforcement and more victim-centered organizations, such as health outreach workers, both agree that IMI employees work in poor labor conditions and are a marginalized community deserving support. Support services could include expanding sexual healthcare resources as well as job skills training to find alternative sources of income.

\subsubsection{Rubmaps.ch (RM)}

\begin{displayquote}
   \textbf{Acronym}. RM refers to Rubmaps.ch. \\
   \textbf{Definition}. RM is a website which lists user-submitted massage parlors. Listings contain information ranging from standard business features (e.g. opening hours, address) to IMI specific information (e.g. whether illicit services are offered).\footnote{Behind a paywall, more information is available, such as how to get illicit services and graphic details of the specific women and services available.} RM also has a forum where thread titles are open to the public but post text requires a (free) account.
\end{displayquote}

Our domain-specific data source is Rubmaps, a site specifically for IMBs.\footnote{Rubmaps title text includes "Erotic massage parlor reviews - find your Happy Ending!" The Rubmaps homepage is shown in \Cref{fig:rmaps_homepage}.} Although there are several sites with user-submitted IMB reviews, we chose to use Rubmaps due to having access to a pre-cleaned dataset courtesy of The Network.\footnote{www.thenetworkteam.org}

\subsection{Computer Science and Counter-Trafficking}

\notegreen{ A sample of the ways in which computer scientists are grappling with counter-trafficking follows. }

\color{black}
\subsubsection{Related Work}
In terms of counter-trafficking, many computer science papers in the last decade focused on the Darknet, including creating ontologies and knowledge graphs for the DARPA MEMEX program \cite{Kejriwal2017AnIS}. Researchers also developed classification algorithms on top of data scraped from Backpages, a website that included adult online classifieds, when it was still online. For example, \cite{Tong2017CombatingHT} developed both the multimodal Human Trafficking Deep Network (HTDN) as well as an annotated dataset of 10,000 posts along with images from escort ads on Backpages. Two works have investigated the use of emoji using Backpages as well: \cite{emojiWhitney2018DontWT} and \cite{NNemojiWang2019SexTD}. The latter includes using t-SNE and word embeddings to handle the constantly changing jargon used by posters to evade automated activity. The advent of
social media presented additional public online sources of data about demand and supply of commercial sex. \cite{simonsonThesisSemiSupervisedClassification} investigated semi-supervised classification of posts on Twitter based on post text and hashtag. \notegreen{Other researchers investigate offline sources of data.} For instance, \cite{Helderop2018HiddenIP} combines prostitution arrest data from the Houston police department, review text from Travelocity (vectorized with FastText \cite{fasttextbojanowski2017enriching}, %
hotels do seem to have higher surrounding rates of prostitution. 
In the field of computer vision, works include the Hotels-50k \cite{stylianouHotels50KGlobalHotel2019}. In that work the authors sought to classify images (with large censored blocks) as belonging to a particular location of a hotel chain. Many rooms within the same hotel are different, while rooms in distant locations may be similar if they belong to the same chain, leading to a challenging research problem. %
\normalcolor

\subsection{Computer Science and the IMI} 

There are two prior works that combine two data sources (Rubmaps.ch and Yelp) and use natural language processing (NLP) to create a classifier to distinguish illicit from legal activity. %

\cite{diaz_natural_2020} present the first attempt to use open web massage data
(i.e. Rubmaps public listings which do not require log-in nor payment) combined
with Yelp reviews to estimate whether a given massage parlor is illicit or not.
They use a term frequency-inverse document (TF-IDF)
weighting of a bag-of-words model and consider several possible classifiers including ensemble models, achieving an accuracy of 76\%. Their dataset includes 11 states, while in our work we cover all fifty states. %
In \cite{li_detecting_2021} the authors focused on classifying individual
reviews. They combined Rubmaps locations with Yelp reviews and ran
classification for five states in the United States. By individually labeling
several hundred Yelp reviews, they also evaluate the overlap between IMBs on
Yelp with IMBs on Rubmaps and show the limitations of blanket use of listings in
Rubmaps as the sole source of ground-truth labels. 

We consider more (academic) inquiries, such as the impact of removing sensitive text from our classifier as well as the previously ignored opening hours features.
\notegreen{
\cite{devriesIdentifyingRiskMarkersRmapsText2022} look at RM data alone, which includes review text (located behind a paywall). They approach it from the linguistic and criminology perspective, and look at the distribution of risk categories words (e.g. the risk markers "new girls" or the clearer flag "assault") in the RM review text (a dataset that we do not use and that is behind a paywall). They find the distribution is not random -- the presence of a clear flag (such as violence) means other risk markers are more likely ("new girls") to be mentioned. They emphasize the lack of causality and simply state that inter-relatedness of the risk markers and the clear flags indicates that the risk markers are informative for the presence of trafficking. They include a detailed discussion of the caveats in using online review text from clienteles for analysis.
}

Finally, in \cite{white_why_2021} researchers used the locations listed in Rubmaps.ch to evaluate what demographic factors influence where IMBs are located. 
For input data, they use public datasets about socioeconomic, population
density, etc. as input features into a random forest classifier. The feature weights then allow inference about what factors do and do not have significant predictive power to whether a specific region will be (relatively) high, medium, or low in Rubmaps listings. They find among other features, international airport distance, state, rent and income levels, racial composition, and religious presence were all significant.

Our work focuses on what conclusions can be drawn from natural language processing techniques and does not seek to evaluate the spatial distribution of IMBs.

\section{Methods and Results} 

The following abbreviations are used. 

\begin{mdframed}
\noindent 
\textbf{BoW} Bag-of-Words. \textbf{MCC} Matthews Correlation Coefficient.
\end{mdframed}

\printsaved{fig_pipeline}

We use Places data to create supervised models that classify individual businesses as more or less likely to have illicit activity.

We discuss the dataset creation process, followed by the classification
specifics for the two research aims. For clarity, we interleave the direct
results directly after the methods of each experiment.

\subsection{Data Labels}

Our labels are \code{1="listed in RM"} and \code{0="not listed in RM."} We treat these in the following text as "illicit" and "licit" for clarity of speech. A discussion of the validity of this assumption is found in the \Cref{sec:discussion}.

\subsection{Raw Dataset Creation}

Our dataset consists of \code{Place IDs} and \code{Place Details} from the Google Maps API. The former are unique IDs for each business, while the latter is a more expensive call that includes both text reviews (up to five per business are returned) and opening hours.

\subsubsection{Place Details from Google Maps API}

For each location we issued an API request for \code{Place Details}. The returned
details include between zero to five reviews per Place ID, as well as the
business opening hours.

To create our positive class (\char`~5,000 businesses), we use a dataset (provided by a third party) of Places IDs that are linked to businesses in Rubmaps.\footnote{Data courtesy of The Network. The dataset was described as: scraping Rubmaps.ch for businesses labeled with by users as \code{is_erotic} on the website, then linking using the business name and location to link to businesses listed in Places.} To create our negative class (\char`~7,500 businesses), we then query Places for additional spas and massage parlors (up to twenty per city, with the list of cities derived from RM) and make a set of those \textit{not} listed in RM. 

\color{black}
\textbf{Overview of positive data pipeline:}
\begin{enumerate}[topsep=0pt,itemsep=0pt,parsep=-1ex]%
\item Start with PlaceIDs mapped to locations in Rubmaps (this is provided to us by The Network)
\item Filter to locations with reviews as of January 1st, 2019 or later
\item Retrieve reviews using Places API to get positive set
\end{enumerate}

\textbf{Overview of negative data pipeline:}
\begin{enumerate}[topsep=0pt,itemsep=0pt,parsep=-1ex]
\item Extract city names from above list
\item For each city, issues "Nearby" request to get up to 20 additional spas and massage parlors\footnote{More requests are possible, but would have further imbalanced our dataset} returned as PlaceIDs
\item Remove PlaceIDs found in positive set
\item Retrieve reviews using Places API to get negative set
\end{enumerate}

\medskip
\noindent In the next paragraphs, we describe the steps in more detail.
\normalcolor 

\subsubsection{Positive Class}

Our Rubmaps data is provided by Heyrick Research (now known as The Network). The fields used include business name and street address, as well as a matched Google Maps Place ID.\footnote{Similarly, note that \cite{li_detecting_2021} independently scraped RM and mapped RM locations to Yelp locations. \notegreen{Our dataset should be replicable by scraping the RM public listings, removing businesses marked not \code{is_erotic} on the site, and matching to businesses in Places using business name and address. Specifically, one can use the \code{FindPlaceFromText} Google Maps API call.}}%

In the RM dataset, since IMBs will have
relatively high turnover (as they change locations in response to e.g. law enforcement), we remove locations that only have reviews before January 1st, 2019. This left approximately 6,000 locations
across the fifty states and DC, each with a matched Places ID.
After removing duplicates, the result is a set of 4,719 locations in our positive class, for which we retrieved Place Details.

\subsubsection{Negative Class}

For our negative class, we compiled a list of all cities present in the filtered RM dataset, 1,745 cities in total. For each city, we used the Places  \code{Geocoding} service to get a latitude and longitude, then made a Places  \code{Nearby Search} request for that GPS point. This call returns up to 20 places ordered by distance from the bias point. In addition to city's latitude and longitude, we also added the keyword "massage" and that results be restricted to type "spa" (one of 96 possible types supported by Places API).\footnote{\url{https://developers.google.com/maps/documentation/places/web-service/supported_types}} 

In total, this created a list of 17,247 Place IDs. Of these, 1,541 Place IDs overlapped with the positive set and were removed, leaving 15,706 Place IDs in the negative set. 
Due to budget constraints, we randomly sample half the negative set.\footnote{Educated conjecture as code is lost to time. Supporting details: last character of Place IDs are similarly distributed between the listed zeros and the downloaded zeros. The state and the last character of the Place IDs is similarly distributed between the downloaded zeros and the downloaded ones. Here "downloaded" means "issued request for Place Details."} After removing duplicate Place IDs and one location in Ontario, we made \code{Place Details} API calls.
\normalcolor %

\subsubsection{Cleaning}

Finally, we remove Place IDs returned as obsolete by the Place Details API and also locations with no reviews.
The result is 12,150 locations, split into 4,719 locations in our positive (illicit) class and 7,431 locations in our negative (legal) class.

\subsubsection{Additional Notes} 

On a per-review basis, we have 20,064 reviews in our positive class,
35,311 reviews in our negative class, for a total of 55,385 reviews. \notered{In total, we made \char`~1,800 \code{Places Nearby Search} requests (one for each city), \char`~1,800 \code{Geocoding} requests (to turn city names into GPS coordinates), and \char`~10,500 \code{Places Details} requests (one per Place ID in addition to those with details already reported in the Nearby calls). The total cost of the API calls was approximately \$350.}

\subsection{Part 1: Review Text Classifier}

\subsection{Part 1A: Bag-of-Words and Logistic Regression}

For an overview, refer to \Cref{fig:pipeline} which describes
the pipeline at a high level.

\subsubsection*{Text Preprocessing with NLTK}

For each location, we concatenated all reviews (the API returns up to five per business) into a single string.  We then used the NLTK library \cite{nltkbird2009natural} to pre-process the text.
We removed all non-alphabetical characters and converted to lowercase. 
We then used the NLTK implementation of the Porter stemming
algorithm %
to turn words into their stem.\footnote{For instance, "massage",
"massages", and "massaged" all become "massag."} We did not remove stopwords.
\color{black}
Below we provide an example of the pre-processing for a single place that has three reviews.

\begin{enumerate}[topsep=0pt,itemsep=-1ex,partopsep=1ex,parsep=1ex]
    \item Review 1: "Happy we went there." \\ Revew 2: "The place smelled." \\ Review 3: "Spoke English there."
    \item Concatenate: "Happy we went there The place smelled Spoke English there"
    \item Stem: "happi we went there the place smell spoke english there"
\end{enumerate}
\normalcolor

\subsubsection*{Vectorizing}

We then used the bag-of-words (BoW) representation to turn each (concatenated) review text into a vector of numbers,
or specifically sklearn's \cite{scikit} CountVectorizer function using unigrams
(to reserve computation).\footnote{We chose not to use the (TF-IDF)  weighting, which also creates a list of words but as it reduces the weights terms that appear rarely in a corpus. Our intuition was that slang words that may not repeat often but can be very helpful for determining whether there are illicit services on offer.}

BoW creates a list of words across the whole text input, then for each (concatenated) review records the number of times each word is used, and for our dataset, resulting in a bag of \char`~4,000 words (depending on how the dataset was randomly split into test and train sets). Note that we specifically only used the training set to fit our BoW vectorizer for the cross-validation results.

\subsubsection*{Dimensionality Reduction}

We used the singular value decomposition (SVD) algorithm, specifically TruncatedSVD, to reduce the dimensions of the feature set. After combining with the classifier in the next section, we ran a cross-validated grid search of the number of features to reduce to and chose 150. This allowed for a faster run time.

\subsubsection*{Binary Classification}

We tried a few common machine learning models and found performance acceptable for all models and performance differences relatively minor relative to the noise of the dataset. Given that other papers have focused on improving classification accuracy, we chose without further ado a Logistic Regression model to simplify analysis. After a grid search of the regression regularization parameter C, we chose C=10, and again to regularize the data further we chose to use a \code{l1} penalty. Finally we chose the \code{liblinear} solver as our dataset is relatively small.

\subsubsection{Results}

\printsaved{fig_bow}

Results are shown under "Bag-of-Words (n$\approx$150)" in
\Cref{tbl:logreg}. The confusion matrix is in \Cref{fig:bow}.  The 80\% accuracy is remarkable given the noisiness of the dataset, and is sufficient to warrant further analysis. We also compare against the baseline null accuracy of predicting only the most common class. This is a baseline of 61.4\% accuracy, so the logistic regression compares favorably.

\notered{We include the MCC metric, or Matthews Correlation Coefficient. A good MCC score is well-regarded as it requires good performance on both classes in binary classification. The MCC takes values between -1 and 1. A score of -1 (or 1) indicates perfect negative (or positive) correlation.  A score of 0 indicates random chance.}

\addtolength{\tabcolsep}{3pt}
\begin{table}[]
\caption{\notered{Metrics for 5-fold cross validation of the hand-crafted features. $n$ is the number of features input to the classifier.}}
\label{tbl:logreg}
\centering
\begin{tabular}{lll}
\toprule
        & Bag-of-Words         & Opening Hours \\ 
       & (n=150)           & (n=2)                      \\ \midrule
Accuracy   & 0.794 ($\pm$ 0.038)         & 0.700 ($\pm$ 0.043)   \\
Precision  & 0.718 ($\pm$ 0.057)         & 0.595 ($\pm$ 0.051)   \\
Recall & 0.797 ($\pm$ 0.004)         & 0.721 ($\pm$ 0.010)   \\
F1     & 0.754 ($\pm$ 0.033)         & 0.651 ($\pm$ 0.034)   \\
MCC    & 0.586 ($\pm$ 0.062)         & 0.399 ($\pm$ 0.072)   \\ \bottomrule
\end{tabular}
\end{table}
\addtolength{\tabcolsep}{-3pt}

\color{black}
\subsection{Part 1B: DistilBERT}

We also briefly compare against using a transformers model. Specifically, we chose to use a spatial transformers model, DistilBERT, as part of a classification pipeline. We used the SimpleTransformers \cite{simple} library. In terms of parameter choices, we used a max sequence length of 128 tokens. Since review text could be longer than 128 tokens, the library uses a sliding window function, which we chose to have a 20\% overlap between windows. In this case, the overall prediction for a business is the mode of the window predictions (defaulting to 1 for ties). We trained for 10 epochs and otherwise accepted the default parameters.\footnote{Default parameters can be found at: \url{https://simpletransformers.ai/docs/usage/\#configuring-a-simple-transformers-model}} A few key parameters include: a learning rate of 4e-5 after warm-up, batch size of 8, and Adam optimization epsilon of 1e-8. We also used a seed of 42 for reproducible results (due to length of training we ran only one seed). The transformer output is run through a classification layer with two output neurons, and the prediction is the softmax of two outputs.\footnote{Refer to the source code at \url{https://github.com/ThilinaRajapakse/simpletransformers/blob/master/simpletransformers/classification/transformer_models/distilbert_model.py\#L19}}

\textbf{Results}. We found similar accuracy of about 80\%, which we report in \Cref{tbl:distilbert}.

\textbf{Discussion}. The DistilBERT results are within the standard-deviation estimates of the BoW model. A few hypotheses could explain the comparable results. First, it may be that beyond a certain point, the review text simply does not provide enough information to distinguish places. A comparison to manual classification (by humans) could reveal whether the 80\% accuracy could be a ceiling. If the same businesses fail classification between the BoW and BERT models, that would strongly support this hypothesis as well. The second is the potential for overfitting. Both models may be over-fitting, as the average length of the concatenated reviews for a business falls well under 4,000 characters, but the models both have many more features (approx. 4,000 and 30,000 respectively). In this case, the stability of the DistilBERT model across multiple runs could be meaningful: if the DistilBERT results are very unstable, the likely culprit is overfitting.

\begin{table}[]
\caption{DistilBERT classification results. }
\label{tbl:distilbert}
\centering
\begin{tabular}{@{}llllll@{}}
\toprule
\textit{} & Acc.  & Prec. & Recall & F1    & MCC   \\ \midrule
Value     & 0.789 & 0.729 & 0.725  & 0.727 & 0.555 \\ \bottomrule
\end{tabular}
\end{table}

\subsection{Part 2: Feature Distributions by Class}

In this part, we consider the features a domain expert might define, and whether they vary between the two classes (illicit and legal).  Specifically, we are interested in the aforementioned features: 

\begin{enumerate}[itemsep=0pt]
    \item MONEY - Whether the review text mention money (that is, illicit places may be more likely to only take cash)
    \item ETHNICITY - Whether the reviews mention language or nationality (that is, illicit places may be more likely to employ those with limited English skills or recent immigrants) 
    \item OPEN LATE - Whether businesses is open until late 
    \item OPEN 7 DAYS - Whether businesses is open seven days a week 
\end{enumerate}

The creation of the latter two features is straightforward. We describe as follows how we implemented the first two features. 

\subsubsection{Pre-processing}

As before, we concatenate the review text, as well as remove non-alphabetic characters and lowercase. However, we do not stem the text. 

\subsubsection{Named Entity Recognition}

We then run our text through SpaCy's named entity recognizer. A named entity recognizer attempts to tag words as entities, for "Apple" would be considered an organization and "April" would be considered a date.
SpaCy provides off-the-shelf language models, and we use \code{en-web-core-lg}).  In contrast to the most recent NER models in SpaCy (e.g. \code{en_core_web_trf}) which are transformers based, this model uses a deep neural network architecture similar to a convolutional neural network with a few custom additions (including the concept of attention).  Specifically, it uses a transition-based approach and a framework called "Embed. Encode. Attend. Predict." At a high level, dense embeddings are learned (word vectors that represent how likely words are to be next to each other). These vectors are encoded into sentence matrices. These are fed into an attention model which determines what parts of sentences to attend to with respect to a query.  Finally, the model outputs a prediction, in this case our named entity. For NER, the SpaCy pipeline is trained on OntoNotes 5.0 \cite{hovy2006ontonotes}, a large corpus released in 2013 of many genres of text in three languages with both structural information and shallow semantics (word sense as part of an ontology).

For our purposes, we consider three SpaCy entities. \code{MONEY} indicates "monetary values, including units," \code{NORP} indicates "\underline{n}ationalities \underline{o}r \underline{r}eligious or \underline{p}olitical groups," and finally \code{LANGUAGE} indicates "any named language."

\subsubsection{Hours Distribution}

For each hour of each weekday starting at 1 PM and going until midnight, we tallied the number of businesses that closed at that time.

For opening hours, the distribution differences between the 0 and 1 classes can be seen in \Cref{fig:hours}. The strong difference in trends between the zero and one label cases was reassuring.

\printsaved{fig_hours}

\normalcolor

\subsubsection{Results} 

\Cref{fig:barplot} shows the resulting distribution between 0 and 1 labels for the four features.

\printsaved{fig_barplot}

We see that, as expected, reviews for IMBs are more likely to mention money, ethnicity, and nationality. However, the overall percentage of businesses with these tokens anywhere in their reviews is low. On the other hand, the hours features (open late and open 7 days) appear to be a much stronger signal.

\color{black}
\subsection{Part 3: Removing Sensitive Features}

\subsubsection{Part 3A: Logistic Regression Using Hours}

We then consider the two features created previously, and apply them as inputs to a logistic regression classifier (in essence creating a decision tree). This pipeline is shown in \cref{fig:pipeline_hours_vector}.

\printsaved{fig_pipeline_hours_vector}

We already see a performance achieving over 70\% accuracy using just two features instead of 150.
This may be telling that we can further reduce bias reinforcement by using less
of the text corpus. Refer to \Cref{tbl:logreg}.

\printsaved{confusion_hours_features_logreg}

The confusion matrix is shown in \Cref{fig:confusion_hours_features_logreg}.
\normalcolor

\subsubsection{Part 3B: B.o.W. without Sensitive Features (Ablation)}

\printsaved{fig_pipeline_ablate_bow_logreg}

We look to see if a business has any reviews that get flagged as "NORP" and "LANGUAGE" tokens and remove those words. For each entity SpaCy provides start and end indices into the document, which we use to remove the tagged text. An overview of the pipeline is shown in \Cref{fig:pipeline_ablate_bow_logreg}.

\textbf{Discussion}. From the ablation, we see that the removal of potentially sensitive features
did not have a strong impact on the model. As a preliminary measure, such
features could be removed from the dataset to avoid directly reinforcing existing biases. Further investigation is warranted as the algorithm may be picking up still on other strongly correlated features.

\addtolength{\tabcolsep}{3pt}
    \begin{table}[htbp]
    
            \caption{Metrics for the full text combined with the ablation experiments. Null accuracy (not shown) is 61.4\%. Overall, the ablations appear to have minimal effect on the scores of the logistic regression classifier.}
            \label{tab:ablations}
            \centering
            \begin{tabular}{lll}
            \toprule
                        & Original &  After Ablation   \\ \midrule
            Accuracy    & 0.794 ($\pm$ 0.038)   & 0.794 ($\pm$ 0.038)  \\
            Precision   & 0.718 ($\pm$ 0.057)   & 0.718 ($\pm$ 0.057)  \\
            Recall      & 0.797 ($\pm$ 0.004)   & 0.797 ($\pm$ 0.004)  \\
            F1          & 0.754 ($\pm$ 0.033)   & 0.754 ($\pm$ 0.033)  \\
            MCC         & 0.586 ($\pm$ 0.062)   & 0.582 ($\pm$ 0.065)  \\ \bottomrule
        \end{tabular}
        
    \normalcolor
    \end{table}
\addtolength{\tabcolsep}{-3pt}

\section{Discussion}
\label{sec:discussion}

\textbf{Ground Truth Labels}. For our ground truth labels the assumption we make is that
all places listed in RM are IMBs, and if a place is not listed in RM then 
it is a legal massage parlor. Both \cite{white_why_2021} and \cite{diaz_natural_2020}
make this assumption also, but a quick exploration of the RM review text (which
is de-identified and does not contain usernames) shows users commenting that they
were \textbf{not} able to buy sex (of various kinds) at specific locations.
Unfortunately the RM text is not only extremely graphic but also disturbing such
as trading tips on how to pressure for sexual services, and so far all
researchers have avoided working with the RM text directly. For evaluating whether one can find illicit activity using information on websites like Places, the ground truth labels are sufficient despite the noise. However, as a consequence of these assumptions our classifier should never be used on its own to definitively classify a business as illicit or not.

\color{black}
Additionally, \cite{diaz_natural_2020} removed certain keywords (e.g. "hair") from their listings as likely to be hair salons instead of massage parlors. We chose not to do so, as a brief manual inspection of our dataset revealed many locations that offered haircuts indeed also offered massage services.\footnote{A random sample found less than 1\% of listings fell in the category of offering hair services exclusive to massage parlors.} If the focus is only on accuracy and other metrics, this would be a point of concern, but if the goal is understanding the spatiotemporal trends of the IMI, a relative-to-baseline measure is sufficient. Indeed, for classifying any individual business, the limited accuracy of our classifier (at \char`~~80\%) is mediated by the fact there would always be manual gathering of additional corroborating evidence.
\normalcolor

\notegreen{
\textbf{Accuracy}. The fact that the more sophisticated deep learning classifier (DistilBERT) did not achieve improved performance and could only reach similar accuracy as the BoW classifier is interesting. This likely hints at a fundamental noise ceiling in our dataset, in which case the addition of more nonlinearity in our classifier did not. Indeed, preliminary investigations of combining the BoW with an ensemble method, the random classifier, yielded slightly worse but fairly similar accuracy as the logistic regression classifier.
}

\subsection{Conclusion and Future Work}

\notered{
Our experiments clearly indicate that there is enough signal present in just the
Places reviews to give some indication (albeit with low confidence) as to
whether customers encountered illicit activity or not. This already has the
impact of allowing researchers a broader dataset, e.g. RubMaps is primarily used
in the United States but services such as Google Places are used around the
world. Additionally, our work shows that features such as business opening hours served as strong signals worthy of future research. We hope that other researchers will be inspired by this work to investigate the applications of computer science to counter-trafficking within the illicit massage industry. 
}

For improving the classifier, a human baseline would be a straightforward way to resolve the question of why higher accuracy was not achieved with the transformers model. This would verify the hypothesis that the remaining 20\% or so of businesses could not be classified on review text alone. \notegreen{Additionally, one could use the probability estimates from the logistic regression classifier (as opposed to just the binary output) to sample which reviews the classifier was most confidently incorrect on. Both a human baseline and looking at probability distributions of the existing classifier could help assess the validity of our ground truth label assumption.}

Engaging the wider academic community on this topic would also help continue our work. Taking cues from the public availability of the Hotels-50k
\cite{stylianouHotels50KGlobalHotel2019} dataset and its presence now as a
benchmark dataset for the European Conference on Computer Vision (ECCV), this could be as simple as creating datasets with starter code for exploratory analyses. \notegreen{From near- to long-term, actionable tasks to engage the computer science community} would include: 1.
create public datasets on this topic 2. curate quick start documentation 3.
create competitions using the datasets 4. create prizes and funding for this
topic (work ongoing within the AI4SG (Artificial Intelligence for Social Good) and CSS (Computational Social Sciences) communities) 5. create clear use
cases and impact metrics to drive research 6. create institutional support at
the university and funding agency level to support research into human trafficking and other budding niche \notegreen{yet potentially societally-impactful topics}.

\notegreen{In terms of impact on the ground, future work would require engaging interdisciplinary researchers and community stakeholders. Creating and hosting a user-interface for exploring the dataset would help engage others. Working with stakeholders such as non-profits distributing healthcare and other resources would be the next steps to translate this research into actual social impact in improving the lives of those entangled in the IMI}.

\captionsetup{labelfont=bf}

\graphicspath{{./figs/datasets/}}

\newenvironment{img}[4][0.5\textwidth] 
    {\begin{wrapfigure}{r}{#1}
    \centering
        \fcolorbox{gray}{white}{\includegraphics[width=#1]{#2}}
      \caption{#3}
      \label{#4}
    \end{wrapfigure}
    }

\renewcommand{\KFLTtightframe}[1]{%
    \begin{minipage}{\KFLTimageboxwidth}
    \tcbox[size=tight, colframe=black!60, arc=0mm, colback=white]{ %
    #1
    }
    \end{minipage}
}

\chapter{Back to My Roots: Creating Datasets about the Illicit Massage Industry}
\label{chap:datasets}
\chaptermark{MPForum Dataset and Case Studies}

\begin{savedenv}[fig_homepage]
   \keyfig[htbp]{ft,c={%
   For context, we provide a screenshot of the homepage of the AMPReviews website. The backend is a generic forum software called Xenforo. 
        },
        l=fig:homepage%
        }{homepage.png}
\end{savedenv}

\begin{savedenv}[fig_example_review]
   \keyfig[htbp]{lw=0.7,ft,c={%
 Example review post, which contains semi-structured data (e.g. phone number). Each review shares information about a specific visit. Note that there is no field for price, which users often hide from public view under "Private Details." Follow-up posts in the same thread are free-form.
        },
        l=fig:example_review%
        }{review_example.png}
\end{savedenv}

\begin{savedenv}[fig_posts_daily]
  \keyfig[thbp]{lw=1, ft, c={%
Post date for all posts in the MPForum dataset before Jan. 1st 2023. The time period from March to August 2020 is outlined in red. The dip may be due to the impact of the COVID pandemic.%
}, l=fig:posts_daily%
}{posts_per_day.png}
\end{savedenv}

\begin{savedenv}[tbl_locations]
    \begin{wraptable}[21]{r}{0.25\textwidth}
        \renewcommand{\arraystretch}{0.55} %
        \small
        \caption{Category names on AMPReviews.}
        \label{tbl:locations}
            \begin{tabular}{l}
            \toprule
            Location \\* \midrule
            Boardwide Discussion    \\
            Albuquerque             \\
            Allentown               \\
            Anaheim                 \\
            Bethlehem               \\
            Brooklyn                \\
            Central NJ              \\
            Chicago                 \\
            Dallas                  \\
            Delaware                \\
            Denver                  \\
            Flushing                \\
            Fresno                  \\
            Garden Grove            \\
            Honolulu                \\
            Houston                 \\
            Las Vegas               \\
            Los Angeles             \\
            Monroeville             \\
            New Jersey              \\
            NYC/Manhattan \\
            North NJ                \\
            Orlando                 \\
            PA Other Areas          \\
            Philadelphia            \\
            Pittsburgh              \\
            Sacramento              \\
            San Antonio             \\
            San Diego               \\
            San Francisco           \\
            South NJ                \\* \bottomrule
            \end{tabular}
    \end{wraptable}
\end{savedenv}

\begin{savedenv}[fig_example_nonreview]
   \keyfig[htbp]{lw=0.5,ft,c={%
   Example non-review post. \notegreen{Fields include the number of reviews the user has posted, join date of the user, in addition to the timestamp and text of the post itself.}  %
        },
        l=fig:example_nonreview%
        }{post_example.png}
\end{savedenv}

\begin{savedenv}[fig_threads_location] %
   \keyfig[thbp]{lw=0.95,ft,c={%
            The total number of threads for each subcategory, grouped by location. In orange are posts in "Reviews" subcategories. (No date filtering is done).%
        },
        l=fig:threads_location%
        }{threads_per_location.png}
\end{savedenv}

\begin{savedenv}[fig_user_joins] %
   \medskip %
   \keyfig[thbp]{lw=0.95,ft,c={%
 Join date for all authors in the MPForum dataset. (Note that data is clipped to 300 users,) The dip in user join activity potentially caused by COVID is delayed compared to the dip in posting activity.%
        },
        l=fig:user_joins%
        }{joins_per_week.png}
\end{savedenv}

\begin{savedenv}[fig_iqr]
   \keyfig[htbp]{lw=0.5,c={%
   Distribution of the number of posts per user. The majority of users post less than ten times. Other users have accumulated thousands of posts.
        },
        l=fig:iqr%
        }{iqr.jpg}
\end{savedenv}

\begin{savedenv}[fig_top_domains]
   \keyfig[htbp]{lw=0.85,ft,c={%
        Domains retrieved from the text contents of the MPForum dataset. The y-axis is labeled as a percent of all URLs, while the bars are labeled with the absolute number of times the domain shows up in post text
        },l=fig:top_domains%
        }{top_domains.png}
\end{savedenv}

\begin{savedenv}[fig_umap]
\keyfig[htbp]{ft, lw=1,c={%
    We use UMAP to visualize our word embeddings in two dimensions. We see that mamasan and provider are clustered close together and further from other words. Other clustering is less obvious. \textbf{Top.} Each key word is plotted and labeled with its cosine similarity score to the word "worry" in bold. Other words are seeded from the key words. \textbf{Bottom.} The same plot, but the labels are cosine similarity scores to "anxiety".
    }, l=fig:umap}{umap_combined.jpg}
\end{savedenv}

\begin{savedenv}[fig_cosine_dist]
    \medskip
    \keyfig[W]{lw=0.5,c={%
        Cosine similarities between keywords and negative sentiment words. Colors correspond to colors in previous figure.},l=fig:cosine_dist%
        }{cosine_dist_combined.jpg}
\end{savedenv}
        
 \section{Introduction}
In this chapter we show how natural language processing (NLP) can scale research into the illicit massage industry (IMI). Massage businesses (or individual therapists) in the IMI provide commercial sexual services. These businesses can be found in all major cities in the United States. Due to increasing law enforcement and public scrutiny, as well as the potential for labor and sex trafficking of workers, a better understanding of the IMI is urgently needed. In this chapter we look at online forums where users discuss buying sex in the IMI. Specifically, we release an unstructured text dataset scraped from AMPReviews.net as well as the scripts to replicate the dataset. The dataset includes \char`~800k posts posted by \char`~28k users on AMPReviews and covers a timeframe from late 2018 to early 2023. We release this dataset on (github.com, kaggle.com). We further present three case studies using this dataset: acronym expansion, related domain discovery, and studying user concerns and demographics. These use cases are relevant not only to IMI researchers but also to the wider counter-trafficking and social sciences research communities. Finally, the authors present a call to action for cross-domain research with NLP researchers. The release of this dataset serves as a step toward applying advances in natural language processing to studying the IMI in the pursuit of a more just society.

\begin{mdframed}
\noindent\textbf{Content Warning:} This work may contain sexist and racist language.
Reader discretion is advised.
\end{mdframed}

 \label{sec:background_amprev}
 
\subsection{Background}

  \notegreen{As mentioned in \Cref{sec:imi_stats}}, the IMI is a large industry: in 2018 there were over 11,000 such businesses in the United States (according to Polaris \cite{polaris-social}, a non-profit that works to end human trafficking in the US). Polaris estimated a combined annual revenue for IMBs of \$2.5 billion\footnote{Revenues include payment for massage services, not just illicit services.} in 2018 alone. At the city and state levels, researchers have combined website data and in-person foot traffic to estimate annual gross revenue. 
  These estimates include over \$100 million in Houston \cite{boucheEstimatingDemandIllicit2018},  \$180+ million in Dallas \cite{DallasweismannmelissaEstimatingDemandGross2019},  and \$40+ million in Georgia \cite{rodgersbobStreetGraceIllicit2020}. 
  Despite the size of the IMI, due to its underground nature, even basic information such as the number of customers or whether the IMI is growing or expanding is unknown. %
  By extension, the impact of widespread and highly active hobby forums, and the role these play in normalizing misogynistic viewpoints and cultivating violence, can only be speculated at. The 2020 mass shooting that killed several employees at Asian massage parlors in Atlanta \cite{lam_double-edged_2021} serves as a disturbing reminder of the consequences of these gaps in our understanding. 

  Researchers have focused on estimating the location and size of the industry, including creating classifiers for individual businesses (whether they may provide commercial sex or not). This work is vital but concerns mainly the supply side of the IMI. Less attention has been paid to the self-described "mongers" a.k.a. as the buyers or clientele on the demand side of the IMI.\footnote{"Mongers" is short for "whoremongers".} To this end, we created a dataset by scraping the entirety of an online forum dedicated to discussing buying sex via the massage industry. This website is called AMPReviews\footnote{The acronym AMP stands for "asian massage parlors". However, the data contained within concerns all IMIs and is not limited to Asian massage parlors.} and is publicly accessible on the surface web. The forum contains sections for semi-structured reviews (of providers and businesses) as well as more free-form discussions. Reviews cover, among other things, business location, phone number, as physical descriptions of the provider. \notegreen{An example is shown in \Cref{fig:example_review}.}

\subsubsection{AMPReviews.net (AR)}

\begin{displayquote}
   \textbf{Acronym}. We use AR to refer to \urlcolor{AMPReviews.net}. \\
   \textbf{Definition}. AR is a so-called "hobby forum" for buyers of sexual services with some features specific to massage parlors.  Besides a board-wide discussion category, the forum is split by geographic area. Within each geographic area, the forum contains a section for semi-structured reviews (free-form input for fields such as address) of individual visits to physical businesses and a separate section for general discussion. A screenshot can be found in \Cref{fig:homepage}.
\end{displayquote}

\printsaved{fig_homepage}
            
\printsaved{fig_example_review}

\printsaved{fig_example_nonreview}

\subsection{Contributions}
 In \cref{sec:dataset-details} of this chapter, we describe our dataset in detail. Then, in \cref{sec:cases} we cover the three case studies using our dataset. The case studies are as follows.

  \subsubsection{Case 1: Acronym Expansion} 
  As in other online communities, people posting on hobby boards (a general term for forums where buyers of commercial sex gather) use a plethora of acronyms. Understanding these acronyms can greatly aid researchers new to the topic. We show how a simple application of word embedding algorithms to the MPForum dataset can be used to expand two-letter acronyms that are otherwise incorrectly expanded by tools (e.g. Urban Dictionary) aimed at the wider internet community.
  
  \subsubsection{Case 2: Related Domain Discovery}
  Since the 2016 passing of the SESTA-FOSTA bill, which held website owners responsible for content users posted \cite{SESTA} \cite{FOSTA}, hobby boards and websites have split into many smaller domains. Through straightforward parsing of the forum text, we can retrieve a list of what other domains are frequently mentioned. In particular, with an absolute count of domain mentions, investigators can prioritize which sites to investigate further or gain familiarity with.
  
  \subsubsection{Case 3: User Concerns Exploration}
  
  Finally, we report on a more exploratory case study of potential psychological levers (areas of worry and anxiety) in the IMI buyer %
  community using word embeddings of the MPForum data combined with manually curated lists of words. 

\color{black}
\subsection{Limitations}
 As detailed in \cref{sec:ar_data_limits}, the AR dataset is limited to 12 states in the United States, limiting any spatial analyses. It follows that the AR dataset is relatively small, which makes it easier to explore and work with, but limited for research into causality which is important for policy making purposes. This limited data affects the ability to generalize any conclusions such as general trends in the number of visits to IMIs in the United States. For instance, the IMI may be growing in states not listed on the site. An easy way to strengthen this would be to build a second IMI forum dataset. Similar trends on two separate websites would strongly increase confidence in the trend. This work represents a necessary step in the direction toward that. Furthermore, the AR data is self-reported, and as mentioned in \cite{devriesIdentifyingRiskMarkersRmapsText2022}, users are motivated to overlook signs of coercion or trafficking, or at the least not mention it on forums. Any conclusions about the prevalence of trafficking should keep in mind that this dataset should be treated as a floor on the prevalence and not a representative sample.

\normalcolor

\section{Related Work} 
\label{sec:related_datasets}

Uses of NLP in the counter-trafficking research community include in \cite{NNemojiWang2019SexTD}, where the authors investigate the use of emojis in online sex advertisements using word embeddings as a way to expand a pre-existing lexicon of trafficking flags. Many other works exist where authors attempt to use word embeddings such as FastText \cite{fasttextbojanowski2017enriching} or more compute-intensive methods (e.g. Transformers models) \cite{devlin2018bert} to classify posts on a general-purpose forum as belonging to illicit work or not. For instance, in \cite{simonsonThesisSemiSupervisedClassification} the authors also show how semi-supervised learning and FastText can be used to classify social media posts (e.g. on Twitter) as related to sex work or not. Similar work exists for wildlife trafficking as well, for instance \cite{xuPangolinTweetClassificationML2019} used unsupervised topic modeling (clustering) to detect Twitter posts related to trafficked items (ivory and pangolins). Our datasets focus on extracting insights from sites dedicated to illicit activity, rather than filtering illicit activity from a general purpose platform.

\subsection{Case Studies}

The case studies for our dataset cover uses that generalized beyond the illicit massage industry and extend more broadly into other applications of computer science. 

\subsubsection{Acronym Expansion} 
 
Acronym expansion serves an important use in the medical and scientific communities. Older approaches for acronym expansion in e.g. the medical setting used supervised learning algorithms such as Naive Bayes and Decision Trees \cite{joshiSurveySupervisedLearningClinicalAcronym2006}. More recently, in \cite{veysehScienceDocumentAcronymWorkshop2022} the authors describe a Scientific Document Understanding Workshop with tasks that included acronym extraction and acronym disambiguation. Many methods rely on variations of processing a sentence (where the location of the acronym is known) through a pre-trained model (e.g. BERT \cite{Liu2019RoBERTaAR}) to get word embeddings. These embeddings are then processed through a multi-layer perceptron to get a prediction score for expansion candidates, as described in \cite{songT5TransformerAcronym2022}.  
Our thesis represents the first application of acronym expansion to the IMI domain. Existing acronym datasets tend to use Wikipedia and other sources (e.g. clinical notes) that have a different distribution of acronyms than used in the IMI. We use unsupervised learning (word embeddings) to do so.

\subsubsection{Domain Discovery}

Cybersecurity researchers are interested in detecting malicious domain names, which can be generated in large quantities by Domain Generation Algorithms (DGAs). In \cite{bermanSurveyMLDomainNameGeneration2019}, the authors survey of the machine learning approaches (primarily using LSTMs and RNNs) for detecting malicious domains.\footnote{The same survey points out that machine learning algorithms tended to rely on using purely text-based features as those were the most widely available datasets, showing the importance of datasets in influencing research directions.} For our case study, we found that a simple parser for links and counter for popularity was sufficient to provide a list of top sites of interest. In our use case, generally a human (rather than an algorithm) has picked a reasonable URL intended for other humans to use (vs. a URL for bots to evade automatic security measures).

\subsubsection{Sentiment Analysis and User Profiling}

Sentiment analysis has wide applications, with the most well-known being applications in advertising and recommendations. For instance, a user may leave a review for a movie, and analyzing the sentiment of that review (positive or negative) can lead to recommendations for other movies. Applications in public health include analyzing social media text to understand user sentiment toward healthcare interventions. 
\cite{liSpaceTimeDepressionSignsCOVIDSocialMedia2020} classified tweets from January to April of 2020 by sentiment (stressful vs. not-stressful), then manually coded discussion topics for each sentiment, in order to find trends in what were top user concerns over time (e.g. shifting from concerns about infection to concerns about finances as the pandemic continued). We do not explore spatiotemporal patterns and instead focus on a binary question ("is this a concern or not"). In our case, we are exploring possible deterrents to discourage users from illicit activity.

\subsection{Previous IMI Datasets}

Previous datasets used to study the IMI face various ethical concerns for public release. For instance, the Rubmaps dataset contains detailed business address information. The Rubmaps forum discussion text also requires logging in, a distinction that precludes straightforward release of the dataset. On the other hand, our forum dataset is public access and does not require any login. Much of the data is in fact already archived (e.g. on the Wayback Machine on archive.org). Thus, our dataset represents the first public release of natural language text related to the IMI.

\section{Dataset Methods} 
\label{sec:dataset-details}

  In this section, we cover the origin of the dataset, the methods used for dataset creation, and exploratory analysis of the dataset itself. This gives important context for understanding the case studies described in the next section.
  
  \subsection{AMPReviews.net}
  The AMP Reviews forum is dedicated to discussions and reviews of illegal massage parlors in the United States. (Note that they are legal elsewhere). The forum consists of \textbf{categories, subcategories, threads, and posts}.
  
\printsaved{tbl_locations}
  
  \textbf{Categories.} The forum has 61 categories that are public
  access (with many more forums that are private access). There is a
  single "Boardwide General Topics" category. Otherwise, each category
  corresponds to a physical location (either state, city, or general
  geographic area). These areas span 12 states. \textbf{Subcategories.}
   Besides the "Boardwide General Topics" category, within each category
  there are two subcategories: discussions and reviews. Thus, there are
  30 locations represented in this forum. \textbf{Threads.} Discussions
  are similar to discussions on other forums. Each thread represents a
  different topic. For reviews, users post about their IMI experiences.
  Each thread represents an individual visit to a massage business, and
  is usually titled with the name of the sex worker as well as the
  business. \notegreen{An example can be found in \Cref{fig:example_review}.} \textbf{Posts.} Individual messages in a thread are called
  posts. Other users are free to add comments. \notegreen{An example can be found in \Cref{fig:example_nonreview}.}

  \subsubsection{Dataset Limitations}
  \label{sec:ar_data_limits}

  \textbf{User-base.} We might expect that users that comment on a forum dedicated to IMIs would form a skewed subset of the IMI customer base, namely that they would be more regular or frequent customers (which could correlate with higher income and an older population on average).
  
  \noindent\textbf{VIP Access.} There are many more forums and threads that are private access (a.k.a. VIP Access). Access to private areas can be acquired by either paying money or by posting reviews. Purchasing access costs in the range of 30 USD per month, with discounts for 2 and 6 month memberships, charged via credit card.\footnote{In total, there are \textbf{four} such private subcategories, ones for: New York, Allentown (Pennsylvania), Philadelphia (Pennsylvania), and New Jersey.} Meta-information (e.g. the number of threads and messages) is not available for these private categories.
  
  Additionally, even in the public review posts included in our dataset, part of the data is hidden for VIP membership. Though the form is free-form, users are encouraged to hide information such as the price in this section (though users may discuss pricing in replies to the review).
  
  \noindent\textbf{Moderation.} According to moderators on the forum, reviews are automatically publicly posted. The reviews then go through a moderation process (there are multiple moderators) where moderators go back and determine the quality of the review. If the review is good enough (e.g. detailed), the user earns VIP access, otherwise the review is deleted. Finally, some users choose not to display their metadata (e.g. number of reviews posted).

  \noindent\textbf{Limited Geographical Coverage.} Only twelve states are included in the AMPReviews forum. See \Cref{tbl:locations}.

  \noindent\textbf{Data Not Collected.} The current dataset does not include the public direct messages between individual users ("profile posts"). As mentioned before, we did not collect any data that required a login or paid access.
  
  \subsubsection{Data Collection Method}
 
  We relied on a popular open-source tool, a Python library called Scrapy \cite{scrapy}, and built a custom scraper to interact with the Xenforo pages. Our Scrapy spider crawled all categories, then all subcategories, then all thread URLs, following "next page" links as appropriate. From that (static) list of URLs, we visited each URL and scraped all the posts within each thread. For each post, we used CSS selectors to select desired information (such as post text). 
  
  In human-readable form, the fields processed from the raw HTML cover roughly the following topics (number of fields shown in parenthesis):
  
    \begin{itemize}%
        \setstretch{1.5} %
        \item \textbf{(1) Timestamp} of scrape
        \item \textbf{(4) Thread:} title, category, total pages, URL
        \item \textbf{(6) Author:} username, user title, \# of reviews (if shown), \# of posts (if shown), profile URL, join date
        \item \textbf{(4) Post:} thread page number, post ID, posted date, text
        \item \textbf{(4) Misc.}: \# of likes on post, liker usernames (if exists),
          quoted text in post (if exists), author of quoted post
    \end{itemize}
    
    Note that we used conservative intervals between HTTP requests such that the data was collected over multiple days. As a result, some users could have multiple post counts associated with their record. For instance, say user \code{xyz123} posted on thread IDs \#1 and \#3. We could scrape their profile on thread \#1 and find their \# posts is two. Then, if they post two additional replies in thread ID \#1, by the time we scrape thread ID \#3 their profile would list their \# of posts as four. For this reason we include the timestamp of scrape in our dataset.
  
\notegreen{
In our analysis, we focus on the user information and the post text, ignoring the miscellaneous features (quotes and likes).
}

 \subsubsection{Dataset Processing}
 
  As previously mentioned, we performed light processing on the raw HTML page in order to separate out our desired information. We pre-cleaned the text by removing all HTML tags using the Beautiful Soup \cite{bs4} and lxml libraries \cite{lxml}. We also separated quoted text from the post itself (e.g. when users are replying to earlier posts). We parsed user information using CSS selectors (a feature built into the Scrapy library).
  
  \textbf{Case Studies.} Each case study had additional processing. At a high level, for the acronym case study, we performed standard steps (for training word embeddings) such as removing stop words and stemming words. We then trained a word embeddings model using word2vec \cite{word2vec}. For the domain discovery study, we stripped HTML tags from the post contents, then used the URLExtract library to pull out top-level domains mentioned in the text. For the user concerns exploration, we did not do further pre-processing. We used the previously trained embeddings and also used the Uniform Manifold Approximation and Projection (UMAP) algorithm \cite{umap} to visualize the results. Further details are provided in the respective case study sections.

  \subsubsection{Dataset Characterization}

  \printsaved{fig_posts_daily}
  
  As previously mentioned, our dataset contains 800k posts made by 29k users. From 2021 to 2022 (inclusive), there was an average of 522 new posts and 323 unique users posting each day. %
  
\printsaved{fig_threads_location}
     
  \textbf{Posts over Time.} \Cref{fig:posts_daily} plots the absolute number of posts posted on a daily basis. We
  see that the forum starts expanding rapidly after the demise of the Backpage website due to SESTA-FOSTA, which was signed into law in 2018. The next period of note, boxed in red, is the dip in activity starting in March 2020. This is likely in response to the lockdown restrictions imposed during the COVID-19 pandemic. The gradual increase in posts roughly correlates to the time period where restrictions in the United States were lifted. This provides reassurance that the forum is reflective of general activity in the IMI space and not a completely skewed presentation of IMI activity.

  \textbf{Threads by Location.} \Cref{fig:threads_location} shows the number of threads per category. There are significantly more review posts than discussion posts. The most posts are in New York City, followed by the categories for North and for Central NJ. Note that Flushing is listed as a separate location from New York City.

\printsaved{fig_user_joins}
  \textbf{Users}. In our dataset, we collected information on (28k) users, which is significantly less than the forum self-reported statistics. We speculate this is due to users that only post in private forums, which we did not scrape. In \cref{fig:user_joins} we plot the join date of the users. There is a dip during the same COVID period marked in red, however the downward trend appears to start earlier (in fall 2019). Possible causes are unknown.

\printsaved{fig_iqr}
  Additionally, we report statistics on the number of posts per user in \Cref{tbl:user_posts}. This same variable is posted in \Cref{fig:iqr}. Outliers are described as values more than $1.5\times$ the interquartile range. In our case this is users with $\geq45$ posts. We also plot a strip plot overlaid on a violin plot of the posts for all users, and then a further one focused on outliers (here defined as users with posts more than 2 standard deviations away from the mean post counts). The majority of users make fewer than 20 posts, while one user has made 8,264 posts.\footnote{According to parsing the user profile page. The actual number of posts in our dataset for the same user comes in at 6,814. A possible cause for the difference may be that the profile statistics include private posts, while we only scraped public posts.}%

  Considering that AMPReviews covers only a subset of the U.S. (12 states) and is one of several other forums dedicated to mongers of the IMI, as well as that our data contains only the publicly accessible portion of the website.
    
  \subsection{Selected Quotes}
  
  To illustrate the potential of this dataset, we also present a few direct quotes (some excerpted from longer text). Though all information is self-reported, the responses give some sense of the breadth of the monger population in terms of age, income, education, marital status, and ethnicity. In \Cref{tbl:posts_money_freq} we display selected quotes from the dataset about money and frequency. The self-reported results speak to some extent to income levels of the population (likely high, to support thousands of dollars in expenditures) as well as habits (fluctuating over time, and frequency can be quite high, ranging up to every other day).

\begin{table}[htbp]
\caption{Example forum posts about money and frequency. All text is actual quotes.}
\label{tbl:posts_money_freq}
\begin{tblr}{
colspec={X[0.9]|XX},
  row{1} = {gray!40},
  row{even} = {gray!20},
            }
\hline[1pt]
\textbf{Thread Title}                & \textbf{Example Posts \#1--4}         & \textbf{}         \\ \hline[1pt]
So who spent how much in 2022?”      & mine was around 6150   & around 8 k per year                     \\
                                     & 15k+                   & 30k                           \\ \hline
                                     
How many times do you visit a month? & used to do once a week & I binge while traveling \\
 & Averaging 3 times a week and go to 4 only on occasions & Personally, I would sell the house so I can have more free time and money for AMPS \\ 
 \hline[1pt]
\end{tblr}
\end{table}

                  Previous researchers found that white men are the primary 
                  clientele. The MPForum dataset shows that is not a full
                  understanding of the monger demographic. For instance, some usernames
                  are in other languages (e.g. Korean). Others mention ethnicity
                  directly in posts, e.g. 
                  
                  \begin{itemize}
                    \item "I'm $\frac{1}{2}$ Chinese/Korean"
                    \item "I'm Hispanic and grew up with [...]" 
                  \end{itemize}
                  
                  Or gender:
                  \begin{itemize} 
                    \item "I saw a female monger on rubmaps that contributes reviews
                  quite often"
                  \end{itemize}
                  
                  Note though that this singular quote is in the third person, so an anecdote of an anecdote.\footnote{This does bring up interesting questions, e.g. are there incidents of women seeking lesbian sex at massage parlors?}

            \begin{wraptable}[8]{r}{0.25\textwidth}
                \renewcommand{\arraystretch}{0.6} %
                \small
                \caption{Posts per author, according to profile text}
                \label{tbl:user_posts}
                \begin{tabular}{@{}ll@{}}
                    \toprule
                    Mean       & 30.8   \\
                    Std. Dev. & 140.4  \\
                    Min.       & 0      \\
                    25\%       & 2      \\
                    50\%       & 5      \\
                    75\%       & 19     \\
                    Max        & 8,264  \\  \midrule
                    All Posts & 28,268 \\  \bottomrule
                \end{tabular}
            \end{wraptable}

                  Others may mention age:
                  
                  \begin{itemize}
                    \item "I am almost 40 and have been mongering since my early 20s" 
                    \item "Of they are 45, they are still 25 years younger than me." 
                    \item "been there a couple times when I was in grad school (2009 ish)"
                  \end{itemize}
                  
                  And other quotes mention marital status: 
                  \begin{itemize}[itemsep=0pt]
                      \item "Single 40, never married and no kids." 
                      \item "If I had to guess, 85\%+ of mongers are married or in a committed relationship" 
                      \item "married once , never again. [...] I have my adult children \& grandchildren."
                  \end{itemize}
                  
                  Other posts mention industry or profession directly, e.g. the
                  following quotes are all present in the dataset: 
                  \begin{itemize}
                  \onehalfspacing
                      \item "I work in finance and logistics [...]", "I work in customer service", "I work in a medical setting", "I work in midtown Manhattan", "I work in pharmaceutical consulting", "I work in tech", "Being in banking for over 30 years, I can tell you [...]"
                  \end{itemize}
                  
                  We do not explore the full extent of demographic information in the
                  MPForum dataset, but hope the inclusion of these quotes helps explain
                  our research direction and motivate other researchers.

  \section{Case Studies}
  \label{sec:cases}
  
  \subsection{Acronym Expansion}
  
  The use of acronyms often serves as a barrier to entry for researchers
  and other stakeholders. Consider the text: "Once the MMS trusts you,
  then you can call ahead." Readily available sources such as Google,
  Urban Dictionary, and others do not readily return the expansion of
  the acronym. Even using the
  AMPReviews forum search tool
  returns no results as the minimum search length is four characters.
  Without time spent reading through the (at times toxic) text, the
yntheit
expansion is not obvious.
  Although there exist lists of acronyms on these forums, these lists
  can stretch to hundreds of acronyms and contain duplicate or outdated
  acronyms. Thus, for this case study, we show that with the MPForum
  dataset we can automatically find candidate expansions for some of the
  obscure acronyms used in the monger community. For the case study, we
  focus on two-letter acronyms, but the method could easily be extended
  to longer acronyms. Two-letter acronyms are important due to the
  limitations of the built-in forum search functions as mentioned above.
  Specifically, our algorithm focuses on initialisms, where each letter
  stands for the beginning of a separate word.
  
  \subsubsection{Methods}
  We created a word embedding for our dataset. To do so, we first
  pre-process the text, construct bigrams, and use a library to create a
  word2vec model. We then pull the most similar bigrams to our
  initialism and return that.
  
  \textbf{Pre-processing.} We customized a few pre-processing steps to work better for our task. We do not lowercase, as casing can help distinguish acronyms from normal words. We also decrease the minimum word length so that we keep two-letter words (vs. GenSim by default keeps only three-letter words and up). 
  We also customize the list of stop words to our task.\footnote{Stop words are English words (e.g. a, an, the) that are commonly removed as they can detract from creating semantically meaningful word embeddings.} For instance, "other" is a stop word in the default GenSim list.
  We choose to remove a relatively limited set of stop words
  (NTLK's 179 words, vs. the 300+ words by default in
  GenSim, SpaCy, and sci-kit learn). As a concrete example, these steps allow us to retain both so and SO in our text corpus. Ordinarily, "so" and "SO" would be one entity, but the case distinction is important since the latter could be an abbreviation for "\underline{s}ignificant \underline{o}ther". The words would also otherwise be too short and also included in stop word lists.

  \textbf{Bigram Construction.} We also use GenSim Phraser, which implements an algorithm that
  constructs bigrams (two word phrases) from the dataset. Specifically,
  it uses the algorithm described in \cite{word2vec} which creates bigrams
  from words that appear frequently together and not often separately.
  
  \textbf{Word Embeddings Construction.} We used the word2vec architecture. This family of models trains
  we predict the current word given the surrounding context (we use a
  window size of 5, so two words of context before and after to predict
  the missing middle word). We use the mean of the context word vectors
  in order to predict the missing word (this is the default in the
  GenSim library). We do not limit the final vocabulary size of our
  model.\footnote{Note that Skipgram is
  theoretically more appropriate as our dataset is small and has many
  infrequently seen words, however practically the continuous bag-of-words CBOW model was sufficient for this case study.}
  
  Once this model is trained, we can use the "most similar" function in
  GenSim to find co-occuring words. In the GenSim implementation, this
  method computes the cosine similarity between the means of the weight
  vectors of the given search term of each key (word) in our word
  embeddings. Note that this means that we can only compute the
  similarity between words present in the original vocabulary.
  
  \textbf{Expansion Algorithm.} We return the most similar bigram to our acronym (in terms of word
  embedding metric space) as the candidate expansion, with a fallback to
  the most similar unigram, in the top fifty results returned by the
  most\_similar function. We use the default GenSim function for this
  task.

    \begin{singlespace}
\begin{mdframed}
    \begin{verbatim}
function expand_initialism():
    - Find top 50 most similar words to query (using cosine word embedding
    vector distance)
      Word can be bigram or unigram
    - We check all bigrams in descending similarity:
        - If initial letters match query -> return bigram
    - Else, check all unigrams in descending similarity
        - Return the first unigram that starts with same letter 
        - Else if no unigram matches -> return None
    \end{verbatim}
\end{mdframed}
    \end{singlespace}

\subsection{Qualitative Results}

      We sought to demonstrate how our tool provides a strong complement
      to existing general-purpose tools. We consider a few acronyms that
      are very specific to the IMI. For the acronyms our algorithm
      expanded correctly, we compare the expansions from Urban Dictionary
      and ChatGPT. Note that for the ChatGPT and GPT4All \cite{gpt4all}
      queries, we provided additional helper text, as otherwise the
      results were lackluster. These prompts are described in the table.
      All queries with these general-purpose tools were conducted in a
      single chat session.
                      
        \begin{table}
          \centering
          \caption{We use three existing tools to qualitatively evaluate how our acronym expansion method performs. To help the GPT tools, we gave contextualized prompts. Prompt 1: "you are a sex trafficking researcher. what does \_\_ stand for?". Prompt 2: "In the context of massage parlors, what does \_\_ stand for? be concise" Prompt 3: "You are researching massage parlors. What does \_\_ stand for?" For N/A, the ChatGPT interface returned "I'm not aware of any common industry-wide abbreviation"}
          \begin{tabular}{c}
            \frame{\includegraphics[width=\textwidth]{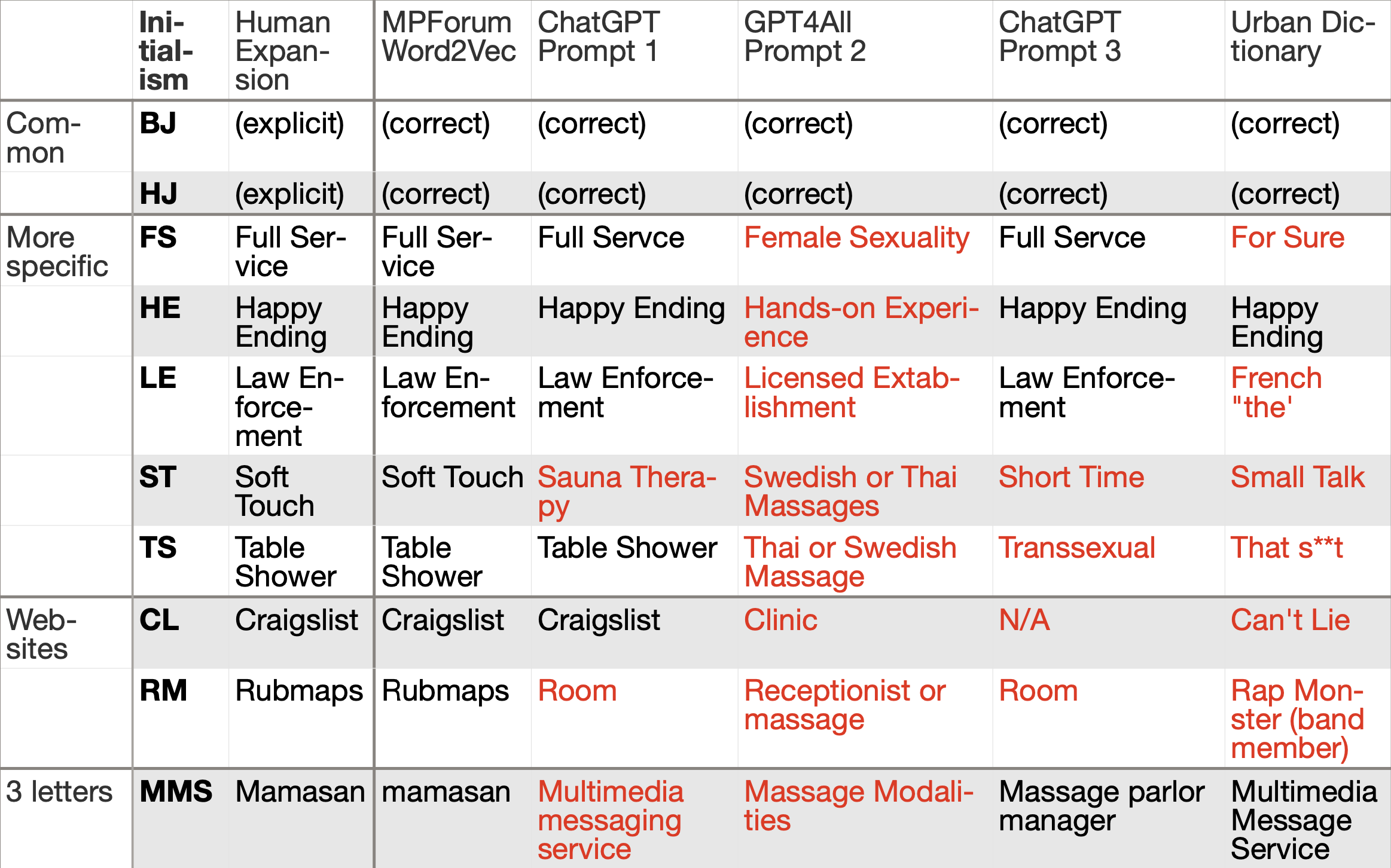}} \\
          \end{tabular}
          \label{tab:image-as-table}
        \end{table}

     \subsection{Discussion}
        
     This case study shows that we can use the MPForum dataset to
     investigate slang and acronyms in the IMI monger community. The
     expanded term is significantly easier than the acronym to search for
     on Google or the forum directly. Our algorithm is entirely
     unsupervised, fast to train, and easy to understand. This mean it
     can be easily generalized to other niche communities, including ones
     in other languages. This tool is powerfully specific: if a result is
     returned, it is guaranteed to be present in the dataset and
     researchers can then gain a better picture of how the word is
     used.
        
      On the other hand, our algorithm cannot pull from a wider base of
      language knowledge. Our algorithm fails to expand "SO" to
      "significant other". Preliminary investigations showed that
      increasing the continuous BoW window size did not improve results. Likely the
      failure results from a fundamental limitation of our dataset: it is
      small. Additionally, due to our removal of stopwords such as "the",
      websites such as TER (\underline{The} Erotic Review) will not expand correctly
      either.
      
      General purpose tools such as ChatGPT not only expand the terms but
      also explain them. However, these tools can require training to use
      effectively. As mentioned, ChatGPT did not give relevant responses
      without prompt engineering to give it conversational context (e.g.
      explicitly stating the IMI context). However, applying that same
      prompt for acronyms that actually stand for websites (e.g. RM) can
      lead to ChatGPT giving incorrect answers. Additionally in some cases
      the acronym is expanded correctly but the explanation not fully
      correctly; this can cause confusion for non-specialists in
      understanding where ChatGPT may fail.
      
      On the whole, given the speed and ease of training and querying,
      open source code, and powerful specificity to an individual
      subculture or community, this case study shows how the we can use
      MPForum dataset can fill in gaps in existing acronym expansion
      tools. For future work, it would be interesting to see how this tool
      performs on forums in other languages or regions.\footnote{A quick inspection of Spanish-language forums indicates that most acronyms are actually standardized from English.}
      
      \subsection{Related Domain Discovery}
      
\printsaved{fig_top_domains}                      

      In this case study, we show how the MPForum dataset can be used to
      automatically detect related domains of interest. 
      After the SESTA-FOSTA law was passed in the United States in 2018, website operators could be held responsible for content on their website. SESTA-FOSTA stands for two separate bills: the FOSTA (Allow States and Victims to Fight Online Sex Trafficking Act) is the House bill and SESTA (Stop Enabling Sex Traffickers Act) is the Senate version \cite{SESTA}\cite{FOSTA}. Before these bills became law, online sexual services activity centered around the Backpages.com website.\footnote{The website is archived on the
      Wayback Machine at \url{https://web.archive.org/web/20170101000000*/backpages.com}} After Backpages was shut down (as a result of the law), users dispersed to several different domains. Due to limited researcher time, prioritization of the large list of domains is desperately needed.
      
      \subsubsection{Methods}
      
      We parse all post text to retrieve any URLs directly mentioned in
      the text using the URLExtractor Python library. This library looks
      for known top-level domains and extracts urls using delimiters such
      as spaces. We use TLDExtract to extract domains and ignore subdomains. For example, \urlcolor{www.example.com} and \urlcolor{example.com} are counted as a single entity. We then used the data processing library Pandas to do value counts.
      
      We then split the domain data into two time frames, one from 2018 to the end of 2020, and one from 2021 to the end of the collection period. We again used the Python library Pandas \cite{pandas} to do so. 
        
      \subsubsection{Results}
      
\begin{table}
\centering
\caption{Here we see the change in the top URLs over two time periods, from 2018 to the end of 2020, and then from 2021 to the end of the collection period (early 2023). The top 5 domains remain unchanged (shown in green). In bold are domains that appear in one time period but not the other.}
\begin{tabular}{c}
\frame{\includegraphics[width=\textwidth]{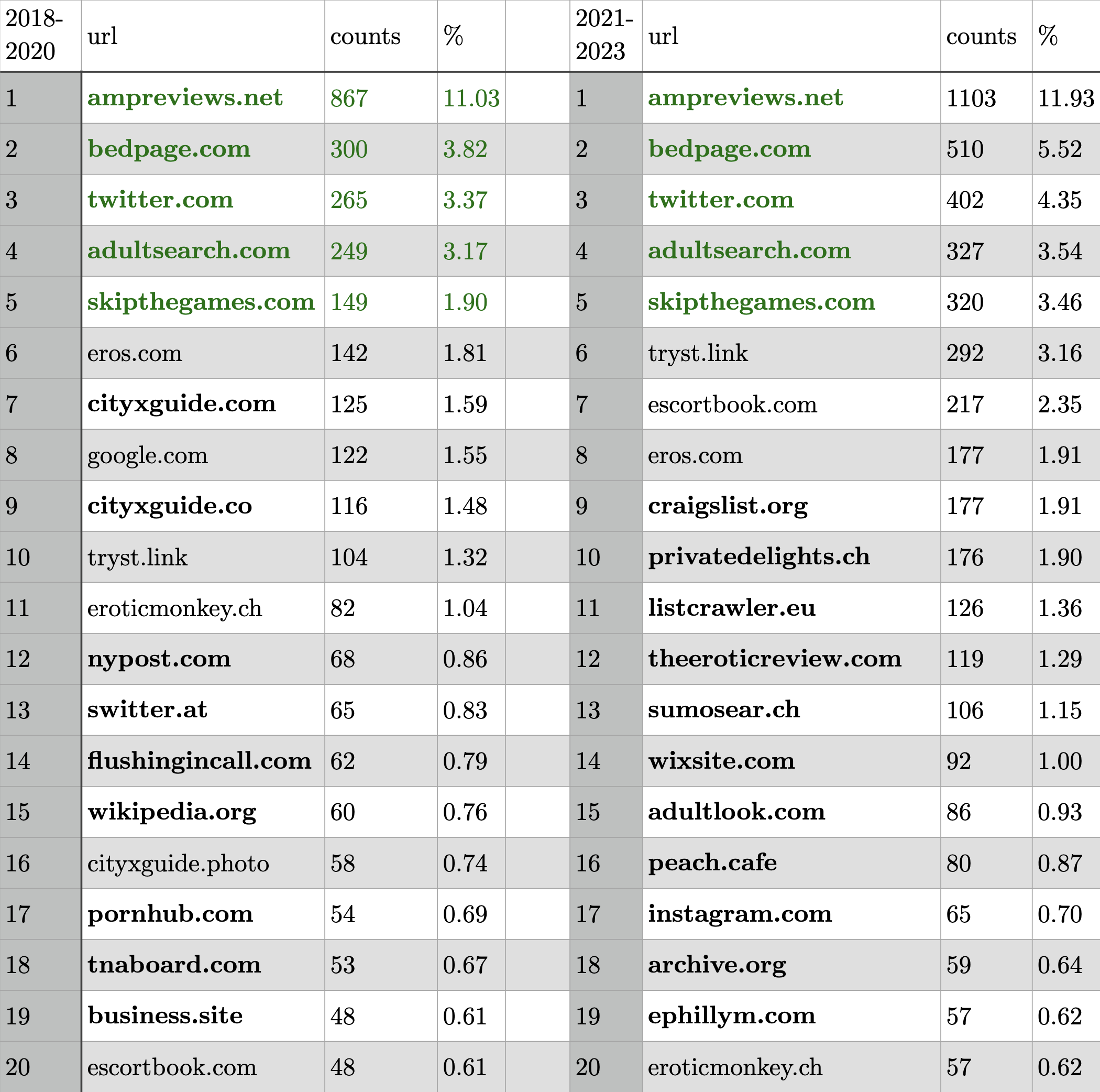}} 
\end{tabular}
\label{tbl:delta_domains}
\end{table}
        
      The results of the value count across all time is shown in \Cref{fig:top_domains}. We see that the highest domain is the self-referential domain \urlcolor{ampreviews.net}, after which the clear winner is \urlcolor{bedpage.com}.
      
      The results of disaggregating the data across two time frames is shown in \Cref{tbl:delta_domains}. An easy change to spot is the demise of \urlcolor{cityxguide}.\footnote{Actually, if the \urlcolor{cityxguide.com} and \urlcolor{cityxguide.co} domain counts are summed, they would be the 5th most popular domain in the first time period.} In 2020, cityxguide was taken down by law enforcement when the owner was arrested. By manual inspection, the popularity of Twitter is from links to Twitter accounts for sex workers or agencies. Similarly, the entry of \urlcolor{wixsite.com} into the top twenty in the second time frame is likely from websites custom-made for a particular agency (or trafficking group). Wix allows users to easily create websites without programming knowledge. This shows that workers in the IMI and adjacent industries readily adapt to new platforms for advertising and connecting with clientele. The naming of many of the remaining domains clearly suggest commercial sex, with the notable exception of \urlcolor{sumosear.ch}. The latter turns out to be a search engine specifically designed for users to enter phone numbers to find related websites.

\notegreen{
The plethora of domains emphasizes the difficulty that researchers face in trying to grasp the extent and dynamics of the IMI. In this case, the MPForum dataset offers a fast way to start triaging which websites should be explored. For instance, the fact that the top five domains remained the same across both time periods is interesting. Our results not only show related top-level domains to investigate, but also give a sense of the importance of each domain based on the frequency (and absolute counts). 
}

  \subsection{User Concerns Exploration}
    
  MPForum can also drive insights into user behavior and demographics.
  The motivation for this case study is as follows:
  IMI policies tend to focus on the supply side in a punitive way.
  Massage parlor businesses are easier to track than individual
  people. Raids conducted (often in response to community complaints,
  which can have anti-immigrant or racist biases) are known to be
  traumatic for the women that all stakeholders are trying to help.
  They can result in arrest, large fines, deportation, and
  even side effects such as falling prey to scams by lawyers. 
  Applicable laws tend to frame the situation as the massage therapist
  mis-using a position of trust (the therapeutic relationship) to
  solicit sex. On the flip side, little research is done into
  protecting massage therapists from harassment.\footnote{From reading our MPForum dataset, we
  know that mongers may pressure women to have sex (or more invasive
  types of sex).} Even the shift in focus over time to "less
  harmful" interventions, which focus educating landlords about the
  IMI, could reinforce existing anti-immigrant or anti-Asian injustice. For details of the above, refer to the interviews summarized in \cite{chinjohnj.IllicitMassageParlors2019}.\footnote{For instance, for some
  police jurisdictions, intervention in the IMI usually is in response
  to neighborhood complaints. The neighborhood complaints themselves
  may arise from anti-immigrant sentiment.}
  
\printsaved{fig_umap}
    
  The MPForum dataset could be used as a discussion tool to widen the
  space of policies people consider, beyond these existing policies
  that focus only on the supply side (the businesses and employees). Prior
  investigations characterize mongers of massage parlors as relatively
  risk-averse. %
  We were therefore interested in understanding
  what users are worried about. Based on manual inspection (directly
  searching through the entire text), users were generally not
  concerned that the women were trafficking victims (and believed they
  could discern if so). Some users did talk about their relationships,
  and one user thought that the majority of mongers are married.
  
  Direct quotes from the MPForum text include:
  \begin{itemize}
  \item "I recently got caught by her [...] it really tore up our
    marriage, but I was able to fix it and we worked things out, now I
    dont care to venture around or monger. I know the consequences
    [...] I am just to (sic) afraid of losing my SO and much more. So I would
    advise if yall continue to do it, do it very discreetly, change your
    clothings, use non scented soaps/lotions [...]"
  \item "this alleged hobby has also interfered with my real life
    relationships"
  \item "Over time you'll find that hobbying comes to mind
    less and less. [...] Mongering really is like an addiction and a
    lot of the same strategies work."
    \end{itemize}
    
    We use word embeddings to explore the hypothesis that mongers as a
    population found "being found out" concerning.
    
  \subsubsection{Methods}

    Intuitively, buyers of commercial sex would be concerned about law
    enforcement or sexually transmitted diseases (STDs). By training a
    word embedding on our dataset, similar word embeddings could be
    interpreted to mean that concerns about damage to interpersonal
    relations are comparable to concerns about law enforcement or STDs.
    Unlike for the acronym case study, we did not generate bigrams.
    Otherwise the pre-processing and training steps for the word embedding
    are identical.
    
\textbf{Keywords}. We used the UMAP method to map word embeddings down to
    two dimensions for visualizing.
    We consider two negative sentiment words and three sets of keywords:
  
    \begin{itemize}
      \item \textbf{negative sentiments:} worry, anxiety
      \item \textbf{hypothesized concerns:} SO (for significant other), marriage
      \item \textbf{other concerns:} LEO (for law enforcement officer), STD
        (sexually transmitted disease)
      \item \textbf{control words:} MMS (for mamasan), provider, parking,
      table
    \end{itemize}
    
  We also programmatically picked four additional words to each of
  these keywords using the \code{most_similar} function in GenSim. This
  helps visualize the clustering of word embeddings in the projected
  space. We additionally label only the initial keywords (shown in
  bold) with their cosine similarity from the negative sentiment words
  ("worry" and "anxiety" respectively). The cosine similarity of the
  negative sentiment word to itself is one. The projected results can
  be seen in \Cref{fig:umap}. For UMAP parameters, we chose
  number of components as 2 (as the goal is visualizing the
  embeddings). We adjusted \code{n_neighbors} to 7 (instead of the default
  10) to better capture the local structure around our words.
  After visualizing the embedding space, we chose a bar plot to
  explore the data in more detail. Specifically, we plot the cosine
  similarities (shown in the \Cref{fig:umap} in bold) as a bar chart.
  
\printsaved{fig_cosine_dist}

  \textbf{Cosine distances}. We select the negative sentiment words \code{worry} and \code{afraid}
  as terms of interest. We then plot the cosine similarities between
  the negative sentiments of "afraid" and "worry" and our keywords.

  The results can be seen in \Cref{fig:cosine_dist}.

\subsubsection{Results and Discussion}

\textbf{UMAP} - Although the data is noisy, \Cref{fig:umap} shows that
word embeddings for words related to the hypothesized concerns
(about relationships) tend to be clustered closer to the negative
sentiment words than to the control words (e.g. table and
provider).

The words for parking are projected to a surprisingly close space
while the STDs are clustered further away. The interpretability of
the UMAP embedding is difficult. The large gap in cosine distance
between "anxiety" and "worry" also makes interpreting the UMAP and
the word embedding similarity scores difficult.

\textbf{\Cref{fig:cosine_dist}} - The significance of the cosine similarity
metric is clearer in this figure. We see that the highest similarity
words from our selected set of keywords to each negative sentiment
word ("anxiety" or "worry") correspond to our intuition -\/- parking
and table are relatively dissimilar, while "STD" and "LEO" are
relatively similar in word embedding space. On this linear scale, we
can see that our hypothesized words ("marriage" and "SO") are also
relatively close to our negative sentiment words. The word
embeddings reinforce the qualitative impression that many mongers
worry about the impact of their actions on their relationships.

\subsubsection{Discussion of Motivation}

Many theoretical concerns that can be raised from both the academic
and public policy perspectives. Using word embedding space to
explore mongers' concerns, although an interesting
application of NLP, feels fairly circuitous compared to asking
directly. The results are hard to interpret. In terms of public
policy, the hypothesis was that mongers may be as concerned about
harming their romantic or sexual relationships as they are about law
enforcement. However, the end result of applying social pressure on
individual behavior can be unpredictable. For instance, it is
possible that backlash from partners or the public could be against
the sex workers rather than the mongers.
In addition, ideally, our work would ask the question "what are concerns that mongers have expressed and how concerned are they relatively" rather than the binary "are mongers concerned about social relationships". We leave this to future work.

The goal of this exploratory research is to broaden the conversation
about potential IMI policies. The concrete data points in MPForum
can create space to discuss a wider range of possible policies than
campaigns to educate landlords or improve human trafficking
awareness.

  \section{Conclusion and Call to Action}
  
  In this thesis, we present a novel natural language dataset, MPForum
  (massage parlor forum), which focuses on the monger (sex buyer)
  population. To the author's knowledge, MPForum is
  also the first publicly released dataset about the IMI. We sought to
  demonstrate how even basic tools and understanding, e.g. what are
  the most popular related domains, are missing in IMI research. With
  our case studies, we not only show the usefulness of our dataset,
  but also how even basic research can fill gaps in our knowledge of
  the IMI.
  
  \subsection{Future work} 
  We present possible future work in the form of
  a call to action to the computer science, computational social
  sciences, and machine learning communities. We list here two main
  questions and a list of the many associated questions that
  researchers could try to answer with the MPForum dataset. The first
  question represents higher hanging fruit and involve questions that
  require interdisciplinary research between the social sciences and
  the computational communities. The latter focuses on more basic
  natural language processing algorithms. Both questions can be
  generalized for understanding other niche online communities.
   \begin{itemize}
    \item \textbf{Question:} To what extent do these forums promote and
      normalize misogyny and contribute to real life
      harm?
    \item \textbf{Motivation:} The shooting deaths of six massage parlor
      workers in 2021 showed that the attitudes formed in these niche
      communities online can have fatal consequences.
    \item \textbf{Preliminary sub-questions:} Based on the MPForum text, can we create a measure of
      toxicity (e.g. the severity of the misogynistic viewpoints expressed)
      of a set of text? Given that, can we measure the toxicity of a forum
      or a series of posts by individual users? If so, do particularly toxic
      users tend to interact more often? Could the toxicity of individual
      users be measured over time (e.g. can sudden events trigger rapid
      worsening of toxicity, as a warning sign for potential physical
      violence)?
  \end{itemize}
  
  \begin{itemize}
    \item \textbf{Question:} What can we learn about user demographics, such as
      their occupations, income levels, age, and so on?
    \item \textbf{Motivation:} Understanding the user demographics helps put a face on the monger
      population. This can also shift attention from the providers to the
      mongers. Policy-makers can also better estimate the impact of
      implementing any particular laws or deterrent.

    \item \textbf{Preliminary sub-questions:}
    What types of information are present in the dataset?
    How can we extract professions from a natural language dataset? How
    can we separate text about mongers versus about providers (e.g. users
    could discuss their profession, but also the discuss the profession of
    someone in the news, or even other professions or previous professions
    of the massage therapists)? How can we verify the skew of the forum
    userbase (e.g. %
    mongers come from a wide range
    of incomes, but likely those that participate in a niche forum will be
    the very active customers and are likely to have an income above the
    average monger population.)
  \end{itemize}

  \subsection{Concluding Remarks}
  
  Current legislation focuses on the massage therapists and businesses
  instead of the mongers. However, these laws can backfire and harm the
  women involved. The researchers hope that more attention to the monger
  population can lead to legislation that is more equitable and just. In
  particular, the 2018 Robert Kraft case shows the limitations of our
  existing policies and law enforcement and criminal justice frameworks
  around the IMI. Robert Kraft is a billionaire charged in 2018 with
  soliciting prostitution from massage parlors in Florida. Ultimately,
  all charges against Kraft were dropped after the video evidence was
  rendered inadmissible, and he walked away with no consequence. The
  justice system ultimately punished the workers instead. Our current
  legal frameworks focus on the employees and place requirements on the
  women to e.g. self-identify as victims, which IMI employees may be
  reluctant to do. Though the women involved were initially identified
  as human trafficking victims, they were in the end charged with
  prostitution, enduring not only shame but also thousands of dollars in
  fines and a criminal record. A better understanding of mongers can
  allow for laws that address the demand side instead.
  Despite the large size and pervasiveness of the illicit massage
  industry in the United States, the academic community knows very
  little about the IMI. We hope our work motivates and accelerates
  more research into the illicit massage industry.  
\graphicspath{{./figs/synthetic/}}
\captionsetup[figure]{hypcap=false} %

\begin{savedenv}[fig1]
   \keyfig[H]{lw=0.8,c={%
		Impact of varying the mean transaction time for NAgents from (left) noon to (right) 5 PM. The \# of agents and SAgent mean transaction time is kept fixed.%
        },
        l=fig:1%
        }{transactions_different_mean_hr.jpg}
\end{savedenv}

\begin{savedenv}[fig2]
   \keyfig[H]{lw=0.8,c={%
        Impact of varying the ratio of suspicious to normal accounts from 10:1, 100:1, to 1000:1. We fix the number of NAgents to 1000 (relatively small to minimize computation time).
        },
        l=fig:2}
        {varying_percent_suspicious.jpg}
\end{savedenv}

\begin{savedenv}[fig3]
   \keyfig[H]{lw=1,c={%
       Decision Tree with max depth n=1. Left: Decision tree. Right: Jitter plot of every transaction, split by the true label (of the sender), and then colored according to the predicted label (blue for normal, orange for suspicious). For a perfect classifier, the top half would be all blue and the bottom all orange.
       },
       l=fig:3}
       {DTree_depth_1.jpg}
\end{savedenv}

\begin{savedenv}[fig4]
   \keyfig[H]{lw=1,c={%
        Decision tree model with max depth n=2. Left: Decision thresholds.
        Right: Jitter plot of every transaction; refer to \Cref{fig:3} for explanation. The decision tree behaves better with a second layer (3 thresholds, at 11, 72.5, and 76.5 timesteps).
        },
        l=fig:4}
        {DTree_classification.jpg}
\end{savedenv}

\begin{savedenv}[fig5]
   \keyfig[H]{lw=1,c={%
        \color{black}Result of GMM unsupervised outlier detection, with n=2 components.  On the left, we graph all transactions as individual points on a jitterplot by time, split by class. On the right, we graph the same information (transactions per timestep) as a histogram. Note that the blue and orange bars are directly overlaid. For convenience we have provided a zoomed image (boxed in black) of the histogram of the GMM predictions at timestep $t=50$, where an example of the overlaid bars may be seen. \normalcolor
        },
        l=fig:5}
        {GMM_outlier_zoom.jpg}
\end{savedenv}

\begin{savedenv}[fig6]
   \keyfig[H]{lw=1,c={%
        Results of the Isolation Forest algorithm, an unsupervised
        classifier which detects outliers. Synthetic data generated 
        },
        l=fig:6}
        {isolation_outlier.jpg}
\end{savedenv}

\begin{savedenv}[fig7]
   \keyfig[H]{lw=1,c={%
    Pair plot of the synthetic data generated %
    The data is colored according to their true label. Note: Although histograms are more appropriate for the diagonal (due to discrete features such as \code{num_txns}), the KDE is shown for legibility.
        },
        l=fig:7}
        {graph_features_distribution.png}
\end{savedenv}

\begin{savedenv}[fig8]
   \keyfig[H]{lw=0.8,c={%
    Difference between the spring layout and tuned Kamadai-Kawai layout on the same dataset.
        },
        l=fig:8}
        {spring_vs_kk_layout.jpg}
\end{savedenv}

\begin{savedenv}[fig9]
   \keyfig[H]{lw=1,c={%
         Network structure of agents transacting within one-hour windows starting at midnight, 4 AM, 8 AM, 12 PM, 4 PM, and 10 PM. Nodes are labeled with their agent ID. Suspicious accounts (nodes 1010-1019) are highlighted in orange. The (rather illegible) labels include the aforementioned parameters as well as how many agents of each type are present in that window (out of 1000 NAgents and 10 SAgents), as well as the mean transaction time parameter (noon and 10 PM for NAgent and SAgent respectively). Note: we used the spiral layout for NetworkX on the noon sub-figure due to the density of points.  Loops are present as agents are allowed to transact with themselves.
         For the window starting at noon, we switched to a spiral layout due to the density of points.
     },
        l=fig:9}
        {24_hr_network.jpg}
\end{savedenv}

\begin{savedenv}[fig10]
   \keyfig[H]{lw=1,c={%
           Decision Tree with varying number of input features (note that we are now classifying agents, not transactions).  The feature based solely on \code{in_degree} reaches 100\% accuracy with one level, while the \code{out_degree} model requires more.},
        l=fig:10}
        {DTree_comparison_graph_features.jpg}
\end{savedenv}

\begin{savedenv}[fig11]
   \keyfig[W]{lw=0.4,c={%
        Each outlier detection model on four sets of features. Here mcc stands for Matthew's correlation coefficient.
        },
        l=fig:11}
        {outlier_detection_feature_importance.jpg}
\end{savedenv}

\chapter{Faking It: Modeling Transaction Data for Counter-Trafficking in the Financial Sector}
\chaptermark{Agent-Based Synthetic Financial Data}
\label{chap:synthetic}

\section{Introduction}

Financial institutions are legally required to invest in anti-money
laundering (AML) \cite{fedregisterDueDiligence2016} and fines can range in the hundreds of millions of dollars. Current AML algorithms tend to be rules-based, such as thresholding on simple features (in essence creating a decision tree) such as size of transaction \cite{richardsonRulesBasedFPRateMcKinsey2019}. The benefit of this method is that the decisions are easy to explain to investigators who can then provide input on the rules used. However, downsides include a massive amount of alerts leading to false positives \cite{richardsonRulesBasedFPRateMcKinsey2019}, as well as difficulty adapting to new trends and behaviors. %

As a result, the financial industry has begun adopting machine learning (ML) techniques and now routinely use machine learning for AML purposes.\footnote{Of 14 major North American banks, a McKinsey survey in 2022 \cite{doppalapudiMLGameChangerMcKinsey2022} found more than 10 institutions had begun adopting ML solutions.}

\color{black}
In this chapter, we investigate how generating synthetic data could help with AML in banking, specifically in the context of counter-human trafficking. We use agent-based modeling to generate our data.

We generate not just time-series table data but also the graph data associated with transactions, both of which are important for AML investigators. Prior work targets other illegal activity (such as credit card or account fraud) or aims to mimic either tabular or graph data separately. 

We additionally experiment with varying the parameters (a.k.a. the behavior assumptions of the agents) of our model and show the impact on the downstream task of outlier detection. We  outline how we could validate the synthetic data against real data. Finally, we look into generating more complex (heterogeneous) graph data and discuss directions for future work.

In the following sections, \notegreen{we consider both a simple model (where agents simply have one-sided transactions at different times) and a more complex model (where additionally there we vary the transaction probability by pairs of agent types). For each model,} we first describe our algorithmic choices, second describe what the synthetic data looks like, and third describe the results of a downstream task, namely outlier detection. We used AgentPy \cite{foramittiAgentPyLibraryJOSS2021} to implement
our models. We use common packages for the rest: Pandas \cite{pandas}
  (processing data), Scikit-Learn \cite{scikit} (machine learning models), and SciPy \cite{scipy}
  and NumPy \cite{numpy} (various calculations). Plotting uses Matplotlib \cite{matplotlib} and
  Seaborn \cite{seaborn}, and we use NetworkX \cite{networkx} for visualizing graphs. Data is stored using Python3's joblib library to pass data between scripts.
\normalcolor

\notegreen{
Synthetic datasets allow researchers to rapidly test new algorithms. This allows faster translation of research from academia to industry, as research teams in financial institutions can more easily evaluate new algorithms on toy datasets before investing the engineering effort to try such algorithms on real-life data. For instance, our ABM allows for varying ratios of fraudulent to normal transactions (or agents). If an algorithm fails to perform when the ratio of fraudulent transactions (or labeled fraudulent transactions) is too low, then industry scientists can know that if their ratio is that low, they should wait for further improvements to the algorithm.
}

\section{Background}
\subsection{Bank AML Workflow and Acronyms}
\label{sec:bank-aml-workflow-and-acronyms}
\color{black}
Banks face increasing regulatory pressure to ensure that they are not accomplices to terrorism or money laundering. One report found that fines for insufficient anti-money laundering infrastructure were reported as having totaled \$321 billion between 2008 and 2017 and included \$42 billion in a single year (2016) \cite{fineTotalsUNODCMcKinsey2019}. Additionally, 
because human trafficking involves large flows of money, almost inevitably that money will flow through the formal financial sector. As a result, over the last few years many banks, such as Scotiabank \cite{ScotiabankHT2021}, have taken the initiative to expand their anti-money laundering efforts to include fighting human trafficking and other crimes.
\normalcolor

The general workflow is data flowing from customer activity to data
scientist teams and then investigators, as described in \cite{capitalOneHowUseML2021}. After internal modeling and discussion, teams then flag accounts for manual investigation. Investigators then choose whether to
investigate or throw out flagged accounts. %
Manual investigation can then lead to account closure
and mandatory filing of Suspicious Activity Reports (SARs) with
government regulators. Bank tellers closer to the point of contact may
also file Unusual Transaction Reports (UTRs) as they conduct
due diligence, which generally falls under KYC (Know Your
Customer). The idea of KYC includes not just AML but also responding to
sanctions (e.g. against Iran), and determining whether the owners of a business fall under sanctions can be difficult.

\color{black}
Depending on the specific task, recall might be important (not detecting
fraudulent activity leading to severe fines, up to hundreds of millions
of dollars). Regulators must understand why accounts were
flagged. On the other hand, for other work (e.g. counter-human
trafficking), high precision (high confidence in a flagged accounts) may
be required due to limited investigator resources. Additionally,
heuristic-based flagging is becoming problematic as money-laundering
activity has been increasing over the past few years; some sources say
alerts can range to upwards of 98\% false positives \cite{richardsonRulesBasedFPRateMcKinsey2019}. Being able to
explain the machine learning results therefore critical for moving AML
investigators from heuristics to machine learning models.
\normalcolor

\section{Related Work}

Financial fraud and AML have long been a focus of academic research and
government agency attention. Below is a selection of related work for inspiration; many more papers exist.

\subsection{Agent-Based Modeling}

\color{black}
The idea behind agent-based models is that although the behavior of the individual agents may be straightforward (described by simple rules), the resulting system can still result in unexpected or counter-intuitive behavior. Famous examples include neighborhood segregation, where researchers (in 1971) showed that even if any given individual only had a preference that 30-50\% of their neighbors to be of the same type (otherwise they would move), as a group level this led to highly segregated neighborhoods  \cite{schelling1971neighborhoodsegregation}. Another commonly used example of ABMs is in wealth transfer.  Should the pool of wealth remain equal, if agents choose with random probability other agents to transfer all their money to, then eventually the inequality of wealth will follow a Boltzmann distribution, where most people will have 0 or 1 units of wealth and a lucky few will have 5 or more units of wealth \cite{angle1986abmwealthdistribution}. Additional examples of ABMs include modeling forest fires and flocks of birds. Interactive implementations of the aforementioned can be found at the AgentPy documentation.\footnote{ \url{https://agentpy.readthedocs.io/en/latest/model_library.html}}
\normalcolor

\subsection{Synthetic Data}
\label{sec:synthetic-data}

Synthetic data has a wide variety of approaches. Of special note, there
is the Synthetic Data Vault, which uses a suite of algorithms to
generate tabular data and relational data \cite{zhangSequentialModelsSyntheticDataVault2022}. They additionally
provide a set of metrics for measuring how well synthetic data compares
to real data (along many axes). However, these generators focus on
tabular data and do not feature an easy way to emulate how criminal
behaviors may shift over time. Additionally, there is no direct method
of simulating graph data, nor benchmarks and metrics for doing so.

Other work addresses generating and labeling using graph networks,
including investigating the impact of homophily assumptions (that is,
that connected nodes have the same label) and the implications for
fraud
detection \cite{pereiraHowEffectiveAreGNNBrazil2021}. We also look at how noise impacts the performance of
these datasets.\footnote{A framework for using injected noise in synthetic data to comparatively evaluate the robustness of fraud detection algorithms can be found in 
\cite{dosSantosLabelNoiseInjectionRobustnessFraudDetectionMS2020}}
Finally, \cite{darabiFrameworkLargeGraphDataGeneration2022} presents one recent approach
to generating graph networks which addresses scale issues and also
learns parameterized models that can be adjusted by researchers.

\subsection{Agent-Based Modeling for Synthetic Transactional Data}
\label{sec:agent-based-modeling}
The most direct inspiration for our work comes from PaySim \cite{paysim2016}, a 2016 implementation in Java of an ABModel that generated transactions according to mobile payments from a mobile provider. Upon investigation, unfortunately the details of the simulator were not clear (e.g. column names that seemed similar, assumptions that weren't obvious, and values that did not match the documentation). Additionally, network data cannot be created as no IDs are available for the agents; only tabular data is available at the end \cite{voutilaMobileFraudPaySimFork2020}.  \notegreen{In other words, Paysim only produces transactions \cite{paysim2016} without attaching them to individual agents, which means that classifiers working with the model are limited to classifying transactions and not entire accounts. Our model provides a way to model accounts in addition to transactions.}

This simulation is preceded by RetSim by the same authors which looks at fraud in retail stores, and operates on similar principles \cite{l-rojasSocialSimulationForFraudDectection2014}. Other work in this sphere looks at how to generalize from coarse behavioral models (such as the simple difference-of-Gaussian-means that we use in this chapter) to more detailed behavioral models \cite{wangFineGrainedBehaviorFraudOnlinePayment2022}.

\subsection{Publicly Released Datasets}
\label{sec:publicly-released-datasets}

Recent advances in privacy-preserving ways to release datasets have
filtered into financial AML research as well. One dataset, available on Kaggle (a machine learning competition website), features transactions from 2013 by European cardholders \cite{CreditCardFraudKaggle}. However, they only released composite features that are unnamed and generated from the underlying dataset. Since then, patterns of fraud are likely to have changed in response to regulatory pressure. Additionally, the authors of that paper released a simulator for the data \cite{TransactionDataSimulator}. This simulator includes parameters for the geographical distribution of merchants, but is implemented not in an agent-based fashion (the source code is still valuable however). A more recent release features a pipeline where machine learning models are used to generate tabular data to match opening account information from a bank with ground truth labels, as described in their paper \cite{jesusBiasedTabularDataBAFKaggleNeurips2022}. The input data has injected noise so that the model cannot actually match the real dataset. The output data is also filtered appropriately to e.g. ensure no actual data points are
present.

\section{Methods}

In the section we include details of how we used ABModels to generate synthetic data, including details for replicating our work. The codebase can be found at \url{https://github.com/nro-bot/fake-banking-data}. %

Our work includes two different ABModels: a simpler model that only creates
tabular data, and a more complex model that generates graph data and has
more parameters to tweak. We call these
\textbf{SimpleModel} and \textbf{GraphModel}. 

\color{black}
\subsection{Feature Selection: Transaction Time-of-Day}

Our model is based on reports from FinCEN, FINTRAC, and FATF, organizations defined as follows. FinCEN is the United States Financial Crimes Enforcement Network, part of the United States Treasury.  FINTRAC is the Financial Transactions and Reports Analysis Centre in Canada. And FATF stands for Financial Action Task Force, an independent inter-governmental agency that develops gold-standard  recommendations for the global financial sector's anti-money laundering and counter-terrorist financing policies.

We draw from the following sources, which we will refer to uniformly as advisories:

\begin{itemize}
    \item The 2014 FinCEN advisory "Guidance on Recognizing Activity that May be Associated with Human Smuggling and Human Trafficking — Financial Red Flags" \cite{FinCEN-2014-A008}
    \item The 2020 FinCEN updated advisory "Supplemental Advisory on Identifying and Reporting Human Trafficking and Related Activity" \cite{FinCEN-2020-A008}
    \item The 2016 FINTRAC operational alert "Indicators: The laundering of illicit proceeds from human trafficking for sexual exploitation" \cite{FINTRAC-Indicators-2016}
    \item The 2020 FINTRAC updated operation alert "Updated Indicators: Laundering of proceeds from human trafficking for sexual exploitation" \cite{FINTRAC-updated-2021}
    \item The 2018 FATF report "Financial Flows from Human Trafficking" (this specific report co-written with the Asia/Pacific Group on Money Laundering, a regional organization similar to the FATF) \cite{taskForceIntlFinancialFeaturesHT2018}
\end{itemize}

We emphasize here (as echoed in the advisories) that no single transaction or red flag can indicate trafficking, and these features must be used in context with other activity and information. In particular, all reports go through a manual review and reporting process.

\normalcolor

From these advisories, we consider the red flags for sex trafficking and define our agent-based model feature as\textbf{ cash transaction outside of normal business hours.} We chose this feature as being the simplest feature that would still demonstrate the usefulness of synthetic data in counter-trafficking. The details are as follows.

\subsection{Implementation Details} 

For both SimpleModel and GraphModel, we  defined two agent types, a Normal (NAgent) type and a Suspicious
(SAgent) type. We model the probability of transacting at a given time step as a Gaussian.

We chose that one 24 hr period covers 10 standard deviations from the mean (5 per half day), which covers 99.9999\% of the area under a Gaussian curve. Then we arbitrarily picked that all agents are assumed to average three transactions a day. Additionally, for ease of debugging, we also fix that each timestep is 15 minutes. We generated data for one 24-hour period. Thus there are 96 steps per model run. We fixed a seed (42) for the probability of transaction, and report results only on the one seed without cross-validation.

\subsection{Methods for Simple Model}

The SimpleModel mimics cash withdrawal and deposits, which involve only one account at a time. For this model, the input is the number of agents
of each type (normal and suspicious) as well as the mean transaction
hour of each.

\begin{itemize}
\item \textbf{Varying}: \code{mean_transaction_hours, number_agents_per_type}
\item \textbf{Fixed}: \code{mean_number_transactions=4, mins_per_step=15}
\end{itemize}

\subsubsection{Algorithm}

In terms of the implementation, we used \code{scipy.integrate.quad} to integrate
the area under the Gaussian curve, which we discretized into
96 pieces (a.k.a. total timesteps in the run). We thus have a table with
one value (probability of transaction) per 15 minutes (1 step). We shift
the distribution to match the mean.

\subsubsection{Inspection of Generated Data}

The following graphs show the number of transactions that occurred at
each timestep (summed across all agents), as described in the captions.
Data is generated in steps of 15 minutes, then resampled to graph in
one-hour increments.

\printsaved{fig1}

\textbf{\Cref{fig:1}} - In \Cref{fig:1}, note that in the right-most case, due to
the overlapping gaussians, the suspicious transactions are camouflaged
within the normal activity. The model is thus sensitive to the accuracy
of our estimates of agent behavior (that is, if the different means are
more similar than we thought).

\textbf{\Cref{fig:2}} - For \Cref{fig:2}, we see that for the most realistic
scenario of 0.1\% of accounts being suspicious, the suspicious transaction activity is hard to (visually) recognize from the distribution of transactions alone. This indicates that, \textbf{given the low rate of outliers, we may need a large number of total transactions to train an effective classifier}. %

\printsaved{fig2}

\subsection{Methods for Graph Model}
\label{sec:graph-model}

After completing the simple model, we added transactional data as well as several parameters that could be tweaked. The general idea is that there may be variations, such as: suspicious accounts may transact in smaller amounts but more frequently. They may also tend to transact with other suspicious accounts. We added the following
parameters to allow for testing assumptions about these behaviors:

\begin{itemize}
    \item \textbf{\code{agent_type_pair_probabilities}}: \\normal$\rightarrow$normal, normal$\rightarrow$suspicious, suspicious$\rightarrow$suspicious, suspicious$\rightarrow$normal
    \item \textbf{\code{mean_num_transactions}} (for each agent)
    \item \textbf{\code{mean_transaction_amounts}}\footnote{We used
        constant values when generating data for what is presented here as
    the \code{mean_txn_amts} is not fully implemented} (mean, for Gaussian
  distribution)
\end{itemize}

\subsubsection{Algorithm}

For this initial version, agents choose a partner uniformly at random
from the entire set of agents (both NAgents and SAgents). The main
implementation detail here is to pick a random generator for a overall
ABModel, which is then used to generate the seeds for each agent. Each
agent uses their seed to generate randomized transaction amounts. The
latter is generated at model instantiation as a lookup table for all 96
timesteps in our simulation.

\textbf{Parameters} - For the model run for data
for the later task of outlier detection, we used the following
parameters. (for normal and suspicious agents, respectively):

\begin{itemize}
\item  \textbf{1000 NAgents and 10 SAgents} (or 100:1 ratio).
\item Mean hours are \textbf{noon and 10PM}
\item Mean number of transactions are \textbf{4 and 10} respectively
\item Transaction amounts are same for both agent types: mean of \textbf{20}
  (dollars) and standard deviation of \textbf{5}. (Future work would
  allow this to vary)
\item For the pair probabilities, we chose \textbf{90\%/10\% and 70\%/30\%}
  for each agent type, encoding the behavioral rule that agents are more likely to transact with other agents of the same type.\footnote{In other words, if an NAgent decides to send money, it sends to another NAgent with probability 90\%. On the other hand, for SAgents, they transact with other SAgents with probability 70\%}
\end{itemize}

\subsubsection{Inspection of Generated Data}

With this model, for each agent, we have tabular features such as number
of transactions and mean transaction time-of-day, and graph features
such as the in- and out- degrees for each agent.

\printsaved{fig7}

\textbf{Tabular and Graph Features} - we ran the simulation according to
the parameters described at the beginning of \Cref{sec:graph-model}. We then calculated the aforementioned features and then plotted them against each other, resulting in the pair plot shown in \Cref{fig:7}. With the true classes highlighted, these features show clear clustering that should allow our outlier detection algorithms to perform very well.

\printsaved{fig8}

\textbf{Graph} - Out of curiosity, we also visualized the network structure. We visualized the graphs for one hour windows at a few times of day (0:00, 04:00, 08:00, 12:00, 16:00, and 22:00). That is, we graph only the agents that made transactions within that hour.\footnote{We do not include node or edge features. In the in the future, the width of the connecting lines could be scaled according to e.g. transaction amount}

\printsaved{fig9}

\textbf{Technical Note} - For visualization, from an implementation
perspective, at the earlier hours there were few enough agents
transacting with each other that there were many disconnected
components. The NetworkX random layout performed poorly, with many overlapping nodes. We experienced a similar issue with the spring layout, where nodes often overlapped, especially disconnected ones. After an online search, we found out that the Kamada-Kawai forced-directed layout built into NetworkX accepts a distance parameter, allowing any disconnected components to be manually placed far apart and not overlapping. This is shown in \Cref{fig:8}. 

\printsaved{fig10}

\textbf{Discussion on Graph} - The graph data in \Cref{fig:9} shows clearly the impact of the Gaussian and network parameters. There are few transactions at 5AM, which is far from the Gaussian means for either NAgent or SAgent. But at noon the activity is much higher. Furthermore, there are few NAgents transacting at 10PM or midnight, but there are several SAgents transacting. Additionally, setting SAgents to transact over twice as much as NAgents means that they may occasionally transact even at times relatively far from their mean.

\sectionmark{Methods}  %
\section{Downstream Evaluation: Outlier Detection}
\sectionmark{Outlier Detection}  %

\subsection{Simple Model Outliers}

For downstream tasks, we looked at both supervised learning
(classification) and unsupervised learning (outlier detection). For
supervised learning, we used the decision tree (DTree). For
unsupervised learning, we used the gaussian mixture model (GMM)
and isolation forest (IForest) algorithms.
Our rationale is as follows. For supervised learning, we chose to use DTrees, which are straightforward for humans to interpret. Alternatives include logistic regression, but given that there is a single feature, logistic regression will not do much better but will require more interpretation.

For unsupervised learning, we considered both the class imbalance and the underlying assumptions of our model. In AML, the fraudulent/criminal accounts are very few compared to real accounts (in fact, one of the use cases of synthetic data is data augmentation, which can help with machine learning model performance). The classic \emph{k}-means clustering algorithm is more suitable when classes are balanced. GMM should perform well given that the underlying model is Gaussian. Finally, the Isolation Forest model is widely used for outlier detection. It provides an anomaly score so that decisions are not binary.

\printsaved{fig3}

\textbf{Decision Tree} - A decision tree with a single level (max
depth=1), for a one-dimensional input features, produces a simple
threshold. In \Cref{fig:3} we see in the jitterplot that the decision tree (trained on the full labeled dataset) has a threshold at 72.5 timesteps, or about 6 PM.\footnote{A jitterplot is a scatterplot with data points jittered so as to not overlap each other as much.} 

\printsaved{fig4}

In \Cref{fig:4} we see that the decision tree with two levels (four thresholds) does better. We hypothesize this is \textbf{due to the wrapping we used when generating data}, where a mean of 10 PM (88 timesteps) means there will be some transactions at 2 AM (8 timesteps).

\printsaved{fig5}

\textbf{Gaussian Mixture Model} - As shown in \Cref{fig:5}, the GMM behaves
somewhat poorly, mislabeling many normal nodes as suspicious (false
positives). 
\color{black} On the left, we graph all transactions as individual points on a jitterplot by time, split by class. From there we see on the bottom left that almost all suspicious transactions are flagged, but on the top left we see that the GMM has flagged many normal transactions as well. On the right, we graph the same information (transactions per timestep) as a histogram. On the top, we see how the transactions were flagged by the GMM. We see that at the 60 second timestep, approximately half the transactions were mislabeled by the GMM (the 1 and 0 bars are overlaid; a zoomed window is provided boxed in black). This causes the GMM mean transaction time for suspicious transactions to be pulled to the left. The correct distribution of samples is provided in the bottom right of the figure. \normalcolor 
This does mean high recall -- almost all suspicious nodes
are captured. Experimenting with the \code{n_components} variable did show that at n=1 no transactions are correctly classified. On the other hand,
if n=3 the results are fairly poor as well (and require interpretation
as to which label belongs to which Gaussian). The bump of orange on the
right of the histogram plot on the right indicate that the estimated
mean is very shifted from the actual mean for the
SAgent's transactions.

\printsaved{fig6}

\textbf{Isolation Forest Model} - The isolation forest algorithm relies
on the idea that when data points are further apart, longer random forest trees are needed to classify them. The algorithm is thus an unsupervised learning approach. The results of the isolation forest outlier detection are shown in \Cref{fig:6}. We set the contamination parameter (expected proportion of outliers) to 0.1.\footnote{The contamination factor is actually closer to 0.01, that is one outlier to one hundred normal data points, so the performance here despite that is impressive.} The IForest performs similarly to GMM but does slightly better, with fewer false positives (compare to the mislabeled transactions in orange on the right side of \Cref{fig:5}). Additionally, the contamination parameter has a strong impact on results, which can be undesirable in real life conditions where only an estimate is available for what percentage of cases should be considered outliers.

\textbf{Conclusion} - Intuitively, for the SimpleModel, there is only one feature (time of the transaction), so the models do not have a lot of flexibility and thus cannot perform well. We did try adding an $X^2$ term so the models had more flexibility, the graphs were prettier but the outlier results did not improve much. We were surprised at the performance of the GMM, since the samples are drawn directly from a Gaussian distribution. However, the likely cause is our decision to limit the simulation to 24 hrs and wrap the values around. \notegreen{Our agent-based model can have a Gaussian centered at 10PM that outputs suspicious transactions at 2am. However, the GMM classifier cannot group the 2AM and 10PM transactions into a single mixture, and instead has a single mixture with a mean around 4PM.}

\subsection{Graph Model Outliers}

The earlier outlier detection around the single feature as a baseline to compare the more complex GraphModel against. Intuitively, the more features that vary between agent types, the easier it should be for algorithms to flag outliers. Thanks to the fact that the GraphModel version now synthesizes transaction data, we have graph features we can use as features into our machine learning models. (We leave algorithms that operate directly on the graph structure to future work). We also changed from classifying transactions to classifying \textbf{agent type}s directly. From the transaction network, we added the features \textbf{in-degree} and \textbf{out-degree.} These give the number of transactions sent and received from each agent.

\printsaved{fig11}

\textbf{Feature Sets} - For our analysis we looked at the following sets of features. (Note again these are per-agent features).

\begin{itemize}
    \item \textbf{Time data only}: mean transaction time (\code{txn_mean_time}) and number of transactions (\code{num_txns})
    \item \textbf{All features}: The above, with two more features: \code{in_degree} and \code{out_degree} (the number of incoming and outgoing transactions)
    \item \textbf{out\_degree}: Using a single feature. Per the pair plot in \Cref{fig:7}, we expect this to do more poorly than \code{in_degree}
    \item \textbf{in\_degree}: We expect this to do well but less well than using all four features
\end{itemize}

\textbf{Overall thoughts} - The overall idea behind \Cref{fig:11} is to investigate how much the addition of information about the network could assist with the downstream tasks. On the top row, we see on the left the results from using time data-related data, and on the right we see the results with the addition of more features (related to the graph). On the bottom we investigate how well the model did using a single graph feature as a baseline.

\textbf{Decision Tree} - As shown in \Cref{fig:11}, using all the features, the DTree is able to make an exact rule to classify the outliers. Of course, the DTree is likely overfit to the data (given there are only 10 outliers and there are 4 features) given that we are also providing labels to it. In comparison, the unsupervised approaches fare worse but are likely a more realistic approach.

\textbf{Gaussian Mixture Model} - This performed admirably using all
four features, which does better than just using time data, and also better than a single feature (\code{in-degree}), as expected. Surprisingly using a single feature, \code{out_degree}, performed best. This unexpected result was present for all three models.

\textbf{Isolation Forest} - This model performed well when the
contamination parameter was set correctly, it was fairly
sensitive to that.

\textbf{Note about out-degree} - Our initial hypothesis was that the \code{in_degree} would be a more powerful discriminatory feature than \code{out_degree}. Since the SAgents have a strong preference to transact with each other and they have many more transactions on average, the \code{in_degree} might correlate more strongly with suspicious or not than the \code{out_degree} of a node. However our results did not align with this hypothesis. 

\textbf{Discussion} - As a whole, the classifiers are able to demonstrate performance that is better than random and/or the null hypothesis of suggesting that all transactions are normal. Whether the results would hold if the data was split into training and test remains to be seen. Additionally, a larger number of samples may be critical given the small rate (1\%) of fraud -- and this is actually much higher than estimates from other papers.

\section{Discussion}
This agent-based model allows researchers a simple way to model transactions and create network data. \notegreen{
As true for any ABM, one limitation of our work is the 
plethora of assumptions needed. Adding each parameter
was simple but added up and complicated the
documentation.  More black-box synthetic generators can take real data to train on and create realistic data with less researcher time. On the other hand, if the ABM should output data that contradicts real-life data then researchers can say that either the assumptions are incorrect or that there are confounders. For instance, if existing cases do not corroborate suspicious transactions clumping at 10pm, it may be that another group of users also commonly transacts at that time (e.g. night shift workers) or that the behavior patterns of suspicious users have shifted.}
\normalcolor 

We use outlier detection as an example benchmark task enabled by the synthetic data generated by our ABM. In the Simple Model version of our ABM, since the agents transact according to a Gaussian distribution, the outlier detection task degenerates to the task of separating Gaussian mixtures. The conditions for this are well-known and a long line of statistical work addresses and gives provable guarantees for Gaussian mixture separability, including the line of work beginning with \cite{dasguptaLearningGMM1999}. However, fewer theoretical frameworks are given for evaluating algorithm performance on more heterogeneous data such as that given by our Graph Model, which includes network information such as who transacts with whom. In either case, the usefulness of the synthetic data remains unchanged. If a given anomaly detection algorithm does not perform well when applied to synthetic data, then it is unlikely to do well  under the noise of real-life data.

\subsection{Future Work}

For future work, in the short-term, we would add preferential attachment to the
transaction generator: specifically, it would not be realistic for
accounts to transact with others selected at random each time. Most of
the time, agents would transact with their "friends", or in this case an
initial set of nodes. This feature could make the simulation more
realistic.

We also mentioned that adding heterogeneous data as a feature that this
synthesizer easily allows for (and that is not present in future work),
e.g. accounts can have emails, phone numbers, or be business accounts
(with business names) and there can be both cash and other transactions.
Pursuing this further could be interesting, but could also devolve into
a very specialized code base that is hard to understand and modify to
suit a user's own needs.

This work was also motivated by excitement over newer graph-based
machine learning algorithms for semi-supervised labeling, followed by
disappointment about their performance under label sparsity. In the near
future we could implement label propagation as a downstream task and
investigate the results. Potential other work include developing a
streaming version of this simulator and concurrent algorithms, which
would give an interactive feeling to the simulation. We also put our code in an online
repository and plan to refactor and document the code for future
reference.

In the medium term the work is obvious: we would want to find a real-life
dataset and compare with our synthetic data (or modify our ABModel). In
particular, day-of-month information may likely be more manageable at
research-scale than time-of-day. We already implemented the two-sample
Kolmogorov-Smirnov test (using \code{scipy.stats}) in preparation for when we have real data to compare to.

\section{Conclusion}
\noteblue{
In this chapter, we investigated the possibility of using synthetic
data to augment research work in anti-money laundering (AML) in banks.
To do so, we created an agent-based model specific for our interests in
applying in counter-trafficking and successfully demonstrated the
possibility of extracting insights from the data. Although the model is
simplistic,
the results can still serve well as a communication and demonstration
tool. Although other works exist for synthesizing tabular data that could be used for generating networks (e.g. generating the edge list), we believe there is still value in exploring alternative approaches, and agent-based modeling is a promising approach.}\notegreen{The AML space is constantly evolving as fraudulent customers use increasingly sophisticated methods to evade detection. Combining forces between academia and industry and among banks can accelerate progress in fighting trafficking, and there is always room for more collaboration. Our ABM provides a peek at one such avenue for doing so.}

\graphicspath{{./figs/digger}}

\chapter{Seeing Touch, Part A: Digger Finger}
\label{chap:digger}
\chaptermark{Digger Finger}

\section{Introduction}

\newthought{Imagine you are at the beach} with a metal detector which goes off, and you stick your hands in the sand to find the metal object. Even though the granular media (sand) is constantly affecting your sense of touch on your fingers and palm, its acuity combined with your cognition enables you to easily find buried objects. \notered{Yet for robots, manipulating granular media is a challenging task. We take on this challenge} of using touch feedback to search for objects buried in granular media. A robot with such capabilities can prove useful in areas such as deep sea exploration \cite{nadeautactile}, mining, excavation, decommissioning explosive ordinates \cite{burtness2011evaluation, noauthor_t4_2019}, agricultural robotics and other areas where task dependant information can be occluded. In this paper we present a prototype of a tactile sensor that is designed to easily penetrate granular media and is equipped with the GelSight tactile sensor \cite{Johnson2009RetrographicSF, johnson2011microgeometry}. \notered{We call our sensor, shown in Figure \ref{real}), the Digger Finger sensor.} 

\textbf{Contributions}. \noteblue{The four major modifications that we make to previous GelSight sensors to adapt to searching for objects in granular media are }\notered{($i$) sensor shape (wedge-shaped), ($ii$) illumination source (red and green fluorescent paint instead of LEDs), ($iii$) addition of mechanical vibration and ($iv$) the type of gel (VHB double-sided tape)}. %

\keyfig[W]{lw=0.5, l=real, c={%
    \textit{Left:} Digger Finger. \textit{Middle:} Penetration motion. \textit{Right:} Tactile data showing zero contact, granular media (rice), and object contact.}
    }{digfin.png}

\textbf{Outline}. In the next section, we give a brief overview of the related work in the domain of robotic manipulation for granular media. We then present our technical approach in section \ref{approach} starting with the design of the sensor in section \ref{sec:sensordesign} and then the manufacturing process in section \ref{sec:manu}. Section \ref{experiments} describes the two experimental procedures we used to evaluate the performance of the Digger Finger in searching for objects buried inside granular media. Section \ref{sec:fluid} illustrates the ability of the Digger Finger to fluidize granular media during penetration. Section \ref{sec:object} illustrates the ability of the Digger Finger to identify objects that are buried inside granular media. Finally, in section \ref{sec:insights} we provide our concluding remarks. 

\color{black}
\section{Related Works}
\normalcolor

\color{black}

 Manipulating granular media, just like manipulating rigid or deformable objects, comes very naturally to us. For robots, however, it remains a challenging task. Historically, research in robotic manipulation has focused on rigid objects more than deformable objects and granular media. One reason is due to the difficulty of modeling the complex dynamics of both deformable objects and granular media. Additionally, the perceptual understanding of these objects and media via tactile based hardware devices and algorithms \noteblue{(as opposed to vision-based methods) is poor}. Yet, when dealing with physical interactions, tactile sensation can be more critical than visual information. Due to these compounding limitations \notered{in algorithmic modeling and sensor input}, robotic manipulation of deformable objects and granular media remains poorly explored.

Robotics research related to granular media has been fall primarily within the scope of automated operation of construction equipment such as scooping \cite{sarata2004trajectory}, legged locomotion \cite{li2013terradynamics, hauser2016friction}, gripper design\cite{brown2010universal}, manipulators \cite{cheng2012design}, haptic displays \cite{brown2020soft, stanley2013haptic} and in robotic pouring tasks \cite{yamaguchi2015pouring, matl2020inferring}, to name a few. In contrast, work on robotic manipulation of and within granular media has only recently begun receiving attention from the research community. In \cite{schenck2017learning}, the authors teach a robot how to scoop and dig into a pile of beans by learning the dynamics of the media from visual data. Building on \cite{schenck2017learning}, the authors in \cite{clarke2019robot} and \cite{clarke2018learning} learn to use tactile, visual and auditory feedback to estimate the flow and amount of granular materials during scooping and pouring tasks respectively. Regarding object identification in granular media in particular, two non-destructive methods that are popular for finding buried objects are Ground Penetrating Radars (GPR) and ultrasonic vibrations \cite{travassos2020artificial, martin2002ultrasonic}. These methods, though reliable, are only good at estimating rough geometry and approximate location of the buried objects. Closest to our work is \cite{jia2017multimodal} and \cite{syrymova2020vibro}. In \cite{jia2017multimodal} the authors use the BioTac sensors from SynTouch Inc. on a three fingered Barrett hand to detect contact with a cylinder fixed inside a bed of granular media. They take advantage of multi-modal sensory data to classify contact and no-contact events. Similarly in \cite{syrymova2020vibro} the authors again only classify the presence or absence of an object inside granular media. Unlike \cite{jia2017multimodal, syrymova2020vibro}, we use the high resolution data from the Digger Finger to identify different objects inside granular media. Also, our design enables deeper penetration in granular media with the help of mechanical vibrations.
\normalcolor

\notered{Our sensor falls within the GelSight family of sensors.}
\color{black}
The GelSight \cite{Johnson2009RetrographicSF, johnson2011microgeometry} is a vision-based tactile sensor \cite{abad2020visuotactile, shimonomura2019tactile}. It consists of a camera that observes an elastomer gel that is clear, directionally illuminated, and coated in a Lambertian or semi-specular membrane.  Objects pressed into the gel deform the surface, and the illumination allows estimation of surface normal along the deformation. The 3D geometry of the deformation can then be recovered using photometric stereo. Previous GelSight sensors had flat sensing surfaces \cite{Yuan2017GelSightHR, Donlon2018GelSlimAH}, which restricted their integration with manipulators to parallel grippers for planar contact interactions, while more recent work in curved Gelsight sensors can be found in \cite{Romero2020SoftRH}.
\normalcolor

\keyfig[H]{lw=0.25, wlw=0.4, l=CAD, c={%
Exploded view of the Digger Finger. Numbered arrows are 1. Bottom housing, 2. Mirror 3, Fluorescent paint, 4. Clear acrylic tube, 5. Gel, 6. Top housing, 7. Camera housing, 8. Blue LEDs, 9. PCB, 10. Vibrator housing, 11. Vibrator motor
} }{digfin_vibrator_CAD.png}

\normalcolor

\section{Technical Approach} \label{approach}

\subsection{Sensor Design} \label{sec:sensordesign}

The three main goals of our design were ($i$) enable the sensor to easily penetrate the granular media, ($ii$) provide rich tactile sensing to identify objects that are buried inside granular media. ($iii$) achieve human finger like form factor so that the sensor can easily be fitted on existing robot hands.

To do so, \noteblue{we keep two things from previous GelSight designs. First, we use a mirror as did \cite{Donlon2018GelSlimAH} to have a camera view that is perpendicular to the gel sensing surface, rather than looking at the surface at an angle. This allows us to have a slim device without worrying about self-occlusions of the sensing surface when contact is made. Second, we use the concept of light piping illustrated in \cite{Romero2020SoftRH} to have light rays pass across a curved surface with negligible loss in light intensity.} \notered{We also make four major modifications to previous GelSight sensors to achieve these goals: we make the sensor wedge-shaped, use fluorescent paint, add vibration, and use a different type of gel. The details are as follows.} First, we build on the prior work on creating curved GelSight sensors by \cite{Romero2020SoftRH} and add a pointed tip. Second, we replace two color LEDs (green and red) with fluorescent acrylic paint from Liquitex. This allows for a simple and compact design especially at the tip of the Digger Finger which has to face the brunt of digging. We use six blue color LEDs (3528) from Chazon to excite the fluorescent paints from the top of the Digger Finger (ref. Fig. \ref{CAD}). To excite the red and green color paint to the required intensity we shine an excessive amount of blue light into the Digger Finger. This causes the image to be dominated by the blue spectrum compared to the red and green parts. To compensate for this we use a small piece of yellow filter that we put on top of the image sensor of the camera by placing the filter inside the lens assembly. We also experimentally set the white balance setting of the camera to suit our needs. The camera that we use is from Arducam (SKU: B006603) which we interface with using Raspberry Pi 4. Third, this new design aids digging by fluidizing granular media using vibrations. For this we mount a high speed micro vibration motor (6$-$12\,$V$, 18000\,$rpm$) onto the Digger Finger. Fourth, we replace the silicone gel used in the previous GelSight sensors with a 3\,$mm$ wide and 1.5\,$mm$ thick double sided, transparent polyurethane tape (ZSHK Happy Cover HC06).

\subsection{Manufacturing Process} \label{sec:manu} %
The manufacturing process of the Digger Finger consists of several simple rapid prototyping techniques including 3D printing, laser cutting and spray painting, to name a few. The Digger Finger consists of a sensing module and a vibration module. The Digger Finger's sensing module's core is shaped like a cylindrical wedge. We use a piece of optically clear acrylic tube with inner diameter of 16\,$mm$ and  outer diameter of 22\,$mm$ (see Fig. \ref{CAD}). The tube is first diagonally cut using a band saw. The top and bottom surfaces of the cut tube are then sanded and polished until optically clear. The bottom surface of the tube is then painted with red and green fluorescent paint. Afterwards, we design and 3D print two custom housing (top and bottom) to house the necessary components of the Digger Finger (cameras, LEDs, mirror, gel boundaries). A thin piece of mirror is cut in shape of an ellipse using a laser cutter and attached to the center of the bottom housing using VHB double sided tape. Blue colored LEDs are soldered onto a custom designed printed circuit board (PCB) that is milled from a copper plate. This PCB is friction fit to the top housing along with the camera. The polyurethane double-sided tape that we use as the gel is brushed with aluminum flake powder, and then over coated with a thin layer of thermoplastic polyurethane (TPU) dissolved in a solvent. The gel is left to rest until the coating is dry, after which the uncoated side is applied across the clear acrylic tube. With the fluorescent paint and gel on the acrylic tube, we then assemble the top and bottom housing with the acrylic tube. This finishes the complete assembly of the sensing module of the Digger Finger. The vibration module is a 3D printed housing for the vibrator motor which is also cylindrical in shape. This module gets attached on top of the sensing module using screws and nuts. 

\section{Experiments and Results}
\label{experiments}

\subsection{Fluidizing Granular Media} \label{sec:fluid} %
The goal of this experiment is to study the effect of mechanical vibrations on the jamming of granular media. An immersed intruder moving in a granular media will experience strong forces because of the particle jamming effect. These forces are a function of several geometrical and material properties of the intruder and the granular media particles. It is well studied that the force chains formed among the granular particles during jamming can be weakened and the granular media fluidized by blowing air \cite{brzinski2010characterization, naclerio2018soft} or by mechanical vibrations \cite{firstbrook2017experimental, texier2017low}. Following \cite{texier2017low} we use a vibrator motor to study the relation between the force acting on the Digger Finger and its penetration depth in granular media, in the presence and absence of mechanical vibrations.  

\begin{figure}[ht]
\centering
\includegraphics[width=\textwidth, keepaspectratio, trim=0cm 0cm 0cm 0cm, clip]{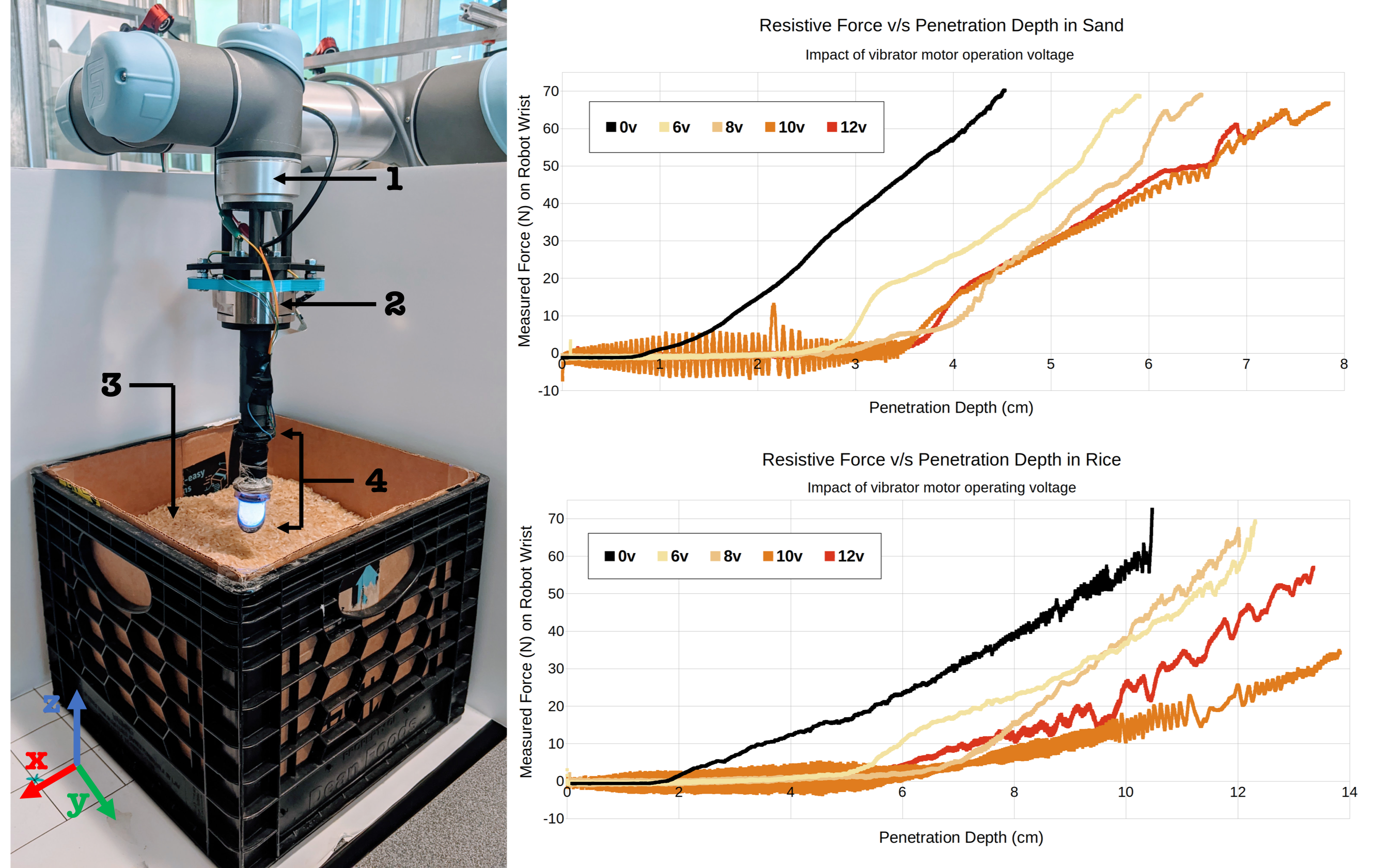}
\caption{\textit{Left:} Experimental setup. Arrows show 1. UR5 robot arm, 2. F/T sensor, 3. Granular media (rice), 4. Digger Finger \textit{Right:} As vibrator motor operating voltage increases, the force (measured by F/T sensor) required to move a given vertical distance in the granular media decreases.}
\label{exp_1}
\end{figure}

The experimental setup as shown in Figure \ref{exp_1} (left) consists of a robot arm (UR5) from Universal Robots Inc. The Digger Finger is coupled to the robot wrist using a 3D printed part. We use two types of granular media in our experiments, sand and rice. We choose these two particular media because of two very distinct effects they have on the object identification task, explained in the next section. For the sake of this experiment it is more important to note that both these media have bulk densities equal to \SI{1578.56}{kg/m^3} and \SI{941.48}{kg/m^3} respectively, resulting in different force versus distance relationships. Both the granular media are placed in a plastic container (12$\times$12$\times$\SI{11}{inches}) up to \SI{6}{inches} deep. We choose these container dimensions to reduce the \textit{edge effect} during penetration \cite{seguin2008influence}. We attach a 6-axis force-torque sensor (ATI Gamma) at the wrist of the robot arm to measure the ground truth forces acting on the Digger Finger during penetration. For each experimental trial the robot arm is first positioned such that the tip of the Digger Finger touches the surface of the granular media. The robot arm is then commanded to vertically penetrate the granular media at a speed of \SI{2}{mm/s}. We keep the velocity of penetration fixed for all trials. The robot arm is moved until it reaches the safety limit of its joints and stalls. We experimentally find that this distance, both in the case of sand and rice, is less than the depth of the container. In the first trial the vibrator motor is off. For each of the next trials we increase the operating voltage of the vibrator motor by 2\,$V$, starting from 6\,$V$ and going up to 12\,$V$, thereby increasing the vibrating frequency and amplitude of oscillation of the motor. Prior to each trial, the state of the granular media is reset by vigorously shaking the container and shoveling the granular media with a tool. For each trial the 3D position of the robot wrist and force from the ATI Gamma sensor are synchronously recorded. The force in the z-axis is first smoothed using an exponential filter ($\alpha=0.1$) and is then plotted against the $z$ position vector of the robot arm for each trial as shown in Figure \ref{exp_1} (right). It is evident from the figure that when the vibrator motor is off i.e. at 0\,$V$, the force on the Digger Finger begins to increase earlier in contrast to when the vibrations are on. This early rise in the force is approximately 2\,$cm$ in the sand case and \SI{4}{cm} in the rice case. Moreover, in the cases when the motor is on, all the curves are slightly less steep resulting in almost double the penetration at stall as compared to the case when the motor is off.

\keytab[W]{lw=0.5, c={%
Vibrating frequency and acceleration amplitude of the Digger Finger tip as a function of vibrator motor voltage.}, l=t1}{
    \singlespace
    \begin{tabular}{p{1.75cm} p{1.75cm} p{2cm}}
        \toprule
        Motor operating voltage ($V$) & Frequency ($Hz$) & Acceleration amplitude ($m/s^2$) \\ \midrule
        6        & 156              & 9.6        \\
        8        & 189              & 19.8       \\
        10       & 213              & 23.6       \\
        12       & 172              & 14.7       \\  \bottomrule
    \end{tabular}
    }

We quantify the vibrations from the vibrator motor at 6\,$V$, 8\,$V$, 10\,$V$ \& 12\,$V$ voltages in terms of frequency and acceleration amplitude using a 3-axis accelerometer. As the motor is rated to run at \SI{18000}{rpm} at 12\,$V$, theoretically the motor cannot vibrate more than 300\,$Hz$. So we use an accelerometer present in LSM9DS1 sensor chip \footnote{\url{https://www.sparkfun.com/products/13944}} that we can sample at a maximum frequency of 500\,$Hz$. The board is attached to the tip of Digger Finger using a double sided tape. The vibrator motor is run at the above specified four operating voltages and data from the accelerometer is recorded for five seconds. The fundamental frequency of the vibrations is found by running fast Fourier transform on the data. The average of the fundamental frequency over two trials for the four operating voltage is shown in Table \ref{t1}. The drop in the frequency and acceleration amplitude going from 10\,$V$ to 12\,$V$ is attributed to the resonance in the mechanical vibrations that we observe at 10\,$V$. Due to this resonance the Digger Finger vibrates with the largest amplitude at 10\,$V$. This is clearly visible in Figure \ref{exp_1}. While all the others curves are smoothed out with the same smoothing constant (i.e. $\alpha=0.1$), the curve at 10\,$V$ still shows periodic oscillation at certain sections of the curve, both for sand and rice, indicating the need for an $\alpha<0.1$ to obtain a smooth curve like the rest. We further verify the vibration frequency using the accelerometer on a smart phone using two Android  software applications. One (iDynamics\footnote{\url{https://www.bauing.uni-kl.de/en/sdt/idynamics/}}) analyzes mechanical vibration and the other (Spectroid\footnote{\url{https://download.cnet.com/Spectroi'.d/3000-20432_4-77833231.html}}) audio. We find that the fundamental frequency values for the four operating voltages range between 160--210\,$Hz$. The above observations clearly indicate that the Digger Finger is able to fluidize densely packed granular media with vibrations of approximately 150--200\,$Hz$ and 10-- 24\,$m/s^2$ acceleration amplitude, resulting in deeper penetration.

\subsection{Object Identification} \label{sec:object} 

\begin{figure}[htbp]
\centering
\includegraphics[width=0.65\textwidth, keepaspectratio, trim=0cm 0cm 0cm 0cm, clip]{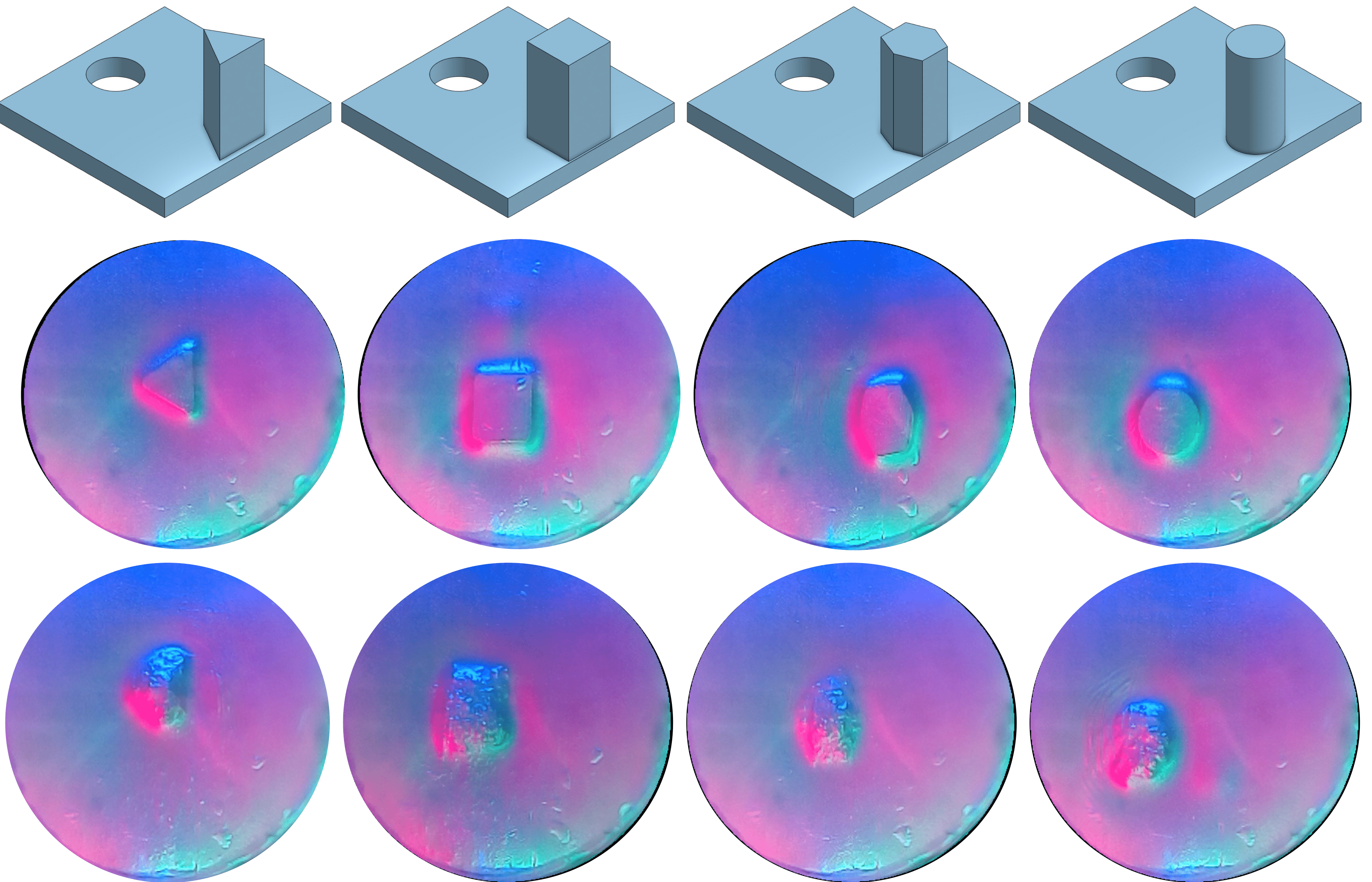}
\caption{\textit{Top row:} Design of buried 3D printed objects. \textit{Middle row:} Digger Finger image showing imprint of object shapes in absence of any granular media. \textit{Bottom row} Imprints of shapes in the presence of granular media (sand).}
\label{shapes}
\end{figure}

This experiment aims to study the performance of the Digger Finger in identifying objects in different granular media. For simplicity we narrow down the problem to classify four simple shapes i.e. triangle, square, hexagon and circle as shown in Figure \ref{shapes}, in two different types of granular media i.e. sand and rice. We choose rice and sand because both media exhibit very different behavior in the way their grains interact with the Digger Finger and the object shapes we are interested in identifying. Through ad hoc experiments with different granular media (washed sand, chia seeds, lentils, and mung beans) we observed that there are three distinct phenomenon that occur when a Digger Finger immersed in granular media approaches a stationary object, depending upon the geometry and material properties of the grains as well as the Digger Finger and the object. The first occurs when very small grains such as sand get permanently stuck between the Digger Finger and the object. The second occurs with grains (in our case rice) that are similar in size to the buried object, in which case the grains can completely block the object from coming in contact with the Digger Finger. The only way to come in contact with the object is to vibrate or twist the Digger Finger to push the grains to the side (refer to supplementary video\footnote{\url{https://sites.google.com/view/diggerfinger}}). We found twisting to be the most efficient way of pushing rice grains to the sides. Third case is when the material properties of the grains e.g. mung beans and lentils are such that they become slippery and do not get stuck between the Digger Finger and object. Based on this ad hoc experiment we narrow down the choice of granular medias to sand and rice. Figure \ref{shapes} shows the different 3D printed objects that we make use of to collect data for this experiment and their corresponding imprints on the Digger Finger as seen from the camera in the form of RGB images.   

For data collection we manually press the 3D printed objects on the Digger Finger and collect around 3000 images for each object shape. We repeat this procedure in a container filled with sand to collect images for cases when the sand grains get stuck between the Digger Finger. As shown in Figure \ref{shapes}, the sand grains distort the boundaries of the object shapes, potentially making them ambiguous. These two rounds of data collection result in a total of eight classes: four classes with just the object shapes and four with sand obscuring the objects. We add a ninth class for the  zero contact case. The dataset for the first eight classes are cleaned by manually removing images where the object is making no or partial contact with the Digger Finger. This data cleaning process leaves us with around 1500 images for each of the first eight classes.

For classification we rely on convolution neural networks because of their growing popularity for image analysis. In particular we train a residual neural network \cite{he2016deep} (ResNet50) on our data set by performing a popular technique in machine learning called transfer learning. Transfer learning focuses on storing knowledge gained while solving one problem and applying it to a different but related problem. For example, ResNet50 network is often first trained to classify real world images from ImageNet data set.\footnote{\url{http://www.image-net.org/}} We use this pre-trained network as the starting point and re-train it on our own object shape data set. We do this by only training the last convolutional block of the network and an uninitialized fully connected layers (classifier block) of size 128 at the end of the network. We split the 1500 images of each class into training (1200), validation (200) and testing (100) data sets. We augment the training data by randomly cropping and rotating the images. We also add Gaussian noise to each color channel. The network is trained for 10 epochs with a learning rate scheduler and batch size of 64, at the end of which it attains 99$\%$ training 98\% validation accuracy. The confusion matrix calculated on the test images is shown in Figure \ref{confusion}. 

\keyfig[H]{lw=0.6, c={%
            Confusion matrix for object identification task.},l=confusion }{cm_pytorch.png} 
            
\section{Conclusion and Future Work}
\label{sec:insights}

Identifying objects buried in granular media using tactile sensors is a challenging task. In our experiments we identify several difficulties. First, it is difficult to actually reach the object because the granular media will start to jam and prevent downward movement. Second, even when the tactile sensor can reach the object, the granular media particles tend to get stuck between the sensor and object, distorting the actual shape of the object. To tackle these challenges we present a novel tactile sensor that we call Digger Finger. We build on previous GelSight tactile sensor designs and introduce several innovations, including the use of red and green fluorescent paint and polyurethane tape as the gel, to design a compact wedge-shaped sensor. We design the Digger Finger to fluidize granular media during penetration using mechanical vibrations. We use the high resolution tactile sensing provided by the Digger Finger to successfully identify different object shapes even when distorted by granular media particles. 

For fluidizing granular media, we have only presented results for vertical penetration, but moving horizontally in granular media is also increasingly challenging at greater depths. Further experiments are required to understand the nature of vibration and robot arm motion needed to fluidize granular media while moving horizontally. For object identification a major issue is that over time the paint on the gel incurs wear causing noticeable artifacts in the RGB sensor image data. The tactile data after wear looks different than the data used to train the neural network for object identification. This causes the network to make false positive predictions, in particular for the zero contact class. The network may also make false predictions when the granular media touches the Digger Finger during free motion or when the gel wrinkles during penetration. Ideally, the network should be able predict the type of the granular media and also robustly predict zero contact. We therefore plan to train the network on additional classes of granular media and also experiment with better data augmentation techniques to make the network more robust to gel artifacts. We also plan to calibrate the Digger Finger in order to construct 3D geometry data from the raw RGB image data. In particular this would let us use algorithms trained on simulated 3D geometry data, as opposed to trying to simulate raw RGB sensor image which requires modeling the exact illumination of the Digger Finger. Using 3D geometry data would also allow object detection algorithms to better transfer to real world tactile data from multiple Digger Finger units. 

We believe that such a tactile sensor paves the way for the robotic manipulation within and of granular media, whether for everyday tasks such as scooping rice or litter, or for more industrial applications such as finding and inspecting buried cables and other objects.

\section{Acknowledgements}
This research was supported by the Toyota Research Institute, the Office of Naval Research (ONR) [N00014-18-1-2815], and the GentleMAN project of the Norwegian Research Council. We would like to thank Achu Wilson, Shaoxiong Wang, Sandra Liu, Yu She, Filipe Veiga, and Megha Tippur for insightful discussions. We also thank Siyuan Dong, Daolin Ma and Alberto Rodriguez for lending us the force-torque sensor.

\fboxsep=0mm%
\fboxrule=0.05pt%

\graphicspath{{./figs/fiducials/}}

\begin{savedenv}[fig_intro]
   \keyfig{lw=0.75,c={%
            Consumer webcams and printed fiducial markers can be used to create a six-axis force-torque sensor. We used four springs to build a platform free to move in all angular directions. We affixed two printed fiducials to the platform, and then aimed a consumer camera up at them. To the right, the camera view  reveals the tag location. The tags are glued to the light shield, which is removable, allowing for easy design changes. Note that cardstock, which was removed for picture clarity, was used to diffuse the LED and avoid overexposing the camera. Green bottle cap is for scale.
            },l=fig:intro%
        }{photo_sensor_intro.pdf}
\end{savedenv}

\begin{savedenv}[fig_diagram]
   \keyfig{lw=0.75,c={%
        The top row shows the viewpoint from the camera when different forces or torques are applied (the dotted gray line shows the center of the camera view). By tracking the movement of the fiducial(s), we can derive the force and torque exerted on the sensor.%
        },l=fig:diagram%
        }{movement.pdf}
    \vspace{10mm} %
\end{savedenv}

\begin{savedenv}[fig_JS_GUI]
   \keyfig[W]{lw=0.7,c={%
        Our prototype JavaScript-based interface (modified from the \texttt{Js-aruco} library example). In this way, sensor data can be read just by loading a webpage.%
        },
        l=fig:JS_GUI%
        }{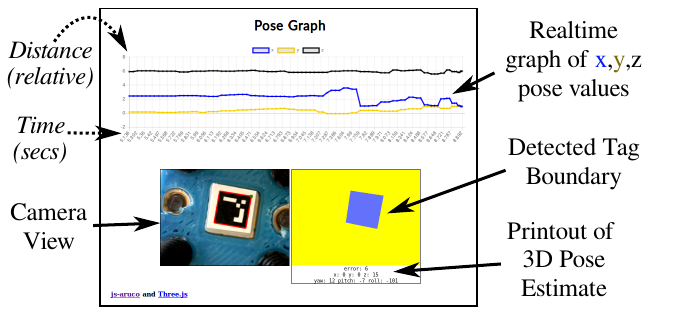}
\end{savedenv}

\begin{savedenv}[fig_pieces]
    \begin{keysubfigs}[H]{2}{c={%
    (a) Left, the four 3D printed parts are shown. (b) Right,  a diagram of the sensor as mounted to the commercial force sensor (in gray on the bottom) used in our experiments. The footprint of the sensor itself is the same as the camera circuit board, 35.7~mm by 22.5~mm. The sensor height $h$ = 51~mm, while the camera lens is approximately $d_{tag}$ = 21~mm from the center of the tags. The light shield is offset on all three sides by $w_{gap}$ = 2.5~mm gap from the camera cover, and has width $w_{shield}$ = 31~mm. The fiducials are each $w_{tag}$ = 3.8~mm wide (or $4.5$~mm including the white border). The mounting plate attaches to mounting holes in the force sensor and has width $w_{bottom}$ = 45~mm.%
    },l=fig:pieces} 
    \keyfig[W]{lw=0.8}{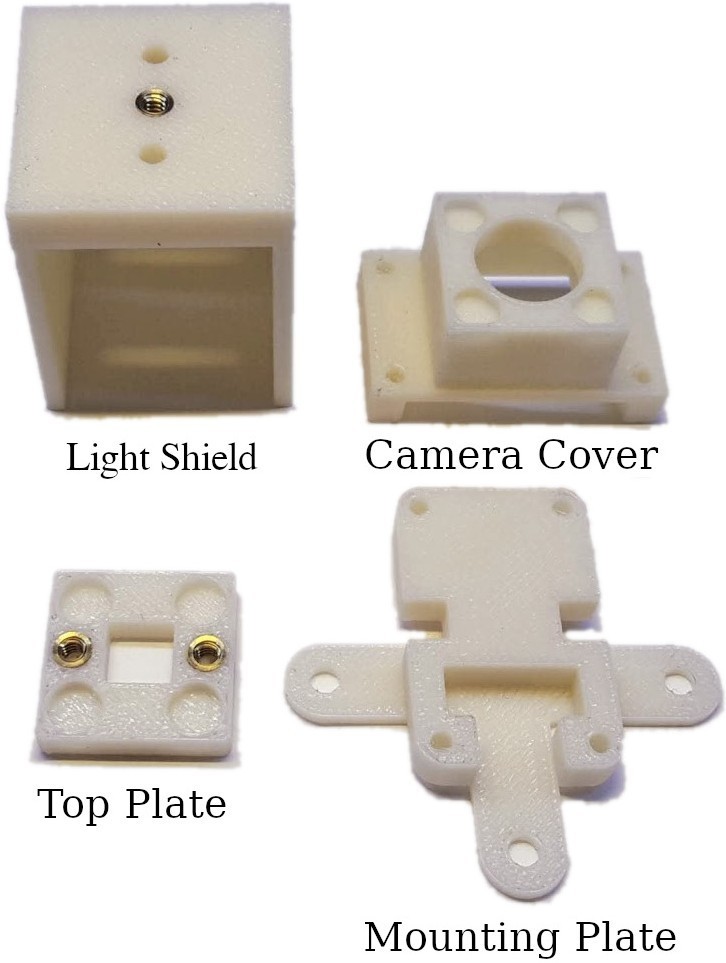}
    \keyfig[W]{lw=1,l=fig:sensorsolidworks}{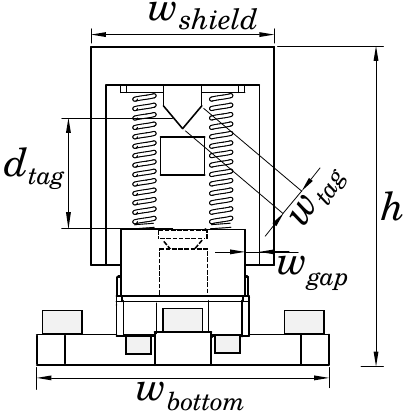}
    \end{keysubfigs}
\end{savedenv}

\begin{savedenv}[fig_xyz_res]
  \begin{keysubfigs}[H]{2}{c={Sensitivity calculation diagrams.}}
    \keyfig[W]{lw=1,c={%
    A frame from the camera. $w_{frame} \times h_{frame} = 640 \times 480$~pixels, and $w_{img} \times h_{img} = 150 \times 240$~pixels.
    },
    l=fig:xy_resolution}
    {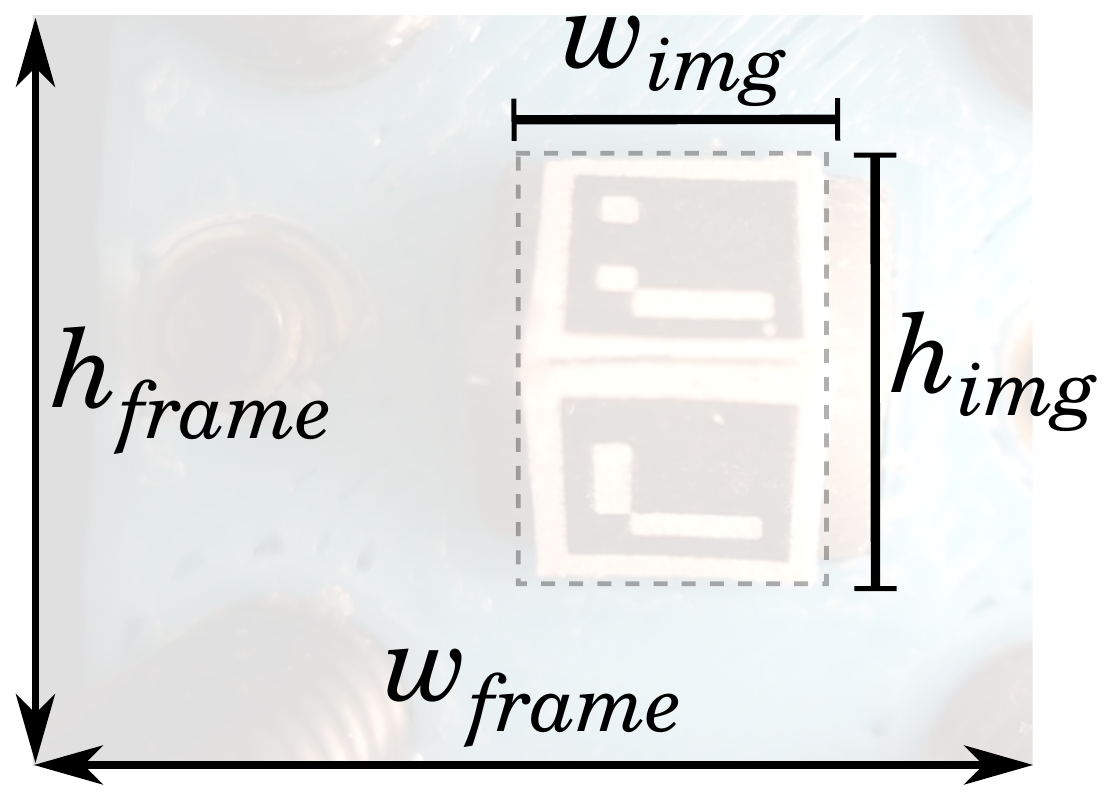}
    \keyfig[W]{lw=0.7,%
      c={%
        Side view. As per \cref{fig:pieces}, $d_{tag}=21$~mm and  $w_{tag}= 4.5$~mm. Light orange indicates original tag position before displacement. Inset shows force applied.%
      },%
      l=fig:zres}
      {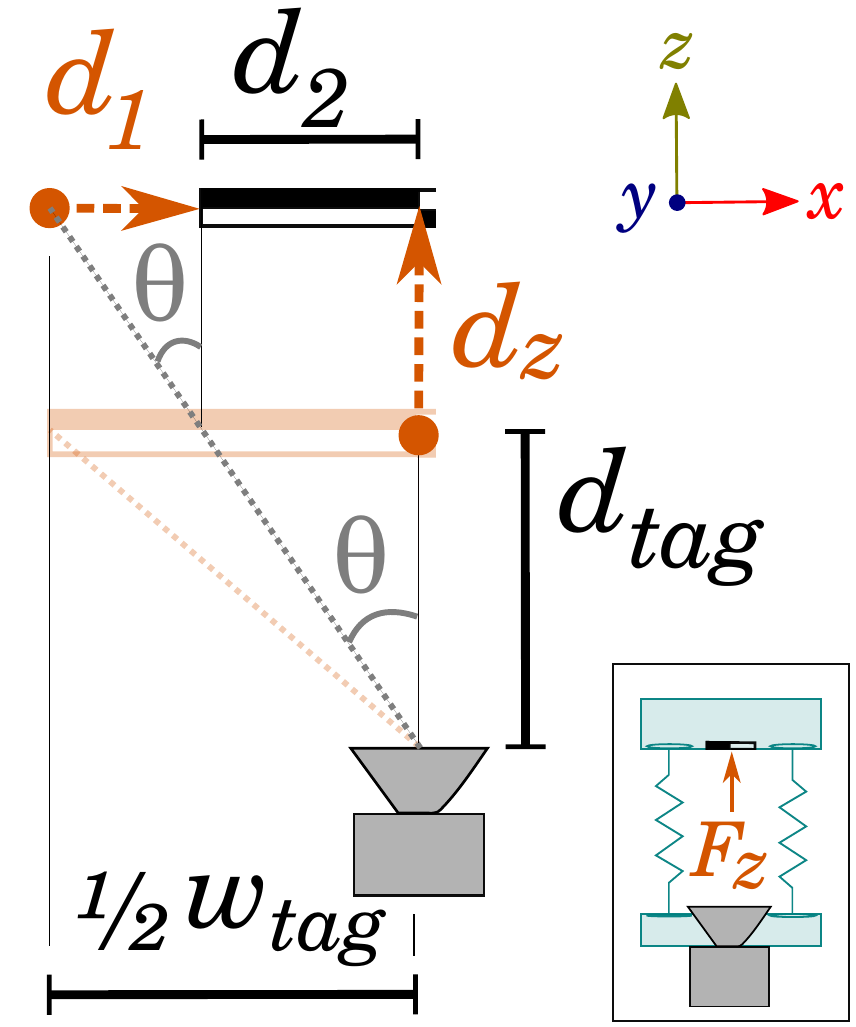}
  \end{keysubfigs}
\end{savedenv}

\begin{savedenv}[fig_tau_res]
\begin{keysubfigs}[H]{2}{%
    c=Sensitivity calculation diagrams. Insets show applied moment.}
    \keyfig[H]{lw=1,c={%
      Camera view. The light orange shows original tag orientation before a \ang{45} turn around the $z$ axis.%
      },
      l=fig:tauz_res}
      {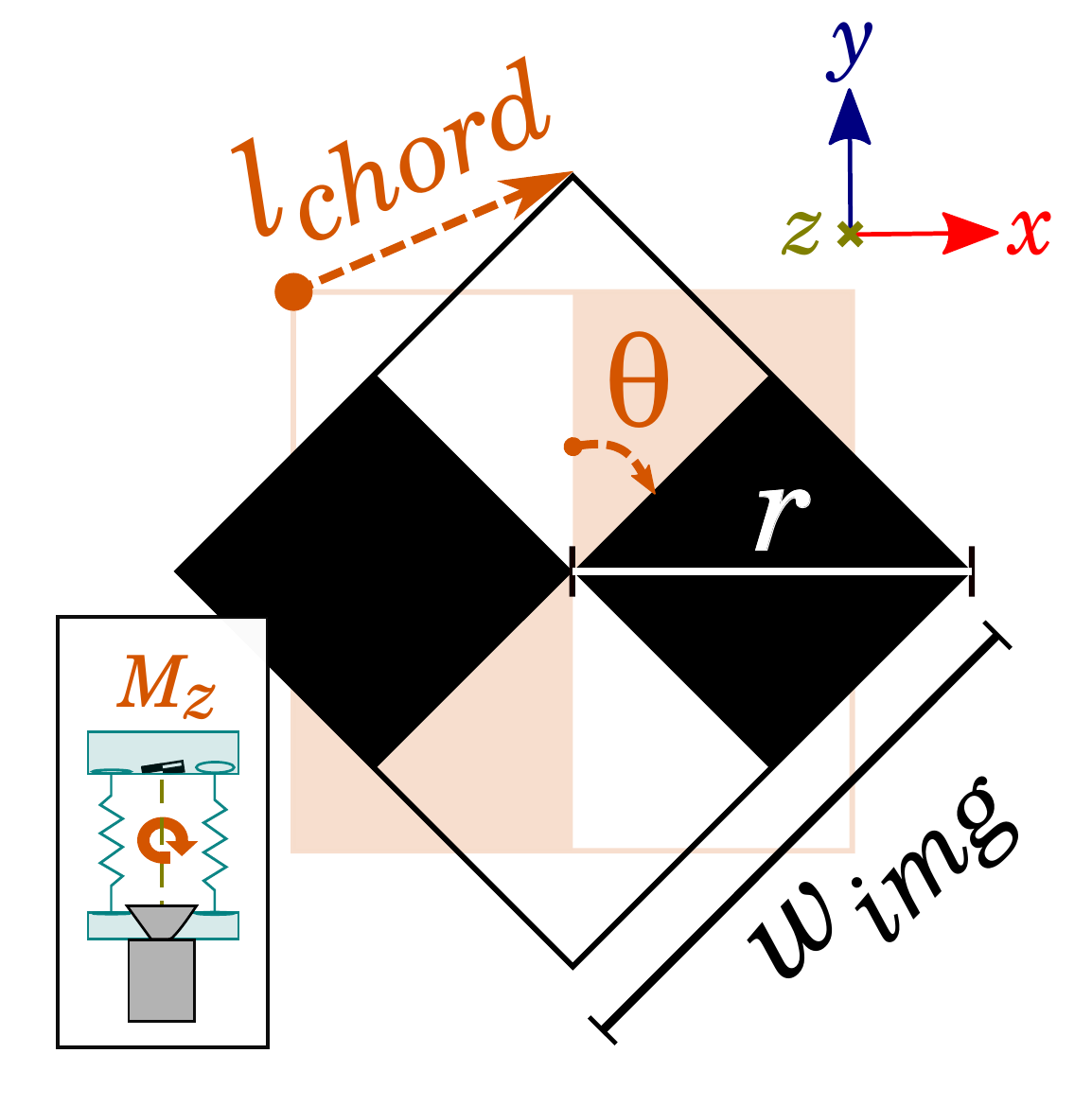}
    \keyfig[H]{lw=1,c={%
      Top-down view. The light orange shows original tag orientation before a \ang{45} turn around the $y$-axis (or equivalently $x$-axis).
      },l=fig:tauxy_res}
      {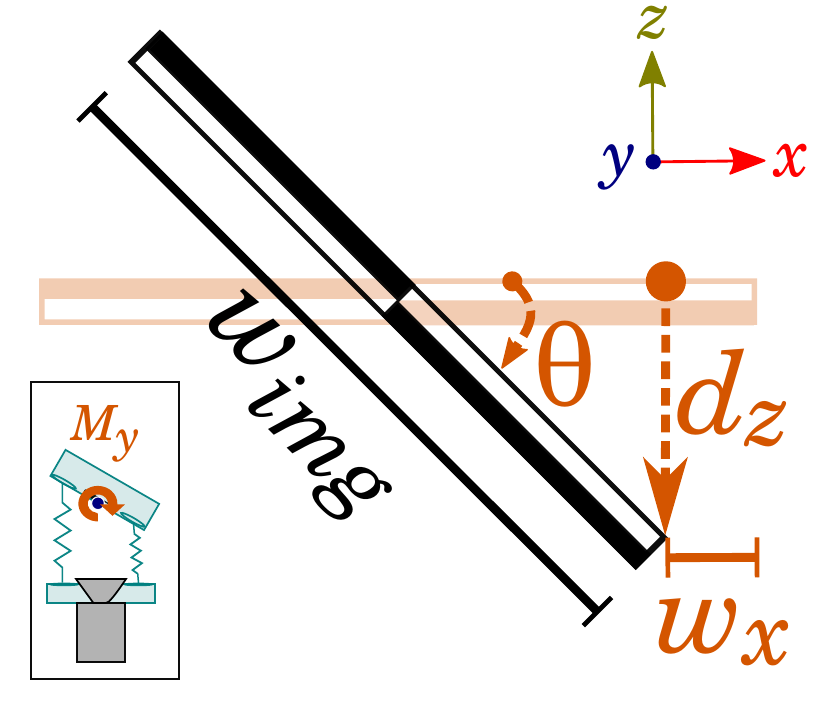}
\end{keysubfigs}
\end{savedenv}

\begin{savedenv}[fig_datacollect]
    \begin{keysubfigs}[H]{2}{c={%
      Left, the data collection setup is shown (with the LED off -- note that out-of-frame, there is an Arduino supplying $\SI{3.3}{V}$ to the LED. Later designs used a 3.3~V coin cell battery to make the sensor standalone). Right, a method to calibrate the sensor without using the commercial sensor is demonstrated. The sensor is mounted upside down and weights are hung by string from the sensor to apply force uniaxially to the +$z$ axis.%
      },
      l=fig:datacollect}
    \keyfig[W]{lw=1, c=Experimental setup.}
        {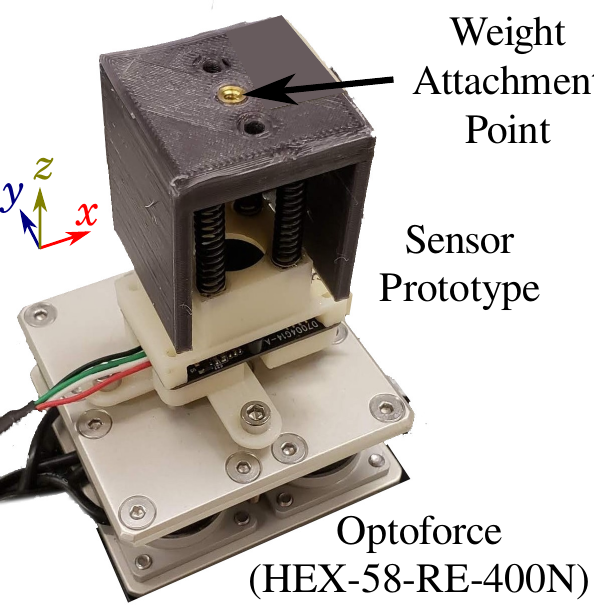}
    \keyfig[W]{lw=1, c=Calibration method.}
        {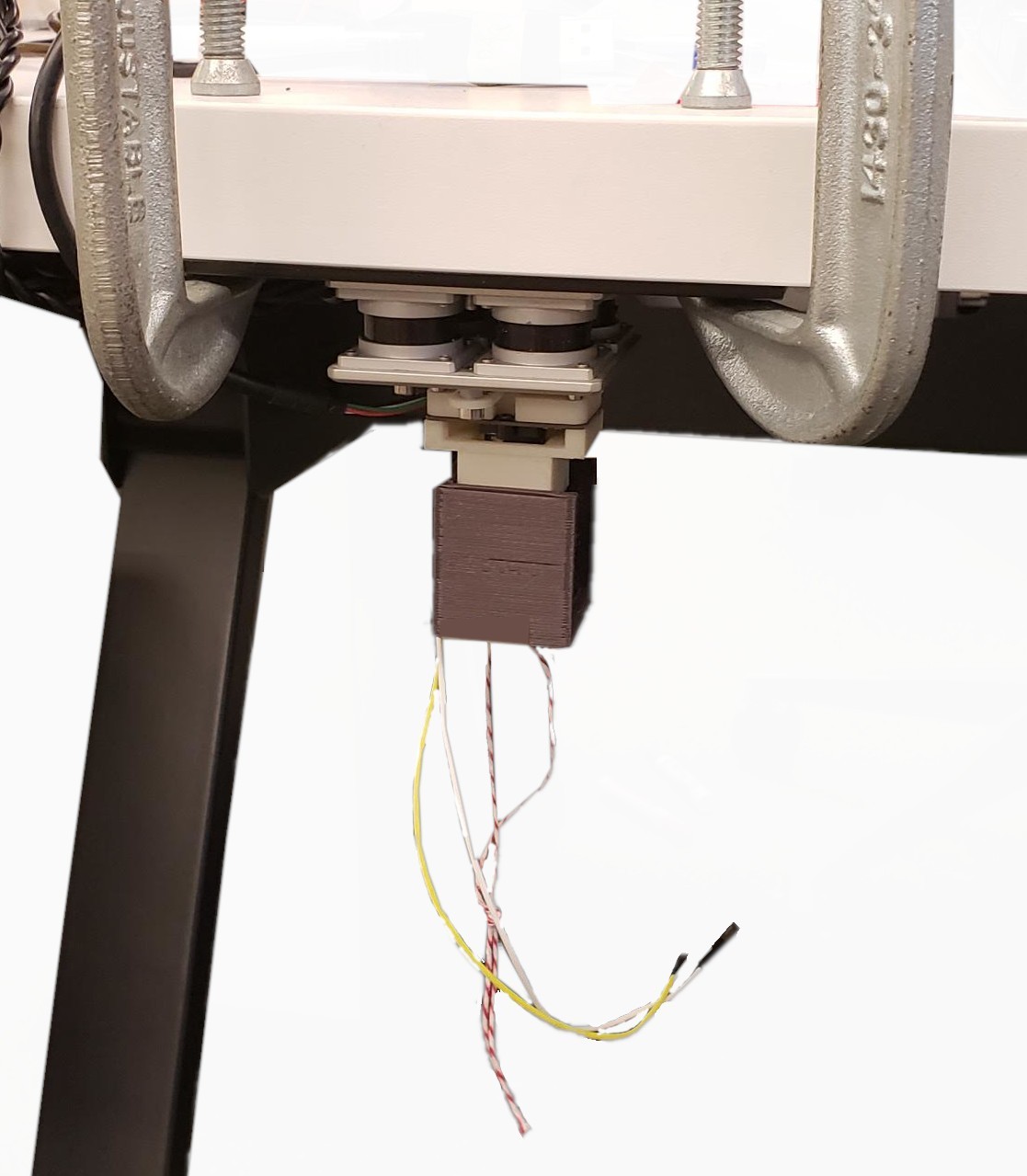}
    \end{keysubfigs}
\end{savedenv}

\begin{savedenv}[fig_linfit]
  \keyfig[H]{lw=1,
    c={%
    The black line represents a perfectly linear response between our sensor and the commercial sensor. The red dots show the actual sensor measurements using the ArUco tags.
    },l=fig:linfit}
    {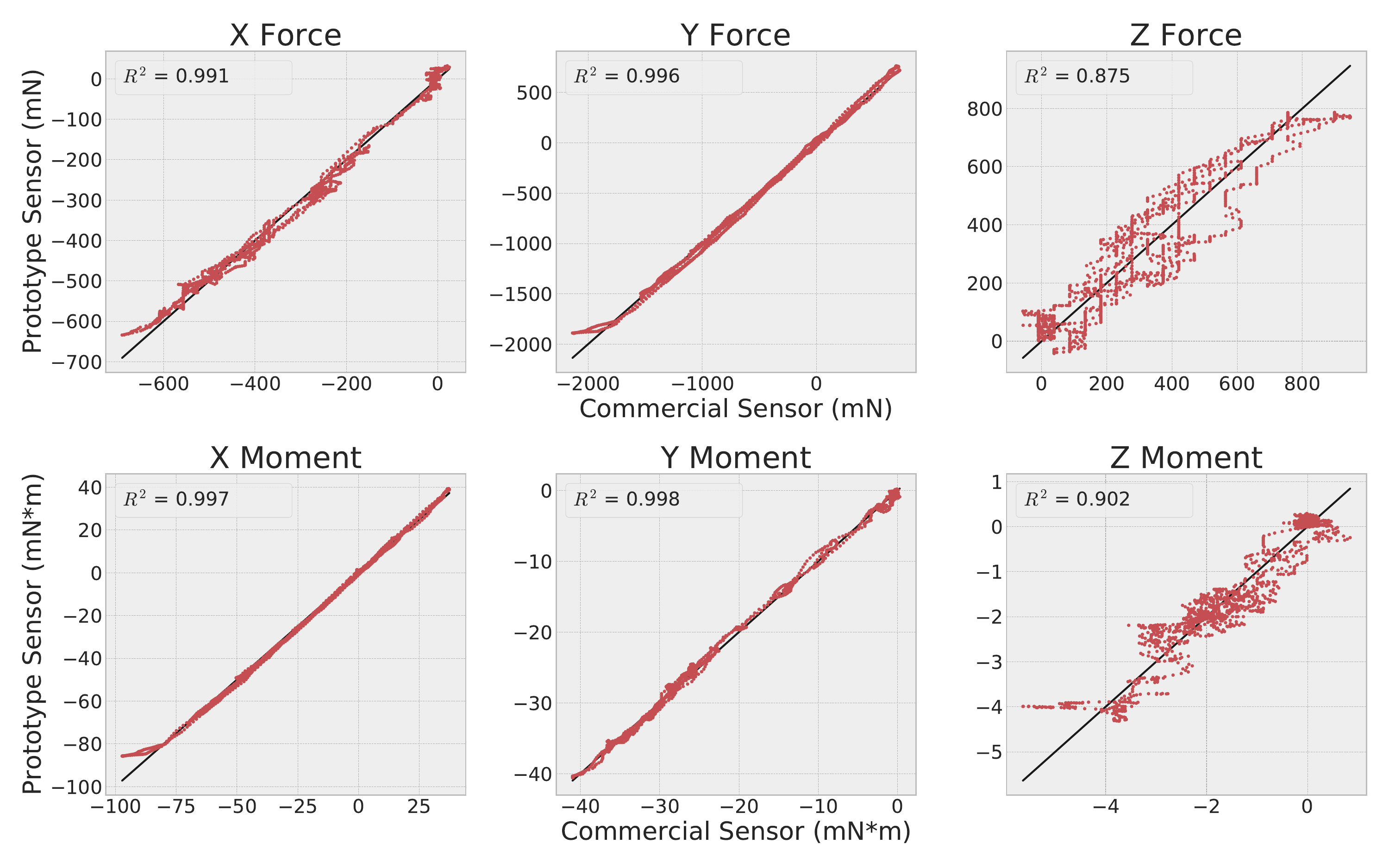}
\end{savedenv}

\chapter{Seeing Touch, Part B: Six-Axis Fiducial-Based Sensor}
\chaptermark{Fiducial 6-Axis F/T Sensor}
\label{chap:fiducial}

\section{Motivation}

    Force-torque sensors are used extensively in both industry and research.
    
 \printsaved{fig_intro}

    We focus here on the use of these sensors in two examples: robotic grasping, where they
    are used to provide tactile feedback (e.g. detecting when contact is made),
    and in human computer interaction.
    However, commercial six-axis force-torque sensors can be both expensive
    and fragile. This combination makes them tricky to use for grasping, where
    controlled contact is desired, but a small coding error could easily smash
    and overload the sensor. One of the most common types of sensors, the
    ATI force/torque sensor, costs tens of thousands of dollars and
    relies on strain gauges that are fragile and have to be surrounded in a
    bulky package.
    For these reasons, we are motivated to consider new sensor designs that could
    promote the use of tactile data in the robotics community through being a
    combination of cheaper, easier to use, and more robust.

\subsection{Related Work}

    Multiple designs have emerged recently taking advantage of the rich
    information available from consumer webcams. Even low-end webcams will
    output 640x480 RGB images at 15 frames-per-second (fps). The webcam-based
    sensors are particularly easy to manufacture and wire. Notable examples
    include the Gelsight \cite{Yuan2015MeasurementOS}, GelForce
    \cite{Sato2010FingerShapedGS}, TacTip \cite{WardCherrier2018TheTF}, the
    Fingervision \cite{Yamaguchi2016CombiningFV}, and others. These sensors rely
    on cameras facing markers embedded in transparent or semi-transparent
    elastomer (often with supplemental LED lighting). These can be used to
    estimate shear, slip, and force, but tend not to do well in cases where the
    object hits the side of the finger instead of dead on. They also require
    casting elastomers.

    Several MEMS multi-axis force-torque sensors have been developed, which use
    the same principle of creating a device free to deflect into multiple axes,
    but then measures them using capacitive \cite{Beyeler2009ASM} or
    piezoresistive \cite{estevez20126} means. In
    \cite{Cappelleri2009TwodimensionalV} the deflection is measured using a
    camera as well, a CCD camera mounted to a microscope, however the device
    only measures two directions of force.

    Prior work used MEMS barometers to create six-axis force-torque sensors
    with very low parts cost and good durability \cite{Guggenheim2017RobustAI}.
    However, fabricating the sensor requires specialized lab equipment such as a
    degassing machine.

    Other work explored estimating fingertip force via video, but only for human
    fingers \cite{yu2008, Sartison2018FingerGF}. Commercial sensors like
    the Spacemouse and the OptoForce use similar ideas, but rely on custom
    circuit boards for a ranging sensor inside. In contrast, our work is
    straightforward to fabricate even for users unfamiliar with electronics. 

\subsection{Contributions}

    In this paper, we investigate novel combinations of readily-accessible
    technologies to create six-axis force-torque sensors that are inexpensive,
    require minimal expertise to design and build, and are easily customized for
    diverse applications.

    The proposed novel type of sensor makes six-axis force-torque measurements by
    tracking position and orientation displacement using the 3D pose estimate
    from fiducial tags, and uses a linear fit between displacement and applied
    force-torque.  Fiducials are markers used to help locate objects or serve as
    points of reference. They can be found in robotics and augmented reality
    applications, where they usually take the form of printed paper markers
    glued onto various objects of interest. Sensors employing these fiducials
    operate by detecting the sharp gradients that are created between black and
    white pixels, such as those
    one might find on a checkerboard. An example of two fiducials can be found
    in the top right of the labeled diagram of our sensor at
    \Cref{fig:intro}. Using the known geometry of the tag (e.g.
    perpendicular sides of checkerboard), as well as known tag size and
    pre-determined camera calibration matrix, the 3D object pose (location and
    orientation) of the object can be estimated. This calculation is known as
    the solving the Perspective-\textit{n}-Point (PnP) problem.  We created prototypes
    utilizing two open-source tag protocols, AprilTags \cite{Olson2011AprilTagAR} and ArUco markers
    \cite{GarridoJurado2014AutomaticGA}; pictured in \Cref{fig:intro}
    are two ArUCo markers. 
    
    In the following sections, we begin with the design and fabrication process
    for our sensor. We follow with a theoretical analysis of how the
    sensor design parameters affect resolution, sensitivity, measurement range,
    and bandwidth. We also present an analysis of data collected from a
    prototype sensor. We conclude with a discussion of the advantages and
    limitations of this sensor.

    \section{Design}
    \subsection{Sensor Design}
    
    At a high level, the sensor consists of two main parts: a base and a platform above the base. The platform is connected to the base with 4 springs and can move in all directions with respect to the base. Two fiducial tags were
    glued to the underside of the platform. Then, a webcam pointed up at the
    tags was installed at the base. As force or torque is applied to the
    platform, the tags translate and rotate accordingly. The camera is used to 
    track the 3D pose of the tags. Should there be a suitably linear
    relationship between the displacement and the force-torque applied, 
    a short calibration procedure using known weights can be used to collect
    data points for regression. Given a known linear fit, the sensor can then output force and torque measurements. \cref{fig:diagram} shows the principle
    behind this fiducial-based force sensor.
    
\printsaved{fig_diagram}

    \subsection{Design Goals}
    \label{sec:designgoals}

    When designing the sensor prototype, a few considerations were made. First
    and foremost, the sensor needs to be sensitive to all six degrees of freedom
    (displacement in $x$, $y$, $z$ and rotation in yaw, pitch, roll). For
    illustrative purposes, the following analysis is performed in terms of
    specific specification values that are appropriate for a sample robot
    gripper. Alternate values for other use cases such as human-computer
    interfaces can be easily substituted. For grasping, between $\pm$ 40~N is
    realistic, and sensitivity of at least 1/10 N is desirable. Qualitatively,
    we want the sensor to be small (for grasping applications, the sensor should
    be roughly finger-sized), inexpensive, and robust. The sensor should allow for
    rapid prototyping and easy customization with minimal technical expertise.
    The sensor should be not only easy to fabricate, but also easy to use.

    \subsection{Fabrication}
    
\printsaved{fig_pieces}

    \subsubsection{Physical Fabrication}

    The four pieces in \cref{fig:pieces} (figure includes dimensions) are
    3D-printed in two to three hours on an inexpensive consumer-grade device
    (Select Mini V2, Monoprice). Epoxy
    is used to glue the springs into the camera cover and top plate. The tags
    are printed on paper and glued in. A small piece of white cardstock is used
    to diffuse the LED (in the future, this would be built into the 3D design).
    Conveniently, the pose estimate is
    relative to the camera frame, and the sensor relies only on relative
    measurements, so the tag placement can be imprecise. The LED is mounted 
    in and connected to a 3.3~V power source. The heat-set thread inserts (for bolting the
    light shield to the platform) are melted in with a soldering iron. The
    camera is placed between the mounting plate and camera
    cover and then everything is bolted together. The springs are steel
    compression springs available online as part of
    an assortment pack from Swordfish Tools. The spring dimensions are
    2.54~cm long, 0.475~cm wide, and wire width of 0.071~cm,
    with a stiffness of approximately $\SI{0.7}{N/mm}$.  Fabrication can be
    completed in a day. The actual assembly, given a complete set of hardware
    and tools, can be completed in 30 minutes, depending on the epoxy setting
    time.

    \subsubsection{Usage and Software}

    The only data cable used is the USB from the webcam to the computer.  
    On the computer, the OpenCV Python library \cite{opencv_library} (version
    4.1.2) is used to
    detect the ArUco markers in the video feed. We used a commercial
    force-torque sensor to characterize our sensor, for which we used another
    freely available Python library (see \cite{python_optoforce}). The data from
    the commercial sensor (Model HEX-58-RE-400N, OptoForce, Budapest, Hungary)
    and the markers are read in parallel threads and timestamped, then recorded
    to CSV. Python is used for further analysis.

\printsaved{fig_JS_GUI}

    By using a consumer webcam, sensor reading is also possible
    without installing Python. To demonstrate this, we developed a simple
    interface using a JavaScript ArUco tag detector library (see
    \cite{JS-aruco}). \cref{fig:JS_GUI} shows a graphical user interface (GUI)
    that plots the $x$, $y$, and $z$-axes of the 3D pose estimate for a single
    tag.

    In theory, the sensor reading can be done on-the-go with a smartphone and a wireless or USB-C webcam (such as inexpensive endoscope inspection cameras found online).

    \subsubsection{Calibration}

    Although we calibrated using a commercial force-torque sensor, the same can
    be achieved with a set of weights and careful clamping. The sensor can be
    clamped sideways to a sturdy surface to calibrate the $x$- and $y$-axes. A set of
    known weights is then attached to the center bolts on the light shield piece
    via a string. The same procedure can be applied to calibrate the $z$-axis,
    with the sensor clamping upside down to a tabletop. Finally, weights can be
    applied to the two side bolts to produce known torques while hanging upside
    down or sideways.

\section{Analysis}

    Considering the above design goals, there are a few primary concerns
    amenable to theoretical analysis: the sensor resolution, sensitivity,
    force range, and bandwidth. Here, sensor resolution is defined in bits
    (relative terms) and sensitivity in millimeters and degrees. 

    \subsection{Resolution}

    Let us conservatively estimate the discernible resolution of the tag
    system to be $d_R = 1/4$ pixel, or $C= 4$ counts per pixel. This factor exists because we
    have more than just binary information (1 bit) for every pixel. For
    instance, if a black / white intersection is halfway between two pixels, the
    pixels will be gray. (Tag algorithms also use the known grid geometry to
    achieve subpixel resolution -- see the \textit{cornerSubPix} function in the OpenCV
    library).

\printsaved{fig_xyz_res}

    In that case, we can determine the resolution of the sensor itself
    geometrically, by looking at the number of pixels. The fact that the tags must
    stay on-screen limits the sensor resolution. 
    
    We can characterize an approximate $y$-axis resolution $r_y$ of the camera by taking the
    number of pixels available, multiplying by $C$, and converting our counts into bits.
    \begin{align}
        r_y &= \lfloor \log_2 \left( C \cdot (h_{frame} - h_{img}) \right) \rfloor + 1
    \end{align}

    For instance, the calculations for our sensor prototype are as follows. In the $y$-axis,
    \begin{align}
        r_y &= \lfloor \log_2{\left(4 \cdot (480 - 240)\right)} \rfloor \\
        r_y &= 10 ~\text{bits}
    \end{align}

    In the $x$-axis, repeating the same calculations we have    
    \begin{align}
        r_x &= \lfloor \log_2 \left( C \cdot (w_{frame} - w_{img}) \right) \rfloor + 1 \\
        r_x &= \lfloor \log_2{\left(4 \cdot (640- 150)\right)} \rfloor + 1 \\
        r_x &= 11 ~\text{bits}
    \end{align}

    In the $z$-axis, our limitation is the same as the $y$-axis, so we have $r_z = 11 ~\text{bits}$.

    \subsection{Sensitivity}

    Let us now calculate the sensitivity of the sensor. We will start by
    looking at the minimum detectable travel in each of the $x$, $y$, and $z$-axes.

    \subsubsection{Translational Sensitivity}

    In the $x$ and $y$ directions, we can measure the mm/px at rest (the sensor
    resolution varies a bit since the tag gets larger or smaller depending on
    the $z$ distance). Roughly, the tag measures $\SI{4.5}{mm}$ and appears as
    $w_{tag} = 150$~pixels in the image. Assuming as above that we can discern 4
    counts per pixel, the theoretical sensitivity is
    \begin{align}
        s_y &= \frac{h_{frame} ~\SI{}{(mm)}}{h_{frame} ~\SI{}{(px)}} d_R 
            = \frac{4.5}{150} \cdot \frac{1}{4} 
            = \SI{0.0075}{mm} 
    \end{align}

    For the $z$-axis sensitivity, we consider that the tag will get smaller as
    it displaces in the $+z$ direction. Using a simple geometrical model (see
    \cref{fig:zres}), given that the smallest detectable change in $xy$ plane is
    $1/4$~pixel, we can calculate what is the resulting change in $z$. 

    Using similar triangles, we see that 
    \begin{align}
        \frac{d_1}{d_z} &= \frac{d_2}{d_{tag}} \\
        d_1 + d_2 &= w_{img}/2 \\
        d_2 &= (w_{img}/2) - d_1
    \end{align}

    We would like to work in $\SI{}{mm}$, therefore we use the fact that the tag is
    4.5~mm and appears as 150~px.
    \begin{align}
        d_1 = d_R = 1/4 \; \text{px} \cdot \frac{\SI{4.5}{mm}}{\SI{150}{px}} 
        &= \SI{0.0075}{mm} \\
        d_2 = \frac{4.5}{2} - 0.0075 &= \SI{2.2425}{mm} \\
        s_z = d_z = \frac{d_1}{d_2} d_{tag} = \frac{0.0075}{2.2425} \cdot 21 &= \SI{0.07}{mm}
    \end{align}

    \subsubsection{Rotational Sensitivity}

    For rotation about the $z$ axis, we can calculate the chord length in pixels
    traveled when a tag is rotated 45~degrees (about its center), and use the
    same assumption of four counts per pixel to estimate our rotational
    sensitivity. Geometrically, we know that
    \begin{align}
        l_{chord} &= 2 ~r \sin\frac{\theta}{2} 
    \end{align}

    In our case, with $w_{img} = \SI{150}{px}$, we see that
    \begin{align}
        \mathit{r} &= \sqrt{2} \cdot w_{img} /2 \\
        l_{chord} &= 2 \sqrt{2} \cdot 150/2 \cdot \sin \frac{\pi /4}{2} = \SI{81.18}{px} \\
        s_{\tau z} &= \frac{\theta}{l_{chord}} \cdot d_R 
                    = \frac{81.18}{150} \cdot \frac{1}{4} 
                    = \ang{0.14}
    \end{align}

\printsaved{fig_tau_res}

    For rotation about the $x$ and $y$-axes, the analysis becomes a matter of
    determining the $z$-axis change in mm, and using that to determine the
    pixels changed in the $x$-$y$ plane.
    Consider a 45~degree rotation around the $z$-axis of a tag that starts out flat
    (facing the camera), as shown in \cref{fig:tauxy_res}. Using $w_{img} =
    \SI{150}{px}$ as before, the $z$ sensitivity is as follows:
    \begin{align}
        w_{img}/2 &= \sqrt{2} \cdot d_z \\
        d_z + w_x &= w_{img}/2 \\
        w_x &= 0.5 \; w_{img} - \frac{0.5 \; w_{img}}{\sqrt{2}} = \SI{21.97}{px} \\
    s_{\tau xy} &= \frac{\theta}{w_x} d_R 
    = \frac{\ang{45}}{21.97} \cdot \frac{1}{4} = \ang{0.51}
    \end{align}

    \subsection{Notes on \textit{z}-axis measurements}
    \label{sec:zres}

    Intuitively, we expect that the sensor is much less reliable in the $z$ 
    displacement direction. For movement along the $x$ and $y$-axes
    axes, the camera sees the entire set of black/white intersections moving
    left or right.  %
    
    For the same reason, in the single tag setup it would be easy to
    detect rotations about the $z$-axis, and difficult to detect rotations around
    the $x$ and $y$-axes. Data collected from this initial (single-tag) design
    exactly reflected the aforementioned issue. Consequently, the design was
    enhanced with two tags oriented at 45 degrees to the camera. This proved
    sufficient for recovering all six force/torque axes.

    \subsection{Force Range Versus Sensitivity}
    \label{sec:40Ncalc}

    There is a clear trade-off between sensitivity (minimum detectable change in
    force) and the maximum force range. As an example, for a%
    desired force range $F_{range} = \pm\SI{1}{N} = \SI{2}{N}$ (close to the observed force
    range for our prototype), and a maximum
    displacement of $y_{range} = h_{frame} - h_{img}$, 
    the $y$ sensitivity $s_y$ in Newtons is as follows.
    \begin{align}
        s_{y} &= \frac{F_{range}}{ C \cdot  y_{range}} 
                = \frac{2}{4 \cdot (480 - 240)} = \SI{0.0021}{N}
    \label{eqn:maxdisplacement}
    \end{align}

    Our $s_y$ is thus 2.1~mN (given our assumption of $d_R = 0.25$). %
    Similarly, for the $x$-axis we find a
    sensitivity $s_x$ = $\SI{1.0}{mN}$ at this force range.  Now consider
    instead the grasping use case, with a desired force range of $\pm$ 40~N, and
    desired sensitivity of at least 0.1~N. If we scale the calculations in
    \cref{eqn:maxdisplacement} by
    40 to get a $\pm$ 40~N force range while keeping the other parameters the
    same, the sensor has 0.04~N and 0.08~N sensitivities in the $x$ %
    and $y$ directions respectively.
    
\section{Sensor Prototype Evaluation}

    \subsection{Linearity}

\printsaved{fig_datacollect} 

    In order to evaluate the linearity (and therefore usefulness) of the sensor, we used a commercial
    force-torque sensor (Model HEX-58-RE-400N, OptoForce, Budapest, Hungary) to
    provide ground truth measurements. Although the OptoForce measures force and
    torque at a different origin than where the load is applied,
    the analysis of the linearity of the sensor holds. Data was collected with a
    Python script which used the OpenCV library to interface with the camera.
    The setup is shown in \cref{fig:datacollect}.

    Autocorrelation was used to determine the lag between our sensor and the
    OptoForce. The sensor lag between the prototype sensor and the OptoForce was
    roughly 40~milliseconds. Next, linear interpolation was used to match our
    sensor data with the OptoForce data, which were output at roughly 25~Hz and
    125~Hz respectively. The sensor data was smoothed with an
    exponential filter with weight of 0.2 to improve the autocorrelation results.

    For calibration, we take a dataset of displacements $D$ and apply linear
    regression (with an affine term) against all six axes.
    $\theta$, $\phi$, and $\gamma$ refer to rotation around the $x$, $y$, and
    $z$ axes respectively. $K$ then forms a 6-by-6 matrix as shown below.
    \begin{align}
        \begin{bmatrix}
            F_x \\ F_y \\ F_z \\ M_x \\ M_y \\ M_z
        \end{bmatrix}
        =
        \left[
        \begin{array}{cccccc}
            \cr \\
            & & K_{6\times 6} & & & \\
            \cr \\
        \end{array}
        \right]
        \begin{bmatrix}
            D_x \\ D_y \\ D_z \\ D_{\theta} \\ D_{\phi} \\ D_{\gamma}
        \end{bmatrix}
        +
        \begin{bmatrix}
            \cr \\
            B\\
            \cr \\
        \end{bmatrix}
    \end{align}

    \subsection{Bandwidth}
    \label{sec:fps}

    Sensor bandwidth is directly limited by the camera framerate.  This must be
    physically measured since the Python script will output at unrealistically
    high framerate -- the OpenCV library reads from a buffer of stale images and
    will return a result even if the camera has not
    physically delivered a new frame.
    The webcam is pointed at a display with a high refresh rate. A script turns
    the screen black, and as soon as the camera detects the black color, the
    screen changes to white, and so forth, and the frames displayed is compared
    to system time to obtain the framerate of the webcam. 

    Note that this calculates our maximum sensor bandwidth; our actual sensor
    bandwidth is determined by the tag detection rate. If dynamic
    instead of quasi-static loading is assumed, then motion blur can lead to tag
    detection failure.

    \section{Results}

    \subsection{Linearity}

    In multiaxial loading, the sensor was manually moved around in all
    directions. As shown in \cref{fig:linfit}, the fits had a $R^2$ of 0.991,
    0.996, 0.875, 0.997, 0.997, and 0.902 for the $F_x, F_y, F_z, M_x, M_y$, and
    $M_z$ axes respectively.  The $F_z$ axis fit is notably worse than the $F_x$
    and $F_y$ fits, which was expected as explained in \cref{sec:zres}. 

\printsaved{fig_linfit}

\noindent
    For qualitative comparison, \cref{fig:linfit} shows an example of a reconstructed dataset, where the linear fits are plotted
    against the original signal for qualitative comparison. This diagram shows
    the relatively large deviations in $F_z$ from the original signal, indicating
    noisiness in the tag measurements.

    \subsection{Bandwidth}

    Our maximum sensor bandwidth is experimentally
    determined to be 25~Hz. Additionally, the camera we used was one of three cameras bought 
    by selecting for low cost, quick availability, and lack of external camera
    case. We also measured the other two cameras which, despite advertising
    similar framerates, exhibited noticeable differences in framerate. Operating
    at 640x480, we measured 25 fps, 33 fps, and 15 fps for the three cameras, as
    listed in \cref{tbl:camera}.     %
    \begin{table}[htbp]
        \caption{Camera Specifications}
        \label{tbl:camera}
        \centering
                \begin{tabular}{@{}llllll@{}}
                    \toprule
                    Camera Module Name    & Nominal Max Res. & Price & Year &
                    FPS @ 640p\\ \midrule
                    D7004G14-A (ours)      & 1280*720p@30fps  & \$20 & N/A  & 25\\
                    OV 2710         & 1920*1080@30fps  & \$20   & 2017 & 33\\
                    ELP Super Mini  & 1280*720p@30fps  & \$30   & 2015 & 15\\ \bottomrule
                \end{tabular}%
    \end{table}

    \section{Discussion}

    Our prototype sensor showed mostly linear responses
    under dynamic loading. While the linearity is not precise, these results
    still validate the underlying hypothesis that with fiducials it is possible
    to collect data on all three axes of force and three axes of torque. Further
    design iterations could improve on these results, although this approach is unlikely to 
    achieve the ~0.1\% accuracy claimed for strain-gauge based force-torque
    sensors.

    \subsection{Design Goals}
    \label{sec:conclusion}

    The sensor can now be evaluated against the goals specified previously in
    \cref{sec:designgoals}. The sensor design is indeed responsive in all six axes
    (after our pivot from one tag to two tags, as well as using a much brighter
    LED). Additionally, for grasping applications, the calculations in
    \cref{eqn:maxdisplacement} shows that if a much stiffer spring were chosen
    so that 40~N of load could be applied without exceeding the $y_{range}$,
    the sensor would still have better than 0.1~N of sensitivity.

    The qualitative design goals were also met. The sensor is small,
    measuring only 3.6~cm by 3.1~cm by 5.1~cm in size.
    The sensor is inexpensive, with the majority of the cost being a \$20
    webcam. The sensor is robust and has survived multiple plane trips and the
    occasional throw or drop. The sensor is also easy to modify. The light shield
    can easily be unbolted to
    change the fiducials, or re-printed in an hour to accommodate different
    designs (e.g. a single-tag vs. dual-tag design). Fabrication is easy and
    non-toxic, requiring no degassing machine (as with elastomer-based sensors)
    nor electrical discharging machines (as with custom strain-gauge based designs).
    The sensor by design does not suffer from thermal considerations (as in
    \cite{Guggenheim2017RobustAI}) or electrical noise (as with designs based on
    strain gauges). %

    \subsection{Error Sources}

    An important consideration is the coordinate origin around which measurements
    are made. As load must be applied to the spring platform on which the tags
    are glued, the origin around which measurements are collected may be
    different than desired, although a linear offset matrix should suffice to
    correct for this. Our six-axis measurement reflects a combination of a camera
    pose estimation and mechanical coupling, each of which can introduce errors.
    In the following section on sensor improvement, we focus on camera sensor issues.

    \subsection{Sensor Improvements}

    \subsubsection{Fiducial Changes}

    Unlike the standard use cases for ArUco markers, we do not care about
    distinguishing multiple objects and care more about the quality of the pose
    estimate for a tag guaranteed to be in-frame. A custom fiducial (perhaps
    solely a checkerboard) could improve the force-torque measurements. 

    \subsubsection{Noise in z-axis}

    The sensor is noisy in force and torque measurements along the $z$-axis.
    To address this, one possibility is to use a mirror and two tags which are laid flat
    on the $xy$ plane and the $yz$ plane respectively. The "sideways" tag (on
    the $yz$ plane) has good sensitivity to $z$-axis displacements, and the flat
    $xy$ plane tag addresses rotations around the $z$-axis. A 45-degree
    mirror then allows the camera to also observe the "sideways" tag on the $yz$
    plane. On the downside, the small mirror could make assembly difficult.

    \subsubsection{Sensor Size}

    Closer placement of the tag, to minimize the size of the sensor, may also be
    desired – this would necessitate a custom lens for the camera to allow for
    closer focus (e.g. a macro lens).  Miniaturization could also be
    accomplished with a smaller camera, as in \cite{WardCherrier2018TheTF}. %

    \subsubsection{Replacing Springs}

    The use of springs means that the sensor may behave poorly in
    high frequency domains. Replacing the springs with another mechanism, such
    as a Stewart platform, could allow custom tuning of the
    response.  Another possibility would be to fill
    the gap between the camera and the tag with optically clear material that
    would be resistant to high frequency inputs. \cite{Kiva2016FingernailWS}
    used a similar idea with a magnet and hall effect sensor, for a three-axis force
    sensor. However, such a design would complicate fabrication and potentially
    make camera calibration difficult due to image warping.

    \begin{table}[htbp]
        \vspace{2mm} %
        \caption{List of components and approximate costs.}
        \label{tbl:bom}
        \centering
            \resizebox{0.9\textwidth}{!}{%
                \begin{tabular}{lll}
                    \toprule
                    Part          & Details   & Cost  \\ \midrule
                    Camera        & Mini Camera module, AmazonSIN: B07CHVYTGD& \$20 \\
                    LED and 2 wires & Golden DRAGON Plus White, 6000K, 124 lumens& \$2 \\
                    4 springs     & Assorted small springs set & \$5  \\
                    3D printed pieces  & PLA filament  & \$5  \\
                    Heat-set Threaded Inserts       & Package of 50 from McMaster-Carr (use 2) & \$1  \\
                    Misc. Bolts   & Hex socket head      & \$1  \\
                    Epoxy         & 5 minute             & \$5  \\ \bottomrule
                \end{tabular}%
            }
        \vspace{10mm} %
    \end{table}

\section{Conclusion}
    We present a novel type of six-axis force-torque sensor using fiducial tags
    and a webcam. The design is fast to fabricate and simple to use, and is also
    strong enough to survive drops and crashes common in contact-rich tasks such
    as robotic grasping.  With only 3D-printed custom components, the design
    needs minimal technical expertise to adapt to applications ranging from
    manipulation to human-computer interaction research. The open-source design
    also allows for direct integration in designs for tasks such as grasping
    where sensor size is important.  This fiducial-based sensor is less accurate
    than commercial force-torque sensors, but is also orders-of-magnitude less
    expensive -- commercial sensors can cost thousands of dollars, while the
    parts cost of our sensor is under \$50 (see \cref{tbl:bom}). These combined
    advantages of our prototype sensor validates the general design principle of
    using 3D pose estimates from printed fiducials to create a six-axis
    force-torque sensor. Future work on improving the $F_z$ and $M_z$ axes could
    allow for an inexpensive, user-friendly, and robust alternative to current
    commercial sensors, opening up a new range of use cases for six-axis
    force-torque sensors. %
\chapter{Offbeat but On-Track: A Conclusion}
\label{chap:conclusion}

\color{black}

In this thesis, my research interests lay in two directions: one on applications of computer science to counter-trafficking, and one on applications of computer vision to tactile sensing. The first three chapters focus on the first topic of counter-trafficking, with the first two chapters devoted to the illicit massage industry and the third chapter focused on the financial sector. The last two chapters focus on the topic of tactile sensing.

In the first chapter, I consider the illicit massage industry (IMI) and the question of how to estimate if a given massage business is more or less likely to have illicit activity, which I approximate as whether a business is listed in Rubmaps or not. I show that text reviews from the general purpose website, Google Maps, can be used to replicate business listings on the IMI-specific website Rubmaps. This allows for a more robust data ecosystem for studying the spatiotemporal trends of the IMI, as Rubmaps does not serve as a single failure point. Additionally, the text reviews provide insights into certain aspects of the IMI. I show that on average, reviews for illicit massage businesses (IMBs) are more likely to mention language or nationality and are more likely to mention money. I also show that the average IMB is open later and more days a week than the average legal massage business. Finally, I consider how one could try to remove sensitive features from the classifier. I compare the accuracy of a classifier that does not use text, finding it performs worse but still significantly above null accuracy. I also consider removing all tagged features mentioning language or nationality, and find that the accuracy does not change significantly.

In the second chapter, I consider the task of developing more research interest into the IMI. I create an additional dataset for the IMI by scraping the AMPReviews forum, which is publicly accessible (and on the Internet Archive). I consider three possible use cases for this data. I show that the forum data could be used to expand certain abbreviations specific to the IMI using a Word2Vec embedding. I also show that other domains that have ties to the IMI can be retrieved from the forum dataset and ranked by popularity (number of mentions). I also find using the Word2Vec embeddings that there is some evidence, albeit weak, as it is less discussed than law enforcement actions, that users (mongers) are considered about relationships, and this  could serve as a demand-side intervention. Finally, I discuss several possible directions for future work, including new research questions and practical applications. These include the idea of measuring toxicity and how it leads to real-world harm (e.g., shootings), as well as understanding how to programmatically retrieve demographic information, such as income and ethnicity.

In the third chapter, I consider applications of synthetic data to anti-money laundering in the financial sector. I create an agent-based model using behavioral models as described in operational alerts (reports of red flags) from the regulatory agencies FINCEN for the United States and FINTRAC for Canada. These describe a pattern of late-night transactions being associated with sex trafficking. I create two models, one that outputs transactions at certain timestamps according to a Gaussian model, and another that creates multiple types of agents that differ para- metrically in their transaction patterns. The model tracks which agents transact with which other agents and thus outputs network data in addition to tabular timestamp data. I show the appli- cations to downstream tasks such as anomaly detection and how the synthetic data can be used to quickly evaluate the suitability of different classifiers, including decision trees, isolation forest, and gaussian mixture models. I show the advantages of having a parameterized agent-based model which is easy to understand and modify: with my model, one can easily adjust what percent of transactions are illicit. I conclude with the observation that the downside of agent-based models is that a lot of hand-crafted work goes into deciding which assumptions to make, and the level of abstraction may not lend itself to data that closely matches real-world data. Both agent-based models and other approaches (e.g. non-parametric or deep learning models) can serve different purposes in this case.

The final two chapters switch to the topic of computer vision and tactile sensing in robotics. In the fourth chapter, I cover my work with my co-authors on the Digger Finger. With this sensor, a small Raspberry Pi camera is pointed at a transparent piece of acrylic cylinder, with the tip cut at an angle to create a wedge shape. A clear elastomer is bonded to the acrylic, and on the outermost surface the elastomer is coated in a Lambertian (or semi-specular) membrane (sprayed on as a pigmented aerosol paint as the gel cures). The camera captures a picture of any object pressed into the gel. A 3D reconstruction of the object is possible with photometric stereo, specifically by comparing the shadows cast by three colors of light from three different directions. Red and green fluorescent paint is used for two of the colors to allow for miniaturization (compared to previous iterations, which required LEDs and a circuit board to fit along the edge of the finger). Integrating a vibrating motor to the Digger Finger, this work also  demonstrates how increasing the motor voltage (and therefore amplitude of vibration) decreases the force required to plunge into two different granular media, sand and rice. For the use case of identifying buried objects, we press 3D printed objects of multiple shapes (circle, hex, square, and triangle) into the sensor in the presence of sand and rice. After augmenting this training data with Gaussian noise, random cropping, and random rotations,  the results show that a convolutional network (ResNet50) that is pre-trained on ImageNet and fine-tuned on this image set of approximately 1,500 images can achieve extremely high accuracy classifying between the different shapes.

In the final chapter, I cover my work on creating a low-cost fiducial-based six-axis force-torque sensor. I use a consumer webcam and off-the-shelf 6D pose estimation (X, Y, Z position and the orientation, or rotation, about each of the XYZ axes). This pose-estimation relies on small tags, or fiducials, with known patterns (namely, parallel straightedges). By sticking the fiducial on a 3D platform that is free to rotate in all axes (floating above the camera using four springs) and then calibrating the displacement against known forces and torques, this creates a force-torque sensor. Despite the downsides of the sensor (e.g., hysteresis of the springs, greater sensitivity moving the platform sideways against the springs rather than directly up or down the springs), the sensor is two orders of magnitude cheaper than other force-torque sensors, and easy to modify. This work is accompanied by  design files, both hardware and software, alongside the published paper.

\textbf{Future Work}

The benefits of academia include the ability to freely release  results and artifacts, including data and machine learning methods, to the public. For the future, I hope to promote and build community around the topic of counter-trafficking in computer science research. To this end, I intend to take the datasets that I have created in Chapters 1  and 2 and organize a hackathon where participants come together to create projects such as visualizations of the data.

\normalcolor

\singlespacing

\clearpage
\printbibliography
\addcontentsline{toc}{chapter}{References}

\newpage

\begin{figure}
  \vspace{50pt}
  \centering
    \includegraphics[width=0.51\textwidth]{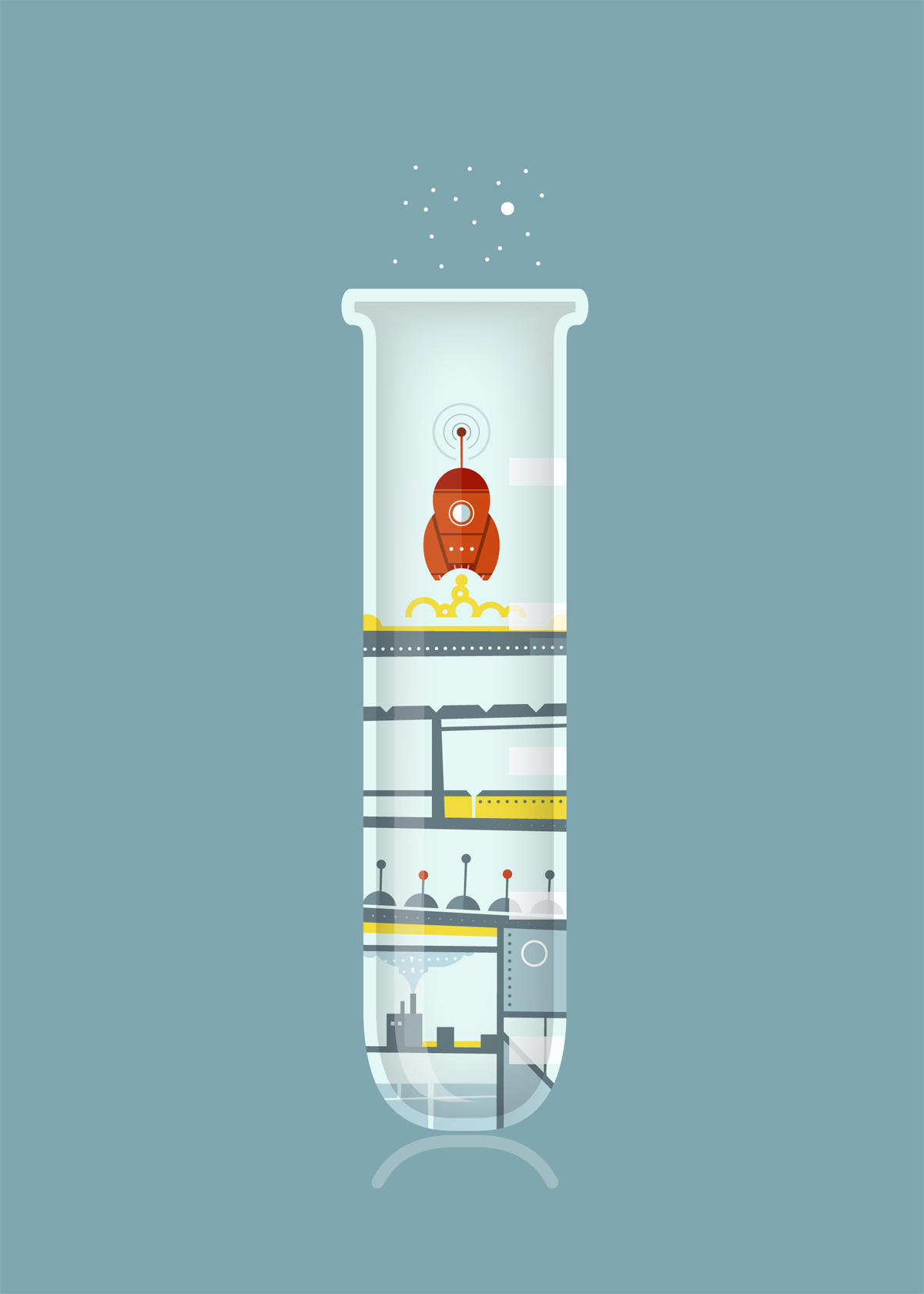}
\end{figure}

\begin{center}
\parbox{200pt}{\lettrine[lines=3,slope=-2pt,nindent=-4pt]{\textcolor{SchoolColor}{T}}{his thesis was typeset} using \LaTeX, originally developed by Leslie Lamport and based on Donald Knuth's \TeX. The body text is set in 11 point Egenolff-Berner Garamond, a revival of Claude Garamont's humanist typeface. The above illustration, ``Science Experiment 02'', was created by Ben Schlitter and released under \href{http://creativecommons.org/licenses/by-nc-nd/3.0/}{\textsc{cc by-nc-nd 3.0}}. A template that can be used to format a PhD thesis with this look and feel has been released under the permissive \textsc{mit} (\textsc{x}11) license, and can be found online at \href{https://github.com/suchow/Dissertate}{github.com/suchow/Dissertate} or from its author, Jordan Suchow, at \href{mailto:suchow@post.harvard.edu}{suchow@post.harvard.edu}.}
\end{center}

\end{document}